\documentclass{SCIS2021}
\usepackage[english]{babel}
\usepackage{makecell}
\usepackage{multirow}
\usepackage{amsthm}
\usepackage{relsize}
\usepackage{amsmath}
\usepackage{enumitem}
\usepackage[colorlinks,linkcolor=black]{hyperref}
\newtheoremstyle{exampstyle}
{0.0em} 
{0.0em} 
{} 
{1em} 
{\bfseries} 
{.} 
{1em} 
{} 
\usepackage{balance}
\theoremstyle{exampstyle}

\newcommand{\tabincell}[2]{\begin{tabular}{@{}#1@{}}#2\end{tabular}} 

\usepackage{CJKutf8}


\DeclareMathOperator{\supp}{supp}

\begin{document}
	\ArticleType{REVIEW}
	\Year{2025}
	\Month{}
	\Vol{}
	\No{}
	\DOI{}
	\ArtNo{}
	\ReceiveDate{}
	\ReviseDate{}
	\AcceptDate{}
	\OnlineDate{}
	\title{Continuous Representation Methods, Theories, and Applications: An Overview and Perspectives}{Continuous Representation Methods, Theories, and Applications: An Overview and Perspectives}
	\author[1]{Yisi Luo}{}
	\author[2]{Xile Zhao}{xlzhao122003@163.com}
	\author[1]{Deyu Meng}{dymeng@mail.xjtu.edu.cn}
	\AuthorMark{Yisi Luo}
	\AuthorCitation{Yisi Luo, Xile Zhao, Deyu Meng}
	\address[1]{School of Mathematics and Statistics, Xi'an Jiaotong University, Xi'an {\rm 710000}, China.}
	\address[2]{School of Mathematical Sciences, University of Electronic Science and Technology of
		China, Chengdu {\rm 610000}, China.}
\abstract{
Recently, continuous representation methods emerge as novel paradigms that characterize the intrinsic structures of real-world data through function representations that map positional coordinates to their corresponding values in the continuous space. As compared with the traditional discrete framework, the continuous framework demonstrates inherent superiority for data representation and reconstruction (e.g., image restoration, novel view synthesis, and waveform inversion) by offering inherent advantages including resolution flexibility, cross-modal adaptability, inherent smoothness, and parameter efficiency. In this review, we systematically examine recent advancements in continuous representation frameworks, focusing on three aspects: (i) Continuous representation method designs such as basis function representation, statistical modeling, tensor function decomposition, and implicit neural representation; (ii) Theoretical foundations of continuous representations such as approximation error analysis, convergence property, and implicit regularization; (iii) Real-world applications of continuous representations derived from computer vision, graphics, bioinformatics, and remote sensing. Furthermore, we outline future directions and perspectives to inspire exploration and deepen insights to facilitate continuous representation methods, theories, and applications. All referenced works are summarized in our open-source repository: {\tt\url{https://github.com/YisiLuo/Continuous-Representation-Zoo}}.}
	\maketitle
\section{Introduction}
In the era of big data, reconstructing high-quality data information from incomplete measurements, noisy observations, or physical rules and attributes remains a fundamental challenge across diverse domains \cite{SIIMS_denoising_survey,LRTFR,SIAM_review_TV,SCIS_rain}, such as medical imaging reconstruction \cite{Medical_imaging_survey}, satellite remote sensing \cite{GRSM_HSI}, scene reconstruction in graphics \cite{CG_SDF_survey}, and numerical scientific computing \cite{TSP_CP}. Traditional data reconstruction methods often rely on discrete grid-based representations (e.g., vectors, matrices, or higher-order tensors \cite{SIAM_review}) and hand-crafted priors (e.g., sparsity or low-rankness \cite{LR_survey}), which may be inadequate for capturing intrinsic geometric or topological structures of complex real-world data. Especially, these limitations are encountered when handling irregular sampled, heterogeneous modality, or cross-resolution tasks, such as arbitrary-resolution imaging \cite{LIIF}, 3D medical image registration \cite{MICCAI_23_registration}, graphics \cite{CG_SDF_survey}, radar imaging geometry \cite{INR_SAR_detection}, and irregular spatial transcriptomics \cite{NM_INR}.\par 
\begin{figure}[t]
	\scriptsize
	\setlength{\tabcolsep}{0.9pt}
	\begin{center}
		\begin{tabular}{c}
			\centering
			\includegraphics[width=1\textwidth]{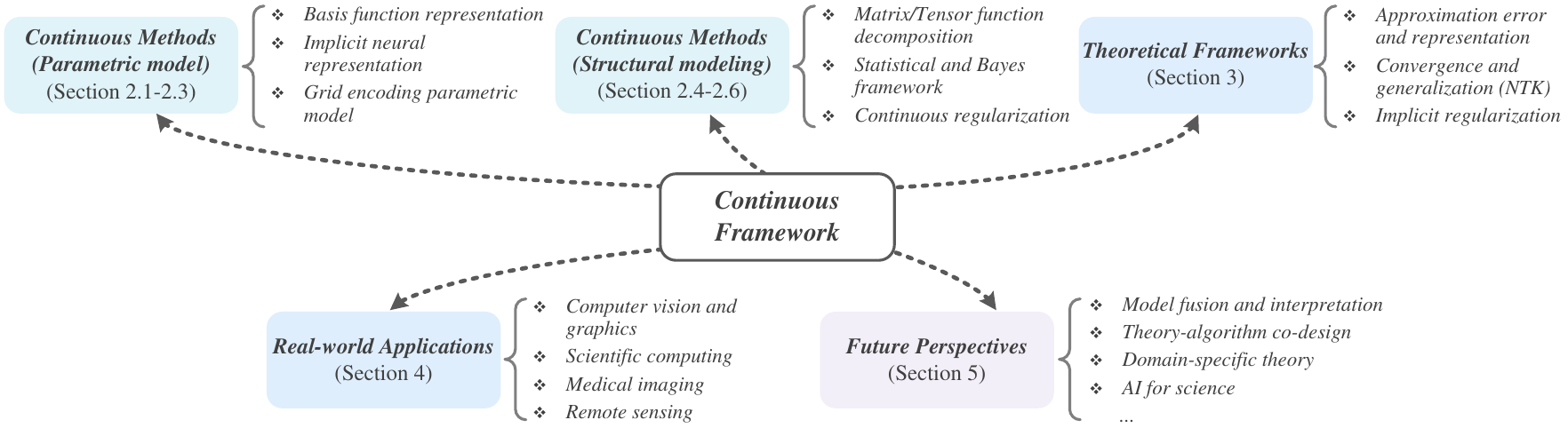}
		\end{tabular}
	\end{center}
	\vspace{-0.4cm}
	\caption{This work is structured to provide a comprehensive overview of continuous representation methods through three perspectives. First, we systematically review continuous representation methods including parametric model architectures and data structural modeling strategies (Section \ref{sec:method}). Second, we review representative theoretical frameworks for formally analyzing the properties and capabilities of continuous representation methods (Section \ref{sec:theory}). Third, we review real-world applications of continuous representation methods derived from diverse fields (Section \ref{sec:applications}). Finally, we discuss future research directions and perspectives aimed at further enhancing the effectiveness, theoretical insights, and applicability of continuous representation frameworks (Section \ref{sec:future}).}
	\label{fig:overview}
\end{figure}
To address these challenges, continuous representation methods (see for classical examples \cite{NeRF,LIIF,SIREN,INR_registration_Jacobian_reg,LRTFR,NeurTV}) have emerged as transformative paradigms for general data reconstruction problems. The {\bf continuous representation} refers to the type of methods that leverage a continuous function representation for discrete data, which maps positional information (e.g., spatial coordinates, temporal time index, or view directions) to the corresponding responses (e.g., image pixels, physical fields, or volume intensities) through certain parametric models such as basis functions or deep neural networks (DNNs). To enhance the effectiveness of continuous representation, regularization terms can be further developed by imposing implicit or explicit structural constraints for the continuous representation function, such as differential operators in loss functions \cite{NeurIPS_nonlinear_PINN_NTK} and functional decomposition \cite{LRTFR}. By embedding discrete data into such smooth, resolution-independent continuous spaces, these methods enable many advantages, such as global feature modeling, adaptability to irregular multi-dimensional data, resolution independence, and interpretable learning theories. For instance, the implicit neural representations (INRs) \cite{SIREN,PE,LIIF} parameterize signals as continuous functions using neural networks, enabling seamless interpolation and super-resolution. Tensor functional decomposition frameworks \cite{LRTFR,Shikai_ICLR_24} exploit the multi-linear structure of high-dimensional data through low-rank factorizations of multivariate functions. Statistical continuous representation methods \cite{Shikai_25_LRTFR,Shikai_ICLR_24,Shikai_ICML_24} encode temporal correlations in continuous time-indexed domains that are important for streaming or time series data analysis. Such continuous approaches inherently avoid discretization errors and achieve parameter efficiency—the discrete data can be implicitly represented by a parametric model that holds much fewer parameters\footnote{This advantage can be shown using tensor representation. Consider a tensor ${\cal A}\in{\mathbb R}^{n_1\times n_2\times n_3}$ and a parametric model $f(x,y,z):{\mathbb R}^3\rightarrow{\mathbb R}$ that maps the index of ${\cal A}$ to its value. According to the universal approximation theorem \cite{Universal_Approximation}, a shallow DNN of depth two can losslessly approximate this continuous function, which holds much fewer parameters than the tensor $\cal A$ itself ($n_1n_2n_3$).}. The parameter efficiency would be critical for tasks with large-scale datasets, e.g., 5D seismic data interpolation \cite{5D_seismic_INR} or video space-time super-resolution \cite{VideoINR}. The continuous representation also holds advantages for further integrating domain-specific knowledge, such as the physics-informed constraints for full-waveform inversion \cite{IFWI_TGRS}, or quasi-static motion priors for magnetic resonance imaging (MRI) \cite{ICLR_MRI_25}, bridging the gap between generic parametric models and application-driven requirements. Overall, continuous representation methods have emerged as increasingly effective paradigms for diverse data reconstruction tasks. \par 
Despite rapid progress, the field remains generally fragmented. Existing works often focus on isolated aspects, such as general continuous representation designs (e.g., Chebyshev expansions for functions \cite{SIAM_cheb}, Fourier features INR \cite{PE}, and wavelet INRs \cite{WIRE}), theoretical-oriented analyses (e.g., convergence analyses for DNNs, e.g., from the neural tangent kernel theory \cite{PE,NTK,untrained_survey_2025} or implicit regularization perspective \cite{Implicit_regularization_deep_CP,LRTFR}), or domain-specific applications (e.g., continuous hyperspectral imaging \cite{HSI_LIIF_24}, deformable registration \cite{MICCAI_23_registration}, and functional decomposing partial differential equations (PDEs) \cite{FTD_PINN}), without synthesizing their interplay. These interplay parts are important for understanding deepen insights into these methods and developing further improvements. For example, the widely-studied INR-based arbitrary-scale image super-resolution \cite{LIIF} actually holds similar training paradigms with the neural operator \cite{NO,FNO}, a way to approximate numerical PDE using neural networks, since both of them take some low-dimensional inputs (coordinates and latent vectors) and synthesize high-resolution responses. For another example, the tensorial radiance fields \cite{TensorNeRF} used for 3D scene reconstruction actually holds intrinsic similarity with a bunch of tensor decomposing functional representations \cite{SIAM_cheb,LRTFR,Functional_CP_Longitudinal,FunTT} analyzed in scientific computing. The classical Fourier basis functional representation \cite{TSP_CP} holds similar ideas with more recent INR methods in Fourier basis-enhanced neural representations \cite{PE}, to name but a few.\par
\begin{figure}[t]
	\scriptsize
	\setlength{\tabcolsep}{0.9pt}
	\begin{center}
		\begin{tabular}{c}
			\centering
			\includegraphics[width=1\textwidth]{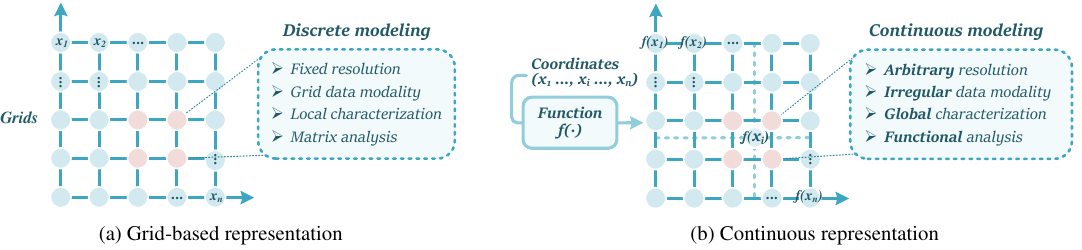}
		\end{tabular}
	\end{center}
	\vspace{-0.4cm}
	\caption{The figure illustrates a comparative analysis between discrete grid modeling and emerging continuous representation modeling approaches. In contrast to conventional grid representations, the novel continuous representation paradigm employs a parametric mapping function $f(\cdot)$ that maps continuous-domain coordinates to their corresponding values, deriving intrinsic advantage such as resolution flexibility, irregular data modeling ability, global characterization of data structures, and interpretable learning theories (such as functional decomposition analysis \cite{NeurIPS_FtSVD,LRTFR}) that extends beyond discrete grids.}
	\label{fig1}
\end{figure}

  
The main aim of this work is to give a comprehensive understanding and review on continuous representation methods including continuous representation parametric model and structural modeling designs, theoretical frameworks and current results that formally analyze the mathematical properties of continuous methods, and diverse domain-specific data reconstruction applications of continuous representation methods. The expectation is to bridge the information gap between different subareas within the continuous representation research field, thus enabling new exciting and transformative research directions unexplored before. An overview of the advancements in continuous representation methods, theories, and applications introduced in this work is summarized in Fig. \ref{fig:overview}, and the fundamental concepts and advantages of continuous representation approaches are illustratively presented in Fig. \ref{fig1}. We outline the main contributions of this review as follows.
\begin{itemize}
\item We provide a systematic review of continuous representation methods for data representation and reconstruction, structured around three parts---review of continuous methodology designs, theoretical insights of these continuous methods, and corresponding applications across diverse fields, deciphering the rapid evolution of related fields.
\item We highlight promising future directions and chart a prospect for addressing open challenges for continuous method designs, theoretical developments for continuous frameworks, and utilizing them to facilitate real-world applications. We aim to inspire interdisciplinary collaborations to advance the data reconstruction methods inspired by continuous representation frameworks.
\end{itemize}\par
The remainder of this paper is structured as follows. In Section \ref{sec:method}, we review the developments of various continuous representation models and structural modeling methods. In Section \ref{sec:theory}, we delve into the theoretical foundations of continuous representation methods such as approximation error analysis and neural tangent kernel theories. In Section \ref{sec:applications}, we review various application scenarios of continuous representation methods such as medical imaging, remote sensing, and bioinformatics. In Section \ref{sec:future}, we introduce promising future directions and perspectives to inspire further explorations of continuous representation methods.
\begin{table}[h]
	\centering
	\scriptsize
	\belowrulesep=0pt
	\aboverulesep=0pt
	\setlength{\tabcolsep}{2pt}
	\renewcommand{\arraystretch}{1.1} 
	\caption{Review of some representative continuous representation parametric models.\label{tab:method}}
	\begin{tabular}{ll|l|c|c}
		\midrule
		\textbf{Category} & \textbf{Method} & \textbf{Description} & \textbf{Year} & \textbf{Ref.} \\
		\midrule
	\multirow{9}*{\tabincell{c}{Basis Function\\ Representation}} & \tabincell{l}{(NMF) Smooth nonnegative\\matrix factorization}  & \tabincell{l}{Utilize Gaussian basis functions to parameterize nonnegative\\ matrix factorization for multi-way data analysis.} & 2015 & {\cite{SP_smooth}}\\
		\cmidrule{2-5}
				  ~& \tabincell{l}{(STD) Smooth tensor \\decomposition} & \tabincell{l}{Introduce smoothed Tucker decomposition represented by \\Fourier basis functions to incorporate smoothness.} & 2017 & {\cite{ICML_STD}}\\
\cmidrule{2-5}
	~& \tabincell{l}{(Chebfun) Chebfun in three \\dimensions} & \tabincell{l}{Trivariate function representation based on low-rank tensor\\ decomposition through finite Chebyshev expansions.} & 2017 & {\cite{SIAM_cheb}}\\
		\cmidrule{2-5}
		 ~& \tabincell{l}{(TT) Functional \\tensor-train}  & \tabincell{l}{Functional tensor-train decomposition parameterized by\\ Gaussian kernels for regression problems.} & 2018 & {\cite{TT_regression}}\\\cmidrule{2-5}
		 ~& \tabincell{l}{(CPD) CP decomposition \\of multivariate functions}  & \tabincell{l}{Generalize the CP decomposition from tensors to multivariate\\ functions by finite multi-dimensional Fourier series.} & 2021 & {\cite{TSP_CP}}\\
		\midrule
	\multirow{11}*{\tabincell{l}{Implicit Neural \\Representation}}& \tabincell{l}{(PE) Fourier features \\positional-encoding INR} & \tabincell{l}{Propose Fourier feature mapping for positional-encoding \\that enables INR to learn high-frequency details.} & 2020 & \cite{PE} \\\cmidrule{2-5}
& \tabincell{l}{(SIREN) INR with periodic \\activation functions} & \tabincell{l}{Leverage periodic activation functions for INR and verify \\ the effectiveness of SIREN for continuous representation.} & 2020 & \cite{SIREN} \\\cmidrule{2-5}
& \tabincell{l}{Meta-learning for INR\\ initialization} & \tabincell{l}{Apply meta-learning to
	learn initial weights for INRs based on\\ signal classes for faster convergence and better generalization.} & 2021 & \cite{Meta_Init_INR} \\\cmidrule{2-5}
& \tabincell{l}{(WIRE) Wavelet INR} & \tabincell{l}{Use the complex Gabor wavelet function as the activation \\function of INR to improve robustness.} & 2023 & \cite{WIRE} \\\cmidrule{2-5}
& \tabincell{l}{(FINER) Flexible spectral-\\bias tuning for INR} & \tabincell{l}{Introduce variableperiodic activation functions and initialize\\ INR bias differently for flexible spectral-bias tuning.} & 2024 & \cite{FINER} \\\cmidrule{2-5}
& \tabincell{l}{(FR-INR) Fourier repara-\\meterization for INR weights} & \tabincell{l}{Learn the coefficient matrix of fixed Fourier bases for\\ linear weights of INR to alleviate spectral bias.} & 2024 & \cite{INR_Reparameterize} \\
		\midrule
		\multirow{9}*{\tabincell{l}{Grid Parametric\\Encoding }} & \tabincell{l}{(InstantNGP) Multi-resolution\\hash grid encoding}&\tabincell{l}{Use a smaller INR augmented by a multi-resolution\\ hash table of trainable feature vectors on grids.} & 2022 & \cite{InstantNGP} \\\cmidrule{2-5}
		&\tabincell{l}{(DVGO) Direct voxel grid \\optimization}&\tabincell{l}{A density voxel grid for scene geometry and a feature \\voxel grid for view-dependent appearance.} & 2022 & \cite{DVGO} \\\cmidrule{2-5}
		&\tabincell{l}{(Plenoxels) Sparse 3D grid \\encoding}&\tabincell{l}{Efficient scene representation as a sparse 3D grid \\with spherical harmonics.} & 2022 & \cite{Plenoxels} \\\cmidrule{2-5}
		&\tabincell{l}{(DINER) Disorder-invariant \\hash-based INR}&\tabincell{l}{Disorder-invariant INR by projecting coordinates into a \\learnable hash table, alleviating spectral bias.} & 2024 & \cite{DINER} \\
		\cmidrule{2-5}
		& \tabincell{l}{{(RHINO)} Regularized grid \\encoding}&\tabincell{l}{Regularizing hash-based grid encoding using coordinate INR\\ concatenated with parametric grid input.} & 2025 & \cite{MLP+InstantNGP} \\
		\midrule
	\end{tabular}
\end{table}
\section{Continuous Representation: Parametric Model and Structural Modeling}\label{sec:method}
In this section, we systematically review the advancements of continuous representation methods, focusing on two groups: {\bf parametric model} design and data {\bf structural modeling} methods. The parametric models, including basis function representation models, implicit neural representations, and grid encoding-based models, aim to construct effective parameterization schemes for the continuous functions, hence enhancing data representation and reconstruction accuracy. The structural modeling approaches, such as matrix/tensor function decomposition, statistical and Bayes frameworks, and continuous regularization methods based on continuous representations, encode prior knowledge of data and structural constraints to further enhance robustness and generalization abilities of continuous representations. An overview of continuous parametric model designs and structural modeling methods is shown in Table \ref{tab:method} and Table \ref{tab:modeling}.   
\subsection{Basis Function Representation}
The concept of basis function continuous representation lies in the core idea of decomposing a multivariate function into a series of low-dimensional functions (in most cases, univariate functions), which are parameterized by some predefined basis functions such as Gaussian kernels or Fourier series. These basis function representations enjoy good interpretability and solid theoretical properties, e.g., existence and identifiability \cite{TSP_CP}, allowing for a wide range of data science problems.\par 
Yokota et al. \cite{SP_smooth} proposed the smooth nonnegative matrix and tensor factorizations for robust multi-way data analysis. This model is based on the matrix or tensor factorization parameterized by Gaussian basis functions in the form of
\begin{align*}
	{\Phi}(i,n) =
	\exp\left[-\frac{(i - n\Delta t)^2}{2\sigma^2}\right],
\end{align*} 
and solves an optimization model based on the nonnegative matrix factorization (NMF) to recover information from an observed data $Y$:
\[
\min_{{W}, {X}} \frac{1}{2} \| {Y} - {\Phi}{W}{X} \|_F^2, 
\quad \text{s.t. } {\Phi}{W} \geq 0, \, {X} \geq 0.
\] 
This NMF model encodes smoothness constraints on nonnegative factors using smooth basis functions, hence facilitating better physical interpretation and robustness w.r.t. noise. Debals et al. \cite{spl_smooth} proposed the NMF using nonnegative polynomial approximations, in which the factors of NMF are modeled by a parametric representation of finite-interval nonnegative polynomials to obtain an optimization problem without external nonnegativity constraints, which can be solved using quasi-Newton or nonlinear least-squares methods. The polynomial model also guarantees smooth solutions for noise reduction through smooth basis function representations. \par
Similarly, Imaizumi et al. \cite{ICML_STD} proposed the smooth tensor decomposition using the Tucker functional tensor decomposition parameterized by basis functions, which leverages smoothness using the sum of a few Fourier basis functions. They theoretically showed that, under the smoothness assumption of the tensor, the smooth tensor decomposition achieves a better reconstruction error bound for tensor recovery and interpolation.\par 
Gorodetsky and Jakeman \cite{TT_regression} proposed functional TT decomposition-based continuous representation model for regression problems, in the form of 
\[
f(x_1, x_2, \ldots, x_d) = \sum_{i_0=1}^{r_0} \sum_{i_1=1}^{r_1} \cdots \sum_{i_d=1}^{r_d} f_1^{(i_0 i_1)} (x_1) f_2^{(i_1 i_2)} (x_2) \cdots f_d^{(i_{d-1} i_d)} (x_d),
\]
where the univariate factor functions $f_k^{ij}(x_k)$ are parameterized by Gaussian kernels $\Phi_k(x)=\exp(-\frac{(x-\theta_k)^2}{\sigma^2})$ with learnable Gaussian coefficients $\theta$ optimized by stochastic gradient descent or quasi-Newton methods. The functional TT model excels in the task of low-multilinear-rank regression.\par
Kargas et al. \cite{Nonlinear_AAAI} proposed the nonlinear system identification via tensor completion, which identifies a general nonlinear function $y = f(x_1,\cdots,x_N)$ from input-output pairs via a tensor completion problem. Their later work \cite{TSP_CP} proposed the functional formulation of the canonical polyadic (CP) tensor decomposition using Fourier basis functions in the form of 
\[
\Phi_0(x) =1,\; \Phi_k(x) =\sqrt
2 \cos(k\pi x), 
\]
where only the cosine extension is used for efficiency. Then the CP decomposition of a multivariate function $f$ parameterized by such basis functions has the form 
\[
f(\mathbf{x}) = \sum_{k_1=0}^{K-1} \cdots \sum_{k_N=0}^{K-1} \sum_{r=1}^R 
\prod_{n=1}^N a_n^r [k_n] \Phi_{k_n}({\bf x}[n]),
\]
where the Fourier coefficients $a_n^r$ are optimized towards fitting the training data. The authors provided an existence and uniqueness guarantee for the CP functional decomposition model from the finite multidimensional Fourier series perspective, and showed that the model can be empirically effective for real-world data regression problems. Similarly, Sort et al. \cite{Functional_CP_Longitudinal} proposed to represent a high-dimensional functional tensor as a low-dimensional set of functions and feature matrices in the form of CP decomposition with Fourier basis functions, along with a probabilistic latent model in the decomposition to enable sparse and irregular sampling, with applications to longitudinal data modeling.\par 
Earlier theoretical works showed that some specific types of multivariate functions can be exactly factorized into explicit forms of low-dimensional functions. For instance, Kunkel and Mehrmann \cite{smooth_matrix_factorization_1991} studied the numerical factorization of matrix-valued functions, and applied it in the numerical solution of differential algebraic Riccati equations. Oseledets \cite{TTF} proposed constructive approximation (explicit representation) of some specific multivariate functions (e.g., the polynomial and sine functions) in the tensor-train (TT) decomposition format: 
\[ f(x_1, \ldots, x_d) \approx \sum_{\alpha=1}^r u_1(x_1, \alpha) u_2(x_2, \alpha) \cdots u_d(x_d, \alpha). \]
For example, for a function $f(x_1, \ldots, x_d) = \sin(x_1 + x_2 + \cdots + x_d)$, the functional TT decomposition has the form 
\[ f = (\sin x_1 \quad \cos x_1) 
\begin{pmatrix} 
	\cos x_2 & -\sin x_2 \\ 
	\sin x_2 & \cos x_2 
\end{pmatrix} 
\cdots 
\begin{pmatrix} 
	\cos x_{d-1} & -\sin x_{d-1} \\ 
	\sin x_{d-1} & \cos x_{d-1} 
\end{pmatrix} 
\begin{pmatrix} 
	\cos x_d \\ 
	\sin x_d 
\end{pmatrix}. \]
The TT decompositions of some other functions, e.g., the polynomial function, were also obtained \cite{TTF}. The obtained functional decompositions are useful for the construction of efficient algorithms in high-dimensional problems. Similarly, Tichavsk and Straka \cite{TT_function_EUSIPCO} proposed an alternating least squares method for fitting a TT to an arbitrary number of tensor fibers, allowing robust and flexible modeling of multivariate functions that contain noise. They provided examples of the algorithm for decomposing Rosenbrock function and quadratic function. Chen et al. \cite{ICML_feature_function_matrix} studied the collaborative filtering problem and formalized the problem as a general functional matrix factorization, which learns feature functions based on
training data, and the learnable feature functions are parameterized by piece-wise linear functions.\par
Hashemi and Trefethen \cite{SIAM_cheb} studied Tucker factorization of trivariate functions and proposed constructive algorithms for approximating trivariate functions in Tucker format, e.g., 
\[
f(x,y,z) \approx \mathcal{T} \times_1 {A}(x) \times_2 {B}(y) \times_3 {C}(z),
\]
where the factor one-dimensional functions are represented by finite sum of Chebyshev polynomials and the core tensor ${\cal T}$ is an $m\times n\times p$ tensor. For instance, the multivariate function
\begin{equation}\label{eq:2.5}
	f(x,y,z)=3x^{7}z+yz+yz^{2}+\log(2+y)z^{3}-2z^{5}
\end{equation}
has an exact decomposition formulation in the Tucker decomposition format:
\begin{equation}\label{f_Tucker}
	A(x) = \begin{bmatrix}1, & x^{7}\end{bmatrix}, \;B(y) = \begin{bmatrix}1, & y, & \log(2+y)\end{bmatrix}, \;C(z) = \begin{bmatrix}z^{5}, & z+z^{2}, & z^{3}, & z\end{bmatrix}.
\end{equation}
Using the Chebyshev polynomial representation algorithm can construct the Tucker decomposition for \eqref{eq:2.5} numerically to obtain the decomposition result \eqref{f_Tucker} \cite{SIAM_cheb}. From the decomposition we can see the trilinear rank of $f$ is $(2,3,4)$: these three numbers are the $x$-rank, the $y$-rank, and the $z$-rank. The $2\times 3\times 4$ discrete core tensor ${\cal T}$ has the following mode-1 unfolding:
\[
{\cal T}_{(1)} = \begin{bmatrix} 
	-2 & 0 & 0 \;\Big{|}& 0 & 1 & 0\;\Big{|}\; 0 \;\; 0 & 1 \;\Big{|}& 0 & 0 &0\\ 
	0 & 0 & 0 \;\Big{|}& 0 & 0 & 0 \;\Big{|}\;0 \;\; 0 & 0 \;\Big{|}& 3 & 0 &0
\end{bmatrix}\in{\mathbb R}^{2\times 12}.
\]
The Tucker decomposition using Chebyshev polynomial representations would benefit fundamental numerical computations of multivariate functions such as differentiation, Laplacian operators, definite and indefinite integrals \cite{SIAM_cheb}. Recently, the Kolmogorov-Arnold network (KAN) \cite{KAN} emerged as a new alternative for function approximation, showing particular interpretability and robustness for scientific applications. Each layer of KAN can be viewed as a combination of spline basis functions, hence linking traditional basis function representation with modern neural representations.\par 
Basis function representations provide principal ways to approximate multivariate continuous functions using predefined basis functions with learnable coefficients, which benefits multi-way data analysis and reconstruction tasks. Nevertheless, predefined basis function representation methods may lack strong representation abilities for depicting complex and irregular real-world data structures. In future work, the spirit of basis function methods can be combined with modern neural representation (such as randomized neural network \cite{Wang_JCP} and KAN-based approaches \cite{KAN}) to enable better characterization of data in signal processing and high-dimensional regression problems in machine learning. Studying the theoretical advantage of basis function methods (such as implicit smoothness \cite{STP_TIP,ICML_STD}) could facilitate understanding and inspire further explorations on the interpretability of more complex continuous representations. 
\subsection{Implicit Neural Representation}
Implicit neural representation has emerged as a transformative parametric model for representing complex, irregular data through continuous coordinate-based mappings using deep neural networks (DNNs). INRs map spatial coordinates to the corresponding values using DNNs. By parameterizing signals as continuous functions of spatial or temporal coordinates, INRs overcome the limitations of discrete sampling, enabling representation at arbitrary resolutions, natural interpolation between samples, and compact modeling of high-dimensional data, which benefit many downstream applications.\par 
The two most classical INRs are the Fourier feature positional-encoding (PE)-based INR \cite{PE} and the sinusoidal periodic activation function-based INR \cite{SIREN}. The PE-based INR \cite{PE} showed that passing input coordinates through a simple Fourier feature mapping enables a multilayer perceptron (MLP) to learn high-frequency functions in low-dimensional
domains. More specifically, given a coordinate vector ${\bf v}\in{\mathbb R}^{d}$, the Fourier mapping maps it to the surface of a higher dimensional hypersphere with a set of sinusoids:
\begin{align*}
	\gamma(\mathbf{v}) = {\rm PE}({\bf v})=\left[ 
	a_1\cos(2\pi\mathbf{b}_1^{\mathrm{T}}\mathbf{v}), 
	a_1\sin(2\pi\mathbf{b}_1^{\mathrm{T}}\mathbf{v}), 
	\dots, 
	a_m\cos(2\pi\mathbf{b}_m^{\mathrm{T}}\mathbf{v}), 
	a_m\sin(2\pi\mathbf{b}_m^{\mathrm{T}}\mathbf{v}) 
	\right]^\mathrm{T}.
\end{align*}
Then passing the positional encoded vector $\gamma(\mathbf{v})$ through a ReLU-based MLP results in effective continuous learning for the low-dimensional input domain of $\bf v$. The underlying reason of the effectiveness of PE is that an MLP with PE (under certain conditions) is equivalent to a kernel regression with a diagonal shift-invariant kernel from the neural tangent kernel perspective (see Section \ref{sec:theory} for details), while conventional MLP without PE leads to a non-diagonal kernel. The kernel regression with a diagonal shift-invariant kernel tends to more easily capture high-frequency information, thus allowing more effective continuous representation using PE.\par 
Similarly, the sinusoidal periodic activation function-based INR \cite{SIREN} (termed SIREN) is also a widely utilized INR method. It was shown that leveraging periodic activation
functions for INRs (instead of using conventional activations such as ReLU) are ideally suited for continuously representing complex natural signals and their derivatives. The authors also proposed principled initialization schemes for SIREN and demonstrated the representation abilities of SIREN for images, wavefields, video, sound, and their derivatives. The SIREN is constructed by stacking sinusoidal layers in the form of 
\[
\mathbf{x}_i \mapsto \phi_i\left(\mathbf{x}_i\right) = \sin\left(\omega_0(\mathbf{W}_i\mathbf{x}_i + \mathbf{b}_i)\right),
\] 
where ${\bf W}_i$ and ${\bf b}_i$ are learnable weights and biases, $\omega_0$ is a frequency hyperparameter (in most times setting $\omega_0=30$). The differentiability of sinusoidal function naturally allows SIREN to tackle physical modeling problems such as surface representation and numerical PDE modeling (e.g., Helmholtz and wave equations considered in \cite{SIREN}). The underlying rationale of the effectiveness of SIREN is similar to that of the PE from the neural tangent kernel perspective. Subsequently, Fathony et al. \cite{MFN} suggested that simply multiplying some sinusoidal or Gabor wavelet basis functions applied to the input coordinates yields comparable performance against PE and SIREN-based INRs.\par 
Later, researchers started to constantly develop more powerful INR paradigms from various aspects. Inspired by harmonic analysis, Saragadam et al. \cite{WIRE}
developed a new INR (termed WIRE) with continuous complex Gabor wavelet activation functions, i.e., 
\begin{equation}
\phi_i({\bf x}_i) = \psi(\mathbf{W}_i\mathbf{x}_i + \mathbf{b}_i; \omega_0, s_0),\; \psi(x; \omega_0, s_0) = e^{j\omega_0 x} e^{-|s_0 x|^2},
\end{equation}
where $\omega_0$ controls the frequency of the wavelet and $s_0$ controls the spread (or width). WIRE enjoys the advantages of periodic nonlinearities such as SIREN due to the complex exponential term and the spatial compactness from the Gaussian window term $e^{-|s_0 x|^2}$. Unlike SIREN, WIRE does not require a carefully chosen set of initial weights due to the Gaussian window, and thus enables robust continuous representation learning w.r.t. the the choice of hyperparameters. Li et al. \cite{AAAI24_Fourierbases} proposed the collaged Fourier bases for INR, which utilizes spatial masks to modulate each global Fourier feature and collages the frequency patches to form better reconstruction of the image's local pattern. Liu et al. \cite{FINER} proposed the variable-periodic activation functions for INRs, where each layer is formulated as
\[
\mathbf{x}_i \mapsto \phi_i\left(\mathbf{x}_i\right) = \sin\left(\omega_0\alpha_i(\mathbf{W}_i\mathbf{x}_i + \mathbf{b}_i)\right),\;\alpha_i=|\mathbf{W}_i\mathbf{x}_i + \mathbf{b}_i|+1.
\] 
It uses the variable-periodic activation function $\sin((|x|+1)x)$ to flexibly tune the supported frequency set of INR to improve performance in continuous signal representation. Jayasundara et al. \cite{WIRE_ICLR_25} proposed the prolate spheroidal wave function (PSWF)-based INR, which exploits the optimal space-frequency domain concentration of PSWF as the nonlinear
mechanism in INRs to better generalize to unseen coordinates. The PSWF $\psi_n(c,t)$ here is the eigenfunction of an integral operator problem 
\[
\int_{-t_0}^{t_0} \psi_n(c,t) \frac{\sin \Omega(x-t)}{\pi(x-t)} dt = \psi_n(c,x) \lambda_n(c),
\] 
which is numerically solved by the expansion of Legendre polynomial $P_k(t)$, i.e., parameterizing the solution as $\psi_n(c,t) = \sum_{k=0}^{\infty} \beta_k^n P_k(t)$ and determining the Legendre coefficients $\beta_k^n$ using recursions. Ramasinghe and Lucey \cite{ECCV_22_unified_activation_INR} proposed an unified perspective on the activation function design of INR, and showed that a large number of non-periodic functions are suitable for continuous representation using INR, such as Laplacian and quadratic functions.\par
Recently, Shi et al. \cite{INR_Reparameterize} proposed the Fourier reparameterized training for INR, which learns coefficient matrix
of fixed Fourier bases to compose the weights of the MLP. The reparameterized training leads to more balanced eigenvalue distribution of the neural tangent kernel, hence leading to improved convergence speed for continuous representation by alleviating the spectral-bias of INR. Hao et al. \cite{MoE_INR} proposed the levels-of-experts (LoE) framework. For each linear layer of the MLP,
the LoE employs multiple candidate values of its weight matrix, with different layers replicating at different frequencies. For each input, only one of the weight matrices is chosen for each layer. This LoE structure greatly increases the INR model capacity without incurring extra computation. Similarly, Shabat et al. \cite{NeuralExperts} proposed the mixture of experts (MoE) INR that learns local piece-wise continuous functions that simultaneously learns to subdivide the domain and fit it locally. They also introduced novel conditioning and pretraining methods for the gating network of the MoE that improves convergence. Vyas et al. \cite{transfer_learning_INR_NIPS} proposed to learn transferable representations for INRs by sharing initial encoder layers across multiple INRs with independent decoder layers. At the inference stage, the learned encoder representations are transferred as initialization for another input sample, thus enabling knowledge transfer. Zhang et al. \cite{ICML_teacher} proposed a nonparametric teaching scheme for INRs, where the teacher selects signal
fragments for iterative training of the MLP to achieve fast convergence. They unveiled the link between the evolution of INR and that of a function using functional gradient descent in nonparametric teaching, thus expanding the applicability of nonparametric teaching towards deep learning using INRs. 
\par
Other external structure designs are also incorporated into INRs for efficient continuous learning. Fang et al. \cite{CycleINR} proposed the CycleINR, which samples unseen points from the learned INR and uses these points to train a another INR, and then impose cycle-consistency loss between the sampled points of two INRs, leading to higher accuracy for super-resolution. Saragadam et al. \cite{MultiScale_INR} proposed the multi-scale INR, which decomposes the signal into multi-scale orthogonal parts in terms of Laplacian pyramid, and represents small disjoint patches of the pyramid at each scale with small MLPs, leading to increased coarse-to-fine approximation using INR. Li et al. \cite{Superpixel_INR_ECCV} proposed the superpixel-informed INR model for multi-dimensional data, which exploits the semantic information within and across generalized superpixels for improved continuous representation. Cai et al. \cite{Batch_normalization_INR} studied the effect of batch normalization for alleviating the spectral bias of INR from the neural tangent kernel eigenvalue distribution perspective. The cross-frequency INR \cite{CF-INR}  explicitly disentangles the multi-frequency characteristics of data by using the Haar wavelet transform, thus achieving superior accuracy for continuous data representation. Kazerouni et al. \cite{INCODE} proposed to use another network to dynamically adjust key parameters (such as frequencies and phases) of the activation function of the INR, enabling more accurate continuous representations.\par
The internal iterative learning process of INR for each observed data from scratch can also be relaxed by using the meta learning method \cite{meta_transformer_INR}, which builds suitable model parameters for the INR through one step forward propagation for each data. Besides, Tancik et al. \cite{Meta_Init_INR} proposed to use meta learning to learn specific initialization schemes for INR parameters for a particular group of signals. The meta learning strategy coupled with dictionary learning of INR \cite{Dictionary_INR} could further uncover improved representation abilities and convergence efficiency across instances. It is a promising future direction to use advancing meta learning methods to generalize INR methods or using INRs as meta learners to enable fine tuning of larger models, such as by using the low-rank adaptation \cite{LoRA}. 
\subsection{Grid Encoding Parametric Model}
In recent years, parametric models based on grid encoding have achieved remarkable progress in continuous representation, especially in scene representation \cite{NeRF,InstantNGP,scene_representation}. By integrating sparse or multi-resolution grids with learnable feature encodings (e.g., hash tables, voxel grids) and interpolation, these methods efficiently capture fine-grained details while accelerating the convergence of continuous models as compared with conventional INRs. We introduce several representative works along this line.\par
\begin{figure}[t]
	\scriptsize
	\setlength{\tabcolsep}{0.9pt}
	\begin{center}
		\begin{tabular}{c}
			\centering
			\includegraphics[width=0.8\textwidth]{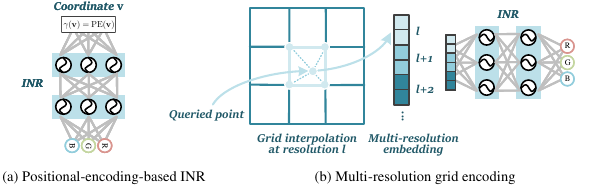}
		\end{tabular}
	\end{center}
	\vspace{-0.4cm}
	\caption{The figure illustrates the structures of the positional-encoding-based INR \cite{PE} and the multi-resolution grid encoding-based INR \cite{InstantNGP}. (a) The PE-based INR maps the input coordinate through a Fourier mapping layer before passing through an MLP for effective continuous representation. (b) The multi-resolution grid encoding-based INR (e.g., InstantNGP \cite{InstantNGP}) constructs learnable feature vectors stored on grids with multiple resolutions. To query the point outside grids, linear interpolation is computed and the interpolated results of different grid resolutions are concatenated and passed through an MLP. Here, the learnable feature vectors are efficiently stored in hash tables.}
	\label{fig:InstantNGP}
\end{figure}
\begin{figure}[t]
	\scriptsize
	\setlength{\tabcolsep}{0.9pt}
	\begin{center}
		\begin{tabular}{c}
			\centering
			\includegraphics[width=0.87\textwidth]{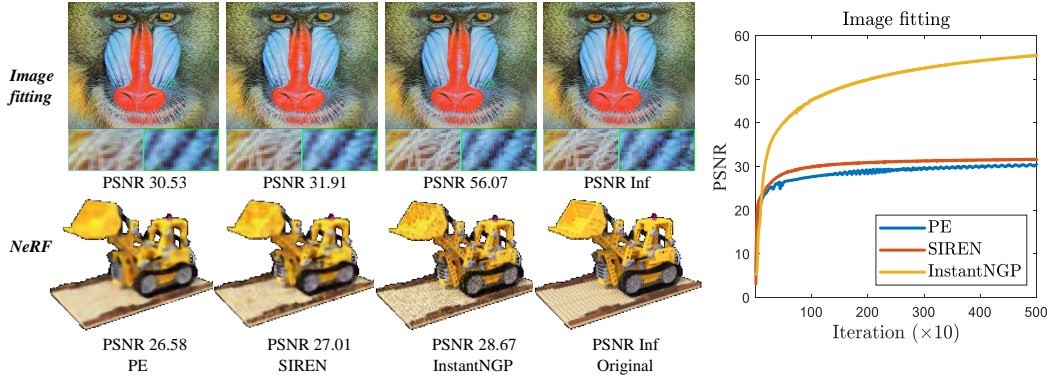}
		\end{tabular}
	\end{center}
	\vspace{-0.4cm}
	\caption{The grid encoding-based method InstantNGP \cite{InstantNGP} achieves faster convergence and captures fine details better as compared with classical INR methods, including positional-encoding (PE) \cite{PE} and SIREN \cite{SIREN}. The top row shows image fitting and the bottom shows novel view synthesis using NeRF \cite{NeRF} of different models with the same iteration step.}
	\label{fig:InstantNGP_com}
\end{figure}

Muller et al. \cite{InstantNGP} proposed the iconic multi-resolution hash encoding method for continuous representation, especially focusing on image or scene representation using neural radiance fields (termed instant neural graphics primitives, or InstantNGP). It uses a smaller INR augmented by a multi-resolution hash table of trainable feature vectors on grids (see Fig. \ref{fig:InstantNGP}). To use the hash table, InstantNGP maps a cascade of grids to corresponding fixed size arrays of trainable feature vectors. The array is treated as a hash table and indexed using a spatial hash function, e.g., 
\[ h({\bf x}) = \left( \bigoplus_{i=1}^d {\bf x}_i \pi_i \right) \mod T, \]
where $\bf x$ denotes the spatial index of grids, $\bigoplus$ is the exclusive-or operation, $\pi_i$ are large prime numbers, and $T$ denotes the size of the hash table. The returning value $h({\bf x})$ indexes the position of the grid $\bf x$ in the hash table. The hash table automatically prioritizes the sparse areas with the most important fine scale detail. Then, any point in the continuous domain can be inferred by using linear interpolation of its neighboring grid feature vectors and passing the interpolated results through an MLP to output the corresponding value. Notably, the InstantNGP can be seen as imposing learnable positional-encoding for INR, where the learnable parameters are stored in discrete grids and those positional-encoding points beyond grids are obtained by linear interpolation using its neighboring grids; see Fig. \ref{fig:InstantNGP}. From another perspective, the InstantNGP can be understood as a complex composite form of linear interpolation functions. It excels in learning fine details of images (scenes) and faster convergence due to the flexibility of learnable positional-encoding embeddings. Some numerical example are shown in {Fig. \ref{fig:InstantNGP_com}}. \par 
Similarly, Sun et al. \cite{DVGO} proposed the DVGO for continuous scene representation, which uses a density voxel grid for scene geometry and a feature voxel grid for view-dependent appearance encoding. The authors also developed post-activation interpolation on voxel density for producing sharp surfaces in lower grid resolution, and imposed special initialization and learning rate schedule to enhance robustness. Yu et al. \cite{PlenOctrees} proposed the PlenOctrees, a method to quickly render novel view images under the scene representation, by training spherical harmonic functions to model view-dependent appearance on the sphere. The color of the novel view image $c({\bf d})$ is calculated by summing the weighted spherical harmonic bases evaluated at the corresponding ray direction $\bf d$: 
\[c({\bf d}) = \left(1+\exp\Big(-  \sum_{\ell=0}^{\ell_{\max}} \sum_{m=-\ell}^{\ell} k_{\ell}^m Y_{\ell}^m ({\bf d}) \Big)\right)^{-1},\]
where $Y_{\ell}^m ({\bf d})$ are spherical harmonic basis functions and $k_{\ell}^m$ are harmonics
coefficients learned by an INR. The trained model can be sampled around the volume to create an octree structure, which can be fine-tuned to improve quality. Keil et al. \cite{Plenoxels} extended PlenOctrees to Plenoxels for efficient scene representation as a sparse 3D grid with spherical harmonics. It achieves end-to-end training of spherical harmonic function representation within a sparse voxel grid without neural networks. The grid encoding coupled with spherical harmonics largely enhances the convergence speed and quality of scene representation for novel view synthesis.\par
Recently, Zhu et al. \cite{DINER} proposed the disorder-invariant INR (DINER) by projecting coordinates into a learnable hash table, alleviating spectral bias of conventional INRs. The hash table rearranges the coordinates of an image where the rearranged image has much lower frequency and thus can be better represented using the subsequent INR. The optimization target of DINER has the form of 
\[
\arg \min_{\theta, \mathcal{H}_M} \mathcal{L} \left( P \left( \{ f_\theta (\mathcal{H}_M^i) \}_{i=1}^N \right), P \left( \{\vec{y}_i \}_{i=1}^N \right) \right),
\]
where $f_\theta$ is an INR, $\mathcal{H}_M^i$ denotes the $i$-th hash key of the hash table, $\vec{y}_i$ denotes the target value, and $P$ is a physical process (e.g., phase retrieval and refractive index reconstruction considered in \cite{DINER}). However, it may not hold generalization to unseen coordinates due to the discontinuity of hash key. To address this limitation, Zhu et al. \cite{MLP+InstantNGP} proposed the hybrid RHINO framework, which utilizes an additional coordinate MLP concatenated with the parametric grid input to regularize hash grid encoding-based methods, enhancing the generalization ability for methods such as DINER \cite{DINER} and InstantNGP \cite{InstantNGP}.  \par 
Grid encoding parametric models have achieved significant success for continuous representation. Future research could address the generalization limitations of current works by integrating advanced interpolation techniques (e.g., cubic splines), low-rank decomposed structures, or more robust hybrid architectures (e.g., the RHINO framework \cite{MLP+InstantNGP}) for broader applications such as operator learning \cite{NO} and full waveform inversion \cite{IFWI_JGR}. These improvements would further amplify the practical value of these grid encoding parametric models, paving the way for broader applications in science and engineering fields.
\subsection{Matrix and Tensor Function Decomposition}
Matrix and tensor function decomposition methods have been widely studied to incorporated structural constraints (such as low-rankness) into the continuous function representation \cite{CoordX,LRTFR,DRO-TFF}. Integrating function decomposition paradigms with INRs has also been an emerging type of structural modeling methods for continuous representation. The superiority of such structure lies in its computational efficiency brought by the compact low-rank factorization and coordinate decomposition, and the encoded low-rankness would benefit many data reconstruction problems. \par
Liang et al. \cite{CoordX} proposed the CoordX, which uses a split structure of MLP for INR. In this method, the initial layers of the MLP are split to learn each dimension of the input coordinates separately. The intermediate features are then fused by the last layers through outer product to generate the learned signal at the corresponding coordinate. It significantly reduces the computational costs and leads to speed up in training and inference, while achieving similar accuracy as the baseline INR. Luo et al. \cite{LRTFR} proposed the low-rank tensor function representation (LRTFR), which learns a low-rank Tucker decomposition parameterized by univariate INRs to generate the factor matrices of the tensor decomposition. It holds superior computational efficiency for grid-structured data and enjoys effectiveness for multi-dimensional data reconstruction problems. Specifically, given an input coordinate $\bf v$, the LRTFR is formulated as
\begin{align*}[\mathcal{C};f_x,f_y,f_z](\mathbf{v}):=\mathcal{C}\times_1 f_x(\mathbf{v}_{(1)})\times_2 f_y(\mathbf{v}_{(2)})\times_3 f_z(\mathbf{v}_{(3)}),\end{align*}
where $\cal C$ is the core tensor and $f_x,f_y,f_z$ are three univariate INRs that generate the factor matrices by taking each separate dimension of the coordinate as inputs. For a tensor of size $n_1\times n_2\times n_3$, the LRTFR costs $O(mrd(n_1+n_2+n_3)+rn_1n_2n_3)$ in a forward pass while a conventional MLP-based INR would cost $O(m^2dn_1n_2n_3)$, where $m$ denotes the INR width, $d$ denotes the INR depth, and $r$ denotes the Tucker rank. Since $r$ is usually much lower than $m^2d$, the LRTFR is practically more efficient than conventional INR methods. And the low-rankness inherent in such type of methods improve the robustness and convergence stability for continuous representation.\par 
\begin{table}[t]
	\centering
	\scriptsize
	\belowrulesep=0pt
	\aboverulesep=0pt
	\setlength{\tabcolsep}{4pt}
	\renewcommand{\arraystretch}{1.1} 
	\caption{Review of some representative structural modeling methods based on continuous representation.\label{tab:modeling}}
	\begin{tabular}{ll|l|c|c}
\midrule
\textbf{Category} & \textbf{Method} & \textbf{Description} & \textbf{Year} & \textbf{Ref.} \\
		\midrule
		\multirow{9}*{\tabincell{l}{Matrix or Tensor\\Decomposition}} & \tabincell{l}{{(CoordX)} Matrix \\decomposition} & \tabincell{l}{Introduce CoordX based on matrix decomposition \\ to accelerate conventional INRs.} & 2022 & \cite{CoordX} \\\cmidrule{2-5}
		& \tabincell{l}{{(LRTFR)} Tensor Tucker\\ decomposition} & \tabincell{l}{Low-rank tensor function representation based on\\ Tucker decomposition for multi-dimensional data.} & 2024 & \cite{LRTFR} \\\cmidrule{2-5}
		& \tabincell{l}{{(F-LRTF)} Tensor SVD} & \tabincell{l}{Learnable functional transform-based low-rank tensor \\factorization for continuous interpolation.} & 2024 & \cite{ECCV_Jianli} \\\cmidrule{2-5}
		& \tabincell{l}{{(DRO-TFF)} Rank-one\\ decomposition} & \tabincell{l}{Parameter-efficient deep rank-one tensor functional\\factorization for multi-dimensional data.} & 2025 & \cite{DRO-TFF} \\
		\cmidrule{2-5}
		& \tabincell{l}{{(F-INR)} Functional \\tensor decomposition} & \tabincell{l}{Functional Tucker, CP, and tensor-train decomposition\\-based INR for efficient continuous representation.} & 2025 & \cite{FTD-INR} \\
		\midrule
		\multirow{7}*{\tabincell{l}{Statistical \\Framework}} 
		& \tabincell{l}{{(SFTL)} Factor\\ trajectory learning} & \tabincell{l}{Streaming factor trajectory learning for tensor decomposition\\ by using Gaussian processes to model factor trajectory.} & 2023 & \cite{NeurIPS_Fang} \\\cmidrule{2-5}
		& \tabincell{l}{{(FunBaT)} Functional \\Bayesian decomposition} & \tabincell{l}{Functional Bayesian Tucker decomposition with Gaussian\\ processes for continuous-indexed tensor data.} & 2024 & \cite{Shikai_ICLR_24} \\\cmidrule{2-5}
		& \tabincell{l}{{(BayOTIDE)} Online \\functional decomposition}& \tabincell{l}{Functional Tucker decomposition with Gaussian processes \\ for online inference of time-series data imputation.} & 2024 & \cite{Shikai_ICML_24} \\\cmidrule{2-5}
		& \tabincell{l}{{(GRET)} Temporal tensor \\decomposition} & \tabincell{l}{Tensor decomposition with continuous spatial indexes and\\ temporal trajectories via neural ODEs.} & 2025 & \cite{Shikai_25_LRTFR} \\
		\midrule
		\multirow{9}*{\tabincell{l}{Continuous\\Regularization}} & \tabincell{l}{(IDIR) Jacobian \\regularization} & \tabincell{l}{Introduce Jacobian regularizer and hyperelastic regularizer \\based on INR for deformable registration.} & 2022 & \cite{INR_registration_Jacobian_reg} \\\cmidrule{2-5}
		& \tabincell{l}{(INRR) Regularize INR \\by itself} & \tabincell{l}{Learning the Dirichlet energy parameterized by another tiny \\INR to improve the generalization of the backbone INR.} & 2023 & \cite{INRR} \\\cmidrule{2-5}
		&\tabincell{l}{(CRNL) Nonlocal \\self-similarity}& \tabincell{l}{Continuous representation-based nonlocal method \\with coupled tensor factorization for data reconstruction.} & 2024 & \cite{CRNL} \\\cmidrule{2-5}
		& \tabincell{l}{(NeurTV) Neural total \\variation} & \tabincell{l}{Higher-order and directional total variation regularization\\ based on continuous neural representation.} & 2025 & \cite{NeurTV} \\
		\cmidrule{2-5}
		& \tabincell{l}{Isometric regularization\\for functional data} & \tabincell{l}{Isometric regularization to preserve geometric quantities\\ between latent space of INR and functional data manifold.} & 2025 & \cite{ICLR_geometric} \\
		\midrule
	\end{tabular}
\end{table}
Recently, Wang et al. \cite{ECCV_Jianli} proposed the functional transform-based low-rank tensor factorization (FLRTF), which parameterizes the mode-3 factor matrix of the tensor factorization with a functional transform parameterized by INR to enable arbitrary-resolution interpolation of multi-dimensional data, such as frame interpolation of videos and spectral super-resolution of multispectral images. For a third-order tensor $\cal X$, the FLRTF is formulated as
\begin{align*}
	\mathcal{X}(\cdot,\cdot,k) := (\mathcal{A} \bigtriangleup \mathcal{B}) \times_3 f(\mathbf{z}_{(k)}), \quad k = 1, 2, \cdots, n_3,
\end{align*}
where $\mathcal{A},\mathcal{B}$ are two tensor factors, $\bigtriangleup$ is the face-wise product between two tensors, and $\times_3$ is the mode-3 tensor-matrix product. Continuous data reconstruction tasks such as video frame interpolation/extrapolation and multi-spectral image band interpolation demonstrate FLRTF's superior performance brought by the functional transform tensor factorization. More recently, Li et al. \cite{DRO-TFF} proposed the deep rank-one tensor functional factorization (DRO-TFF). For a third-order tensor $\cal X$, the DRO-TFF model is formulated as
\[
\mathcal{X} = \psi\bigl(\cdots\psi\bigl((f_{\theta_x}(\mathbf{v}_1)\triangle f_{\theta_y}(\mathbf{v}_2)) \times_3 \mathbf{H}_1\bigr) \times_3 \cdots \times_3 \mathbf{H}_{k-1}\bigr) \times_3 f_{\theta_z}(\mathbf{v}_3),
\]
where $f_{\theta_x}(\mathbf{v}_1),f_{\theta_y}(\mathbf{v}_2)$ are rank-one factors generated by univariate INRs, ${\bf H}$ are weight matrices of the mode-3 deep transform, $f_{\theta_z}(\mathbf{v}_3)$ is the mode-3 functional transform, and $\psi(\cdot)$ is a nonlinear activation. The DRO-TFF excels in lightweight models by virtue of the lightweight rank-one factorization, and thus holds better efficiency among tensor functional factorization methods. \par
Furthermore, Vemuri et al. \cite{FTD-INR} proposed the functional tensor decomposition-based INR, which employs TT, Tucker, and CP tensor decompositions to reconstruct INR. The method reduces forward pass complexity while improves accuracy through specialized learning. The authors applied its to image compression, physics simulations, and 3D geometry reconstruction and shows superior efficiency. Nie et al. \cite{Traffic_LRTFR} proposed the low-rank tensor function representation as a generalized traffic data learner, which enables seamless completion of traffic analysis tasks such as state estimation and mesh-based flow estimation \cite{Traffic_LRTFR}.\par 
Matrix or tensor functional decomposition methods significantly improve the efficiency of INR methods, while the encoded structure constraints such as low-rankness benefit downstream applications. In future research, the combination of low-rank function decomposed INRs with operator learning paradigms \cite{NO} and arbitrary-scale imaging problems \cite{LIIF} are promising directions by leveraging the efficiency of these function decomposition methods. Applying the function decomposition to fine tuning large models are also interesting. It is also interesting to explore more expressive tensor decompositions, such as tensor rings
or topology-aware tensor networks, to enhance INRs. Exploring adaptive rank selection in the low-rank function decomposition (such as K-means and PCA-based approaches) is also interesting.
\subsection{Statistical and Bayesian Framework}
Recently, several continuous representation methods based on statistical and Bayes frameworks are established. These method model continuous trajectories (such as temporal trajectory) by imposing statistical distribution assumptions, and then inferring the model parameters from the given observed discrete signals. These methods are capable of capturing long-term dependency and periodic patterns underlying data, making the structural modeling more accurate. The core concept of statistical and Bayesian modeling is to approximate the exact posterior $p({\cal U},\theta|{\cal X})$ using a variational distribution $q({\cal U},\theta)$, where ${\cal X}$ denotes the observed tensor data, ${\cal U}$ denotes the latent variables, and $\theta$ denotes the statistical parameters to be estimated. The model parameters are inferred by efficient variational inference algorithms, such as variational expectation and maximization or Kalman filtering \cite{Shikai_ICML_24}.\par
To be specific, Fang et al. \cite{NeurIPS_Fang} proposed to model streaming tensors by factor trajectory learning parameterized by Gaussian process (GP), so as to flexibly estimate the continuous temporal evolution of factors for online inference of tensor streams. Their later work \cite{Shikai_ICLR_24} introduced the functional Bayesian Tucker decomposition, which models continuous-indexed tensor data by Tucker decomposition parameterized by GP. The GP is used as functional priors to model the latent functions of Tucker decomposition, formulated as:
\begin{equation}
\begin{split}
&f(i_1, \ldots, i_K) \approx \operatorname{vec}(\mathcal{W})^T \left( U^1(i_1) \otimes \ldots \otimes U^K(i_K) \right)\\
&U^k(i_k) = [u_1^k(i_k), \ldots, u_{r_k}^k(i_k)]^T; \quad u_j^k(i_k) \sim \mathcal{GP}(0, \kappa(i_k, i_k')), \quad j = 1, \ldots, r_k. 
\end{split}
\end{equation}
Here, $\cal W$ denotes the core tensor, $U^k(i_k)$ denotes the factor matrix, $\mathcal{GP}(\cdot)$ denotes the Gaussian process, and $\kappa(i_k, i_k')$ denotes the Matérn kernel function. To infer the model parameters from an observed tensor data $\cal D$, the authors transformed the factor GP into state variables $\mathbf{Z}^{k}$ under the Markov chain, and further assign a Gamma prior over the noise precision $\tau$ and a Gaussian prior over the Tucker core $\cal W$. The joint probabilities are approximated into fully factorized format under the mean-field assumption: 
\[
p(\mathbf{\Theta}|\mathcal{D}) \approx q(\mathbf{\Theta}) = q(\tau) q(\mathcal{W}) \prod\nolimits_{k=1}^{K} q(\mathbf{Z}^{k}),
\]
and then optimize the approximated posterior through expectation propagation \cite{Shikai_ICLR_24}. The algorithm is evaluated on both synthetic and real-world tensor data, and is capable of modeling continuous-indexed tensor data, especially capturing well-documented period patterns underlying tensor data by virtue of the statistical modeling.\par 
Fang et al. \cite{Shikai_ICML_24} further proposed the Bayesian online multivariate time-series imputation method with continuous functional decomposition. Compared with \cite{Shikai_ICLR_24}, the new online method is capable of dealing with time-series data that evolves within time. They modeled the multivariate time-series as a temporal vector-valued function $\mathbf{X}(t)$ consisting of the weighted combination of trend and seasonality factors over time. It is formulated as
\[ 
\mathbf{X}(t) = \mathbf{U}\mathbf{V}(t) = [\mathbf{U}_{\text{trend}}, \mathbf{U}_{\text{season}}] 
\begin{bmatrix} 
	\mathbf{v}_{\text{trend}}(t) \\ 
	\mathbf{v}_{\text{season}}(t) 
\end{bmatrix},\;\mathbf{v}_{\text{trend}}^{i}(t) \sim \mathcal{GP}\left(0, \kappa_{\text{Matern}}\right),\;
\mathbf{v}_{\text{season}}^{j}(t) \sim \mathcal{GP}\left(0, \kappa_{\text{periodic}}\right),
\]
where $\mathbf{U}_{\text{trend}}, \mathbf{U}_{\text{season}}$ are combination weights and the trend and seasonality factors $\mathbf{v}_{\text{trend}}(t),
\mathbf{v}_{\text{season}}(t)$ are modeled by GPs with different kinds of kernels to exploit temporal patterns (i.e., long-term patterns and periodic patterns). Similar to the inference technique considered in \cite{Shikai_ICLR_24}, the inference of GP can be converted to the solution of a time-invariant stochastic differential equation, which is further discretized as a Markov model to infer the state variable ${\bf Z}(t)$ corresponding to the factor ${\bf v}(t)$. Given all observations up to time $t_n$, i.e., ${\cal D}_{t_n}$, and a new observation arrived at $t_{n+1}$, i.e., $\mathbf{y}_{n+1}$, the statistical model parameters $\Theta:=\{{\bf U},{\bf Z}(t),\tau\}$ are inferred by maximizing the online posterior under the Bayes' rule:
\begin{align*}
&p(\Theta \mid D_{t_n} \cup \mathbf{y}_{n+1}) \propto p(\mathbf{y}_{n+1} \mid \Theta, {\cal D}_{t_n}) p(\Theta \mid {\cal D}_{t_n}),\\
&p(\Theta \mid {\cal D}_{t_n}) \approx q(\tau \mid {\cal D}_{t_n}) \prod_{d=1}^{D} q(\mathbf{u}^d \mid {\cal D}_{t_n}) q(\mathbf{Z}(t) \mid {\cal D}_{t_n}),
\end{align*}
where the second approximate posterior is formulated by the mean-field factorization. The authors developed a novel approach to update the posterior $p(\Theta \mid {\cal D}_{t_n})$ in an online manner by using Kalman filter and message merging techniques \cite{Shikai_ICML_24}. The method can handle imputation over arbitrary timestamps in the time-series data, and also offers uncertainty quantification for time-series from the statistical modeling, particularly benefiting downstream applications.\par
More recently, Chen et al. \cite{Shikai_25_LRTFR} proposed an innovative generalized temporal tensor decomposition with rank-revealing latent model for real-world tensor data reconstruction, which integrates statistical modeling with neural representations. This method encodes continuous spatial indexes through an MLP with Fourier feature encoding under the CP tensor decomposition, and employs neural ODEs in latent space to learn the temporal trajectories of factors. Specifically, given a spatial index $i_k$ in the dimension $k$, the temporal factor ${\bf g}^k$ of the CP functional decomposition is modeled by
\[
\begin{aligned}
	\mathbf{z}^k(i_k, 0) &= \text{Encoder}\big([\cos(2\pi\mathbf{b}_k i_k); \sin(2\pi\mathbf{b}_k i_k)]\big), \\
	\mathbf{z}^k(i_k, t) &= \mathbf{z}^k(i_k, 0) + \int_0^t h_{\boldsymbol{\theta}_k}(\mathbf{z}^k(i_k, s), s) \, ds, 
	\;\mathbf{g}^k(i_k, t) = \text{Decoder}(\mathbf{z}^k(i_k, t)),
\end{aligned}
\]
where ${\bf z}^k$ are latent temporal factors and $h_{\boldsymbol{\theta}_k}$ is a state transition function of the dynamics at each timestamp, which is parameterized by an MLP. The index $i_k$ is firstly encoded through an encoder MLP with Fourier encoding, and the neural ODE is implemented to obtain the latent temporal factor through $\mathbf{z}^k(i_k, t)={\rm ODESolve}(\mathbf{z}^k(i_k, 0),h_{\boldsymbol{\theta}_k})$. Finally, the temporal factor ${\bf g}^k$ is obtained by a decoder MLP. A Gaussian-Gamma prior is further imposed to the factor trajectory to automatically reveal the rank of the temporal tensor. The statistical model is also solved by variational inference w.r.t. the approximated posterior. The method outperforms existing statistical methods for tensor recovery and prediction by virtue of the neural representation coupled with statistical modeling \cite{Shikai_25_LRTFR}.\par
Statistical and Bayesian continuous representation methods are capable of capturing periodic and long-term patterns that benefit time-series and streaming data analysis. Future research on enhancing the expressiveness of statistical frameworks (particularly integrating these frameworks with more powerful neural representations \cite{Shikai_25_LRTFR}) is expected to take advantage of both types of methods to better characterize both irregular local variability and long-term dependency underlying signals.  
\subsection{Continuous Regularization for Structural Modeling}
Except for the construction of various continuous representation models, there are also a group of works that focus on the developments of explicit continuous regularization methods conditioned on representations, mainly derived from intrinsic priors of data such as local correlations and nonlocal self-similarity. The explicit constraints lead to more accurate structural modeling of data based on the continuous representation. These continuous regularization methods improve the generalization ability, robustness, and applicability of continuous models. Moreover, continuous regularizations benefit from the resolution-independence and flexibility of continuous representations to handle complex and irregular scenarios that traditional regularization methods (such as NMF and total variation) fail to implement. \par 
As a representative example, Wolterink et al. \cite{INR_registration_Jacobian_reg} proposed the INR for deformable image registration, where several regularization terms were introduced to alleviate the ill-posedness of such inverse problem. The Jacobian regularizer is formulated as 
\[S^{jac}[\Phi] = \int_{\Omega} |1 - \det \nabla \Phi| \, dx,\]
where $\Phi$ is the INR and this regularizer limits the Jacobian determinant around $1$ to restrict the local shrinkage or expansion of the deformation. The hyperelastic regularizer \cite{INR_registration_Jacobian_reg} conditioned on the INR $\Phi$ is defined as
\[S^{\text{hyper}}[\Phi] = \int_{\Omega} \left[ \frac{1}{2}\alpha_{l}|\nabla u|^{2} + \alpha_{a}\phi_{c}(\text{cof}\,\nabla\Phi) + \alpha_{v}\psi(\text{det}\,\nabla\Phi) \right] dx,\]
where $\frac{1}{2}\alpha_{l}|\nabla u|^{2}$ penalizes the length of the deformation, $\alpha_{a}\phi_{c}(\text{cof}\,\nabla\Phi)$ penalizes the expansion of area with the cofactor matrix of the Jacobian $\text{cof}\,\nabla\Phi$, and $\alpha_{v}\psi(\text{det}\,\nabla\Phi)$ penalizes the growth and shrinkage of the deformation with a convex function $\psi(v)=\frac{(v-1)^4}{v^2}$. Expect for the first-order regularizations, INR can also be facilitated with higher-order regularizations due to the differentiability of activation functions such as SIREN \cite{SIREN}. The second-order Bending energy regularization \cite{INR_registration_Jacobian_reg} is formulated as
\[
\mathcal{S}^{\text{bend}}[\Phi] = \int_{-1}^{1} \int_{-1}^{1} \int_{-1}^{1} 
\left( \frac{\partial^2 \Phi}{\partial x^2} \right)^2 + 
\left( \frac{\partial^2 \Phi}{\partial y^2} \right)^2 + 
\left( \frac{\partial^2 \Phi}{\partial z^2} \right)^2 + 
2 \left( \frac{\partial^2 \Phi}{\partial x \partial y} \right)^2 + 
2 \left( \frac{\partial^2 \Phi}{\partial x \partial z} \right)^2 + 
2 \left( \frac{\partial^2 \Phi}{\partial y \partial z} \right)^2 dx  dy  dz,
\]
which penalizes the smoothness of the deformation vector field by using second-order derivatives of INR. These regularization methods can be naturally combined into INRs based on standard automatic differentiation techniques, leading improved accuracy for deformable image registration.\par
Recently, Li et al. \cite{INRR} proposed the Dirichlet energy regularization for INR by utilizing another tiny INR to parameterize the Laplacian matrix of Dirichlet energy, termed INR regularization (INRR). Given an output matrix ${\bf X}$ obtained by an INR with parameters $\theta$, the INRR is formulated as 
\[
\begin{cases} 
	{\cal R}(\theta) = \text{tr} \left({\bf X}^T {\bf L}(\theta) {\bf X}\right) \\ 
	{\bf L}(\theta) = {\bf A}(\theta) \cdot \mathbf{1}_{m' \times m'} \odot {\bf I}_{m'} - {\bf A}(\theta) \\ 
	{\bf A}(\theta) = \frac{\exp \left( g^T (\theta; u) g(\theta; u) \right)}{\mathbf{1}_{m'}^T \exp \left( g^T (\theta; u) g(\theta; u) \right) \mathbf{1}_{m'}}, 
\end{cases}
\] 
where ${\cal R}(\theta)$ is the Dirichlet energy that captures the row/column-wise nonlocal self-similarity of $\bf X$, ${\bf L}(\theta)$ is the Laplacian matrix that constitutes the Dirichlet energy, ${\bf A}(\theta)$ is a weighted adjacency matrix that measures the similarity of $\bf X$, and $g(\theta; u)$ is another tiny INR with input row/column coordinates $u$. The tiny INR here enforces the implicit smoothness constraint for the Laplacian matrix ${\bf L}(\theta)$, which aligns with the piecewise smoothness of natural images. The INRR is especially helpful for improving the generalization abilities of INR for nonuniform sampling coordinates by leveraging the smooth row/column-wise self-similarity underlying signals. Similarly, Luo et al. \cite{CRNL} proposed the continuous representation-based nonlocal method for multi-dimensional data, which captures the nonlocal self-similarity of discrete grid or irregular signals by constructing nonlocal patch groups. Especially, the nonlocal continuous similar cubes of a signal are grouped together, which holds stronger low-rankness than the original data. Then, a coupled low-rank tensor function representation method was proposed to represent the nonlocal continuous cubes to reconstruct the data \cite{CRNL}. The nonlocal method enjoys superior reconstruction abilities for both grid data such as images and irregular non-grid ones such as point clouds by virtue of the nonlocal low-rankness excavation, and holds better efficiency by the coupled tensor factorization.\par 
\begin{figure}[t]
	\scriptsize
	\setlength{\tabcolsep}{0.9pt}
	\begin{center}
		\begin{tabular}{c}
			\centering
			\includegraphics[width=0.8\textwidth]{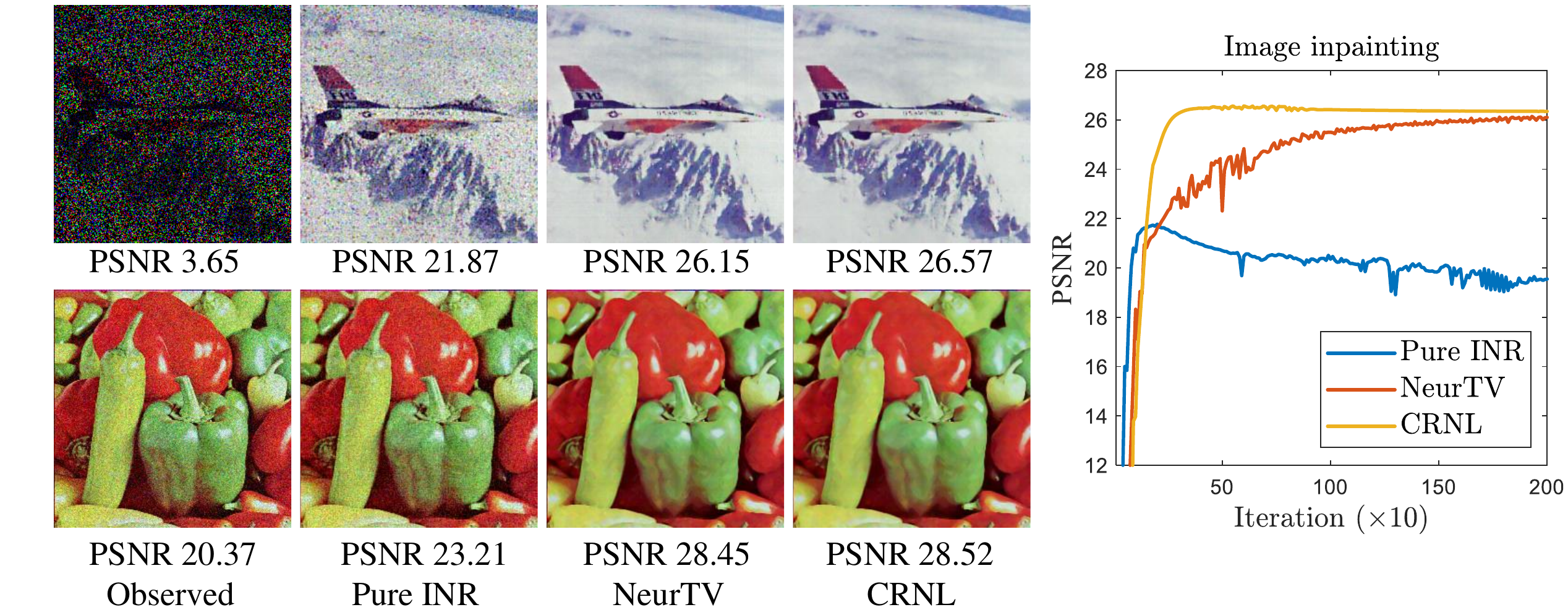}
		\end{tabular}
	\end{center}
	\vspace{-0.4cm}
	\caption{Continuous regularization methods such as NeurTV \cite{NeurTV} and continuous representation-based nonlocal model (CRNL) \cite{CRNL} alleviate the overfitting phenomenon of pure INR for image reconstruction tasks including image inpainting (top row) and denoising (bottom row) by incorporating intrinsic priors of data.}
	\label{fig:reg}
\end{figure}
Recently, Luo et al. \cite{NeurTV} proposed the total variation regularization on the neural domain (termed NeurTV). It uses the directional or higher-order derivatives of INR outputs w.r.t. input coordinates to capture local directional correlations intrinsically existed in grid and non-grid data, and NeurTV demonstrates superior capabilities then traditional discrete TV for image, point cloud, and spatial transcriptomics data reconstruction by avoiding the discretization error inherent in traditional methods. Specifically, the directional NeurTV is defined as
\[
\Psi_{\text{NeurDTV}_{\theta}}(\Theta) := \int_\Omega\left| \frac{\partial f_\Theta(\mathbf{x})}{\partial \mathbf{x}_{(1)}} \cos \theta + \frac{\partial f_\Theta(\mathbf{x})}{\partial \mathbf{x}_{(2)}} \sin \theta \right|d{\bf x},
\]
where $\theta$ is a predefined direction and $f_\Theta$ is an INR with input coordinate $\bf x$. The directional NeurTV can capture local smoothness along a predefined direction $\theta$. Furthermore, the NeurTV can be extended to a space-variant formulation that automatically determines the local directional and scale parameters for each spatial point to better describe local variations:
	\begin{equation}
	\Psi_{{\rm NeurTV}_\theta^\alpha}(\Theta)=\int_\Omega \alpha_{\bf x}\left\lVert\begin{pmatrix}
		{a_{\bf x}}&0\\
		0&2-a_{\bf x}\\
	\end{pmatrix}
	\begin{pmatrix}
		\cos\theta_{\bf x}&\sin\theta_{\bf x}\\
		\sin\theta_{\bf x}&-\cos\theta_{\bf x}\\
	\end{pmatrix}
	\begin{pmatrix}
		\frac{\partial f_\Theta({\bf x})}{\partial{\bf x}_{(1)}}\\
		\frac{\partial f_\Theta({\bf x})}{\partial{\bf x}_{(2)}}
	\end{pmatrix}
	\right\rVert_{\ell_1}d{\bf x},
\end{equation}
where $\alpha_{\bf x},a_{\bf x}$ are local scale parameters for $\bf x$ and $\theta_{\bf x}$ are local directional parameters. The space-variant NeurTV better captures directional variations within a signal, and the directional parameters $\alpha_{\bf x},a_{\bf x},\theta_{\bf x}$ are automatically identified and calibrated during optimization. Some numerical examples of the nonlocal method \cite{CRNL} and the space-variant NeurTV \cite{NeurTV} regularization are shown in {Fig. \ref{fig:reg}} to visually demonstrate the efficacy of these regularization methods for general data reconstruction tasks by merging such intrinsic data priors with continuous representation.\par 
Recently, Heo et al. \cite{ICLR_geometric} proposed an isometric regularization technique for INRs with latent space embedding, such as the generative modeling of shapes using signed distance function \cite{DeepSDF}. The regularization preserves meaningful geometric quantities (such as distances and angles) between the latent space and the functional data manifold by interpreting the mapping from latent variables to INRs as a parametrization of a Riemannian manifold. The regularization encourages the manifold to hold minimal intrinsic curvature, hence enabling robust and smooth geometric representations. One can generalize this regularization to a group of latent space-based continuous representations, such as operator learning \cite{NO,ICLR_geometric} and arbitrary-scale image representation \cite{LIIF}. \par
In future research, it would be important to scale these effective continuous regularization to higher-dimensional cases (e.g., extending the directional NeurTV to higher dimensions beyond two), which would benefit multi-directional characterization such as optical flow in videos and spectral smoothness in multispectral images. Meanwhile, it would be interesting to apply these explicit regularizations to more vision and scientific computing problems such as scene representation \cite{NeRF}, surface representation \cite{DeepSDF}, and neural physical simulations \cite{AAAI_PINN_InstantNGP} to enhance their respect robustness. 
\section{Theoretical Foundations of Continuous Regularizations}\label{sec:theory}
This section introduces several theoretical frameworks that uncover fundamental insights into continuous representation learning methodologies. Our analysis is primarily based on three groups of theories: approximation error analysis and representational capacity theory for continuous representation models, convergence and generalization analysis of continuous methods from the neural tangent kernel theory, and implicit regularizations brought by the structures of continuous models and their corresponding optimization dynamics. These theoretical perspectives are summarized in Table \ref{tab:theory}.
\begin{table}[t]
	\centering
	\scriptsize
	\belowrulesep=0pt
	\aboverulesep=0pt
	\setlength{\tabcolsep}{2pt}
	\renewcommand{\arraystretch}{1.1} 
	\caption{Review of representative theoretical foundations related to continuous function representations. \label{tab:theory}}
	\begin{tabular}{ll|l|c|c}
		\midrule
		\textbf{Category} & \textbf{Theory} & \textbf{Description} & \textbf{Year} & \textbf{Ref.} \\
		\midrule
		\multirow{11}*{\tabincell{l}{Approximation \\and Representation}} 
		& \tabincell{l}{Truncation error of SVD\\ for functions} & \tabincell{l}{Truncation error estimate and decay rate of SVD \\approximation for Sobolev smooth bivariate functions.} & 2018 & {\cite{SIAM_Bivariate_function_SVD}}\\
		\cmidrule{2-5}
		& \tabincell{l}{Functional tensor-train\\decomposition}&\tabincell{l}{Propose functional tensor-train (TT) (and parameterization\\ error) that represents functions in the TT format.} & 2019 & {\cite{FunTT}}\\	
		\cmidrule{2-5}
		& \tabincell{l}{Functional tensor CP\\ decomposition }&\tabincell{l}{Propose the functional CP decomposition (and its contractive\\ error bounds) via RKHS-based power iteration.} & 2021 & {\cite{JASA_FunTSVD}}\\
		\cmidrule{2-5}
		~& \tabincell{l}{Dictionary analysis \\for INR} & \tabincell{l}{Utilize harmonic analysis to characterize the function class \\represented by INR using a signal dictionary.} & 2021 & {\cite{Dictionary_INR}}\\
		\cmidrule{2-5}
		& \tabincell{l}{Complex wavelets analysis\\ for INR}  & \tabincell{l}{Characterize the function class represented by wavelet INR \\using Fourier convolution theorem and $\Gamma$-progressiveness. } & 2024 & {\cite{WIRE_theory}}\\
		\cmidrule{2-5}
		~& \tabincell{l}{Functional tensor SVD\\ framework}  & \tabincell{l}{Propose the functional tensor SVD theorem in continuous \\domains for multi-output regression problem.} & 2024 & {\cite{NeurIPS_FtSVD}}\\
		\midrule
		\multirow{11}*{\tabincell{l}{Convergence\\and Generalization}}  &\tabincell{l}{Neural tangent kernel\\(NTK)}& \tabincell{l}{The training of neural nets follows kernel gradient w.r.t. NTK,\\ enabling generalization and convergence analysis of neural nets.} & 2018 & \cite{NTK} \\\cmidrule{2-5}
		&\tabincell{l}{Convolutional NTK}&\tabincell{l}{Propose the NTK computation of CNN and the non-asymptotic\\ proof that wide net is equivalent to NTK regression.} & 2019 & \cite{wideNN_NTK}\\\cmidrule{2-5}
		&\tabincell{l}{Regularized neural nets\\outperforms NTK}&\tabincell{l}{Regularized neural nets has better generalization than NTK \\and the convergence of $\ell_2$-norm regularized neural nets.} & 2019 & \cite{l2_reg_NTK}\\\cmidrule{2-5}
		&\tabincell{l}{RKHS norm reg.\\for neural nets}&\tabincell{l}{Use the RKHS norm under the NTK as the function-space\\ regularization of deep neural nets.} & 2022 & \cite{RKHS_norm_reg}\\\cmidrule{2-5}
		&\tabincell{l}{Modified kernel{\textquoteright}s\\ spectrum}&\tabincell{l}{Propose modified spectrum kernels and a preconditioned \\method for faster convergence by altering eigenvalues of NTK.} & 2024 & \cite{TMLR_NTK_control} \\\cmidrule{2-5}
		&\tabincell{l}{NTK analysis for \\InstantNGP}&\tabincell{l}{Demonstrate that InstantNGP recovers fine details better\\ from the NTK eigenvalue perspective.} & 2025 & \cite{InstantNGP_NTK} \\
		\midrule
		\multirow{9}*{\tabincell{l}{Implicit\\Regularization}}& \tabincell{l}{Implicit reg. in deep matrix\\ factorization} & \tabincell{l}{Study the implicit nuclear norm reg. of gradient descent\\ over deep linear neural networks for matrix sensing.} & 2019 & \cite{Implicit_regularization_NIPS_2019} \\\cmidrule{2-5}
		& \tabincell{l}{Implicit reg. in deep tensor\\ factorization and CNN} & \tabincell{l}{Analyze the implicit reg. of hierarchical tensor factorization \\(equivalent to deep CNN) towards low hierarchical tensor rank.} & 2022 & \cite{Implicit_reg_CNN} \\\cmidrule{2-5}
		& \tabincell{l}{Implicit reg. in deep CP\\ factorization} & \tabincell{l}{Establish the polynomial growth of implicit low-rank reg. \\w.r.t. the depth of deep tensor CP factorization.} & 2022 & \cite{Implicit_regularization_deep_CP} \\\cmidrule{2-5}
		& \tabincell{l}{Implicit reg. in deep Tucker\\ factorization} & \tabincell{l}{Study the implicit reg. of deep Tucker factorization for tensor\\ completion towards solutions with low multilinear rank.} & 2024 & \cite{Implicit_regularization_deep_Tucker} \\\cmidrule{2-5}
		& \tabincell{l}{Implicit bias of AdamW\\optimizer} & \tabincell{l}{Show that AdamW optimizer converges to a KKT point of the\\ original loss under bounded $\ell_\infty$-norm constraint.} & 2024 & \cite{AdamW_implicit} \\
		\midrule
	\end{tabular}
\end{table}
\subsection{Approximation and Representation Theory}
The approximation and representation theory refer to the theoretical framework that quantizes the error of estimating/representing an unknown or complex function using a more tractable continuous function representation by parametric or non-parametric models \cite{JASA_FunTSVD,FunTT,SIAM_Bivariate_function_SVD,SIAM_cheb}. The goal is to find or demonstrate that a function from a predefined class (e.g., polynomials, neural networks, Gaussians) could closely match the target function in the continuous domain.\par 
First, it is known that deep neural nets serve as universal approximators for any continuous functions in the compact set of Euclidean space.
\begin{theorem}[Universal approximation \cite{Universal_NN,Universal_Ellacott_1994}] 
Let \( K \subset \mathbb{R}^n \) be a compact set, and \( f: K \to \mathbb{R} \) be any continuous function. For any \( \epsilon > 0 \), there exists a single-hidden-layer MLP with activation function \( \sigma \) (any non-polynomial continuous function), such that the network output \( N(x) \) satisfies \[
	\sup_{x \in K} |f(x) - N(x)| < \epsilon,\;
	N(x) = \sum_{i=1}^m a_i \, \sigma(w_i \cdot x + b_i),\]
	where \( m \in \mathbb{N} \), \( a_i, b_i \in \mathbb{R} \), and \( w_i \in \mathbb{R}^n \).  
\end{theorem}
The theorem provides a fundamental theoretical guarantee on the representation abilities of INR. For example, the widely-used SIREN \cite{SIREN} utilizes the sinusoidal activation function, which is a non-polynomial continuous function. Hence, SIREN has the capacity to approximate any continuous functions. However, in practice, INRs such as SIREN still suffer from spectral basis that favors low-frequency information and leads to slower convergence to high-frequency information. This is related to the optimization dynamic and can be well explained by the neural tangent kernel; see Section \ref{sec:NTK}.\par 
From the traditional function approximation perspective, many studies have explored the representation capacity of basis function representations. For example, any integrable periodic function can be represented by a unique combination of Fourier series \cite{Numerical_Fourier_Analysis}. Any compactly supported non-periodic function can be transformed into a periodic function, and hence can be also represented by Fourier series. Similarly, linear combinations of complete wavelets bases (such as Haar wavelets) are capable of approximating any integrable functions from the multiresolution analysis \cite{MRA}. \par  
Following the function representation theory pipeline, several works have investigated the function class that could be captured by general INR structures. Yuce et al. \cite{Dictionary_INR} proposed a theoretical framework that characterizes the function class represented by an INR. Specifically, INR families are structured signal dictionaries with atoms being integer harmonic functions related to the initial mapping frequencies. The INR $f_\theta$ in \cite{Dictionary_INR} is formulated as
\begin{align}
f_{\theta}({\bf r}) = {\bf W}^{(L)} {\bf z}^{(L-1)} + {\bf b}^{(L)},\;
	{\bf z}^{(\ell)} = \rho^{(\ell)} \left( {\bf W}^{(\ell)} {\bf z}^{(\ell-1)} + {\bf b}^{(\ell)} \right), \; \ell = 1, \ldots, L - 1,\;	{\bf z}^{(0)} = \gamma({\bf r}), 
\end{align}
where ${\bf r}$ is the input coordinate, $\gamma$ is the first-layer encoding, and $\rho$ is the activation. Such a structure can only capture the following class of functions.
\begin{theorem}[Function class represented by INR\cite{Dictionary_INR}]\label{th_INR_dic}
	Let \(\rho^{(\ell)}({\bf z}) = \sum_{k=0}^K \alpha_k {\bf z}^k\) for \(\ell > 1\). Let \(\Omega = [\Omega_0, \ldots, \Omega_{T-1}]^\top\) and \(\phi\) denote the matrix of frequencies and vector of phases in the encoding layer \(\gamma({\bf r}) = \sin(\Omega {\bf r} + \phi)\). Then the architecture $f_{\theta}({\bf r})$ can only represent functions of the form
	\begin{equation}
		f_\theta({\bf r}) = \sum_{\omega' \in \mathcal{H}(\Omega)} c_{\omega'} \sin{(\langle \omega', {\bf r} \rangle + \phi_{\omega'})},
	\end{equation}
	where $
		\mathcal{H}(\Omega) \subseteq \left\{ \sum_{t=0}^{T-1} s_t \Omega_t \middle| s_t \in \mathbb{Z} \land \sum_{t=0}^{T-1} |s_t| \leq K^{L-1} \right\}.$
\end{theorem}
The theorem shows that the expressive power of conventional INRs (such as PE-based INR \cite{PE} and SIREN \cite{SIREN}) is restricted to functions that can be expressed as a linear combination of certain harmonics related to the feature mapping $\gamma({\bf r})$. Hence, the INR has the same expressive power as a structured signal dictionary whose atoms are sinusoids with frequencies $\omega'$ equal to sums and differences of the integer harmonics of the mapping frequencies $\Omega$ \cite{Dictionary_INR}. Another insight is that the INR represents this function class with degree of freedom $O(TK^L)$ using substantially fewer parameters (i.e., with $O(T^2L)$ parameters in the learnable weights ${\bf W}$), indicating that the INR imposes a certain low-rank structure over the coefficients. Furthermore, the authors \cite{Dictionary_INR} drew inspirations from meta-learning to construct dictionary atoms of INRs as a combination of examples seen during meta-training. As a result, the target signals would project onto the eigenfunctions
of the neural tangent kernel with the largest eigenvalues, leading to increased convergence speed by meta-learning to reshape the dictionary atoms.\par
Recently, Roddenberry et al. \cite{WIRE_theory} proposed a wavelet analysis framework that captures the function class represented by wavelet INR \cite{WIRE}. The analysis is based on the Fourier convolution theorem of the atoms of the first-layer feature of the INR. Specifically, the signal is represented from coarse approximations performed in the first layer of the INR. The wavelet INR in \cite{WIRE_theory} is formulated as
\begin{align*}
	f_\theta(\mathbf{r}) = \mathbf{W}^{(L)}\mathbf{z}^{(L-1)}(\mathbf{r}) + \mathbf{b}^{(L)},\;
	\mathbf{z}^{(\ell)}(\mathbf{r}) = \rho^{(\ell)}(\mathbf{W}^{(\ell)}\mathbf{z}^{(\ell-1)}(\mathbf{r}) + \mathbf{b}^{(\ell)}),\;
	\mathbf{z}^{(0)}(\mathbf{r}) = \psi(\mathbf{W}^{(0)}\mathbf{r} + \mathbf{b}^{(0)}),
\end{align*}
where $\rho$ are polynomials and $\psi$ is the so-called template function in the first layer, considered as complex wavelets in \cite{WIRE_theory}. This class of INR has a Fourier transform that is fully determined by convolutions of the Fourier transforms of the atoms in the first layer, as demonstrated in the representation theorem.
\begin{theorem}[Function class represented by INR using template function convolutions \cite{WIRE_theory}]\label{thm:inr}Let a point $\mathbf{r}_0\in\mathbb{R}^d$ be given and $\Delta=\{\boldsymbol{m}|\sum_{t=1}^{F_1}m_t=k\}$. Under mild assumptions, there exists an open neighborhood $U\ni\mathbf{r}_0$ such that for all $\phi\in\mathcal{C}_0^\infty(U)$
	\begin{equation}\label{eq:fourier-rep}
		\widehat{\phi\cdot f_{\boldsymbol{\theta}}}(\xi) 
		= 
		\left( \widehat{\phi} \ast \sum_{k=0}^{K^{L-1}} \sum_{\boldsymbol{m}\in\Delta} \widehat{\beta}_{\boldsymbol{m}} {\mathop{\huge \mathlarger{{\ast}}}}_{t=1}^{F_1} \left( e^{i2\pi\langle\mathbf{W}_t^{-\top}\xi,\mathbf{b}_t\rangle} \widehat{\psi}(\mathbf{W}_t^{-\top}\xi) \right)^{\ast m_t, \xi} \right) (\xi),
	\end{equation}
	for coefficients $\widehat{\beta}_{\boldsymbol{m}}\in\mathbb{C}$ independent of $\mathbf{r}$, where $(\cdot)^{\ast m, \xi}$ denotes $m$-fold convolution of the argument with itself with respect to $\xi$.
	Furthermore, the coefficients $\widehat{\beta}_{\boldsymbol{m}}$ are only nonzero when each $t\in[1,\ldots,F_1]$ such that $m_t\neq 0$ also satisfies $\mathbf{W}_t\mathbf{r}_0+\mathbf{b}_t\in\supp(\psi)$.
\end{theorem}
This theorem is an application of the Fourier convolution theorem conditioned on the INR function class introduced in Theorem \ref{th_INR_dic}. Specifically, the Fourier transform $\widehat{\phi\cdot f_{\boldsymbol{\theta}}}$ is fully determined by the self-convolutions of the Fourier transforms of the atoms in the first layer, which generates integer harmonics by scaled, shifted copies of the template function $\psi$, i.e., $\psi(\mathbf{W}_t\mathbf{r}_0+\mathbf{b}_t)$. The support of these scaled and shifted atoms of the first layer, say $\psi(\mathbf{W}_t\mathbf{r}_0+\mathbf{b}_t)$, is preserved as the support of $\psi$, so that the INR output at a given coordinate $\bf r$ is dependent only on the atoms in the first layer whose support contains $\bf r$ \cite{WIRE_theory}. Moreover, the authors \cite{WIRE_theory} used the $\Gamma$-progressiveness analysis tool to show the advantages of using complex wavelets for INR. The $\Gamma$-progressiveness refers to the property of a function that the support of its Fourier transform lies in a convex conic set of ${\mathbb R}^d$. They demonstrated that the wavelet INR is a $\Gamma$-progressive function. Thus, any advantages/limitations of approximating functions using $\Gamma$-progressive template functions are maintained. And if the atoms in the first layer of an INR using a template function $\psi$ have vanishing Fourier transform in some neighborhood of the origin, then the output of the INR has Fourier support that also vanishes in that neighborhood (i.e., band-pass property). Hence, the authors designed a split INR structure that utilizes two INRs to respectively serve as low-pass and high-pass filters for signal representation, which holds good performances. These INR representation theories would provide solid foundations for further developments of continuous representation theories, algorithms, and applications for diverse fields.\par
Furthermore, we shift our focus on the representation and approximation analysis of a class of functional decomposition works, which are related to the INR-based functional tensor decomposition methods \cite{CoordX,LRTFR,FTD_PINN,Separable_PINN}. However, INR-based functional tensor decomposition methods currently lack sufficient theoretical explanations on their representation and approximation abilities. We suppose that existing frameworks from the harmonic and Fourier analysis coupled with theoretical tensor decomposition frameworks \cite{SIAM_review} can pave the way on explaining the expressiveness of these methods \cite{CoordX,LRTFR,FTD_PINN,Separable_PINN}.\par
Griebel and Li \cite{SIAM_Bivariate_function_SVD} studied the decay rate of singular values of a group of bivariate functions in the Sobolev space $W^{s,2}(D)$ where $D\subset {\mathbb R}^n$. Given a bivariate function $\kappa(y,x)$, the singular value decomposition (SVD) refers to the expansion
\begin{equation}\label{FunSVD}
	\kappa(y,x) = \sum_{n=1}^{\infty} \sqrt{\lambda_n} \phi_n(x) \psi_n(y),
\end{equation}
where $\lambda_n$ are eigenvalues of the integral operator $\mathcal{R}$ with associated kernel $R(x,x') = \int_{\Omega} \kappa(y,x)\kappa(y,x')\,dy$, $\phi_n(x)$ are corresponding eigenfunctions, and $
	\psi_n(y) = \frac{1}{\sqrt{\lambda_n}} \int_D \kappa(y,x) \phi_n(x) \, \mathrm{d}x.$
The theory \cite{SIAM_Bivariate_function_SVD} establishes the order $O(M^{-s/d})$ for the truncation error of the SVD series expansion \eqref{FunSVD} after the $M$-th term truncation \cite{SIAM_Bivariate_function_SVD}. In signal processing, SVD serves as a fundamental method for dimensional reduction tasks such as PCA and data compression. Building upon this foundation, a promising research lies in investigating the SVD truncation theory of INR-based continuous representations to uncover theoretical insights of functional tensor decomposition methods \cite{CoordX,LRTFR}. \par 
Following this line of research, Wang et al. \cite{NeurIPS_FtSVD} proposed the functional tensor SVD (t-SVD) via tensor-tensor product, which extends the classical t-SVD to infinite and continuous feature domains. 
\begin{theorem}[Functional $t$-SVD \cite{NeurIPS_FtSVD}]
	\label{thm:ftsvd}
	Let $F : \mathcal{X} \times \mathcal{Y} \rightarrow \mathbb{R}^K$ be a square-integrable vector-valued function with Lipschitz-smooth domains $\mathcal{X} \subset \mathbb{R}^{D_1}$ and $\mathcal{Y} \subset \mathbb{R}^{D_2}$. Then, there exist sets of functions $\{\phi_i\}_{i=1}^{\infty} \subset L^2(\mathcal{X}; \mathbb{R}^K)$ and $\{\psi_i\}_{i=1}^{\infty} \subset L^2(\mathcal{Y}; \mathbb{R}^K)$, and a sequence of $t$-scalars $\{\sigma_i\}_{i=1}^{\infty} \subset \mathbb{N}^K$ with $\lim_{i \to \infty} \sigma_i = 0$, satisfying the functional $t$-Singular Value Decomposition (Ft-SVD):
	\begin{equation}
		F(x, y) = \sum_{i=1}^{\infty} \phi_i(x) *_M \sigma_i *_M \psi_i(y),
		\label{eq:ftsvd}
	\end{equation}
	where $*_M$ denotes the tensor-tensor product induced by the transform $M$. The orthonormality conditions $
		\int_{\mathcal{X}} \phi_i(x) *_M \phi_j(x) \, dx = \delta_{ij}M^{-1}(\mathbf{1})$ {and} $\int_{\mathcal{Y}} \psi_i(y) *_M \psi_j(y) \, dy = \delta_{ij}M^{-1}(\mathbf{1})$ hold, where $\mathbf{1} \in \mathbb{R}^{1 \times 1 \times K}$ is the $t$-scalar with all entries equal to 1, and $\delta_{ij}$ is the Kronecker delta.
\end{theorem} 
The theorem states that the vector-valued function $F$ can be decomposed into the t-SVD form in the continuous Lipschitz domain. The authors applied this novel functional tensor decomposition model to the multi-output regression problem, and resolves the combinatorial distribution shift problem in multi-output regression. They also developed an empirical risk minimization to address the regression problem using Ft-SVD, and demonstrated the theoretical analyses for the performance guarantee of the algorithm.\par
Han et al. \cite{JASA_FunTSVD} proposed the CP low-rank functional tensor decomposition model, a novel dimension reduction framework for tensors with one functional mode and several tabular modes. Let $\mathcal{Y} \in \mathbb{R}^{p_1 \times p_2 \times [0,1]}$ be the functional tensor with approximately CP rank $r$, then the CP functional tensor decomposition model of $\cal Y$ is modeled by
\begin{equation}
	\mathcal{Y} = \mathcal{X} + \mathcal{Z}, \quad \mathcal{X} = \sum_{l=1}^r \lambda_l \mathbf{a}_l \circ \mathbf{b}_l \circ \xi_l \in \mathbb{R}^{p_1 \times p_2 \times [0,1]},
	\label{eq:cp_decomp}
\end{equation}
where $\mathbf{a}_l,\mathbf{b}_l$ are factor vectors and $\xi_l$ is the factor function in a reproducing kernel Hilbert space (RKHS) spanned by a positive-definite kernel function. The authors provided the identifiability condition of the CP functional model. Then, an RKHS-based power iteration method was proposed to estimate the model parameters from a discrete tensor, which iteratively updates the tabular factors $\mathbf{a}_l,\mathbf{b}_l$ and the function $\xi_l$ without function parameterization. Finally, the authors developed a non-asymptotic contractive error bounds for the proposed power iteration algorithm and applied the algorithm to real tensor data analysis.\par 
Gorodetsky et al. \cite{FunTT} developed a new approximation algorithm for representing and computing with multivariate functions using the functional tensor-train, which replaces the three-dimensional TT cores with univariate matrix-valued functions, similar to that of \cite{TT_regression}. They provided parameterization error of using the TT model for approximating a multivariate function, and applied the numerical functional TT algorithm for approximating functions with local features (such as discontinuities) and adaptive integration and differentiation of multivariate functions. Recently, Fageot \cite{Season_trend} proposed the theoretical guarantee for the inverse problem of recovering a continuous-domain function from a finite number of noisy linear measurements, where the target function is the sum of a slowly varying trend function and a periodic seasonal function, and the reconstruction is formalized by introducing a convex generalized total variation regularization over functions. These theoretical frameworks provide fundamental insights and performance guarantee for continuous representation methods. In future research, it would be interesting (and more challenging) to further investigate tractable and theoretically guaranteed computation, decomposition, and reconstruction algorithms for multivariate neural continuous functions, such as INRs and neural operator learning that operates directly on INRs \cite{SP_for_INR,O-INR}. 
\subsection{Generalization and Convergence}\label{sec:NTK}
In this section, we review several neural tangent kernel (NTK)-based theoretical frameworks that reveal the underlying generalization and convergence capabilities of continuous representation methods using INRs (and also more general deep neural nets). The NTK describes the behavior of infinitely wide neural networks during gradient descent training using the corresponding kernel gradient descent under the NTK, and also explains several optimization challenges (such as the spectral bias in INRs that the network favors low-frequency information more that high-frequency ones \cite{PE,INR_Reparameterize}). The NTK measures how a small change in parameters affects the outputs for two network inputs $x$ and $x'$. For wide enough neural nets, this measure stays fixed, simplifying training analysis.\par 
Jacot et al. \cite{NTK} firstly introduced the NTK framework for neural network convergence and generalization. During gradient descent on the parameters of a neural network $f_\theta$ with depth $L$, the network follows the kernel gradient of the functional cost w.r.t. the NTK, where the NTK function is defined as 
\begin{align*}
	\Theta^{(L)}(\theta) = \sum_{p=1}^P \partial_{\theta_p} F^{(L)}(\theta) \otimes \partial_{\theta_p} F^{(L)}(\theta),
\end{align*}
where $\Theta^{(L)}(\theta)$ is the NTK, $F^{(L)}(\theta)$ is the realization function mapping $\theta$ to $f_\theta$, and $\otimes$ is the tensor product between two functions. While the NTK is random at
initialization and varies during training of the neural network $f_\theta$, in the infinite-width limit the NTK $\Theta^{(L)}(\theta)$ converges to an explicit limiting kernel and stays constant during training. These are the two key theoretical results presented in \cite{NTK}.
\begin{theorem}[Convergence to the NTK with infinite-width at initialization \cite{NTK}]
For a neural network of depth \( L \) at initialization, with a Lipschitz nonlinearity \(\sigma\), and in the limit as the network layers width \( n_1, \ldots, n_{L-1} \to \infty \), the NTK \(\Theta^{(L)}\) converges in probability to a deterministic limiting kernel $
\Theta^{(L)} \to \Theta_{\infty}^{(L)} \otimes {Id}_{n_L}$, where \(\Theta_{\infty}^{(L)} : \mathbb{R}^{n_0} \times \mathbb{R}^{n_0} \to \mathbb{R}\) is a scalar kernel and ${Id}_{n_L}$ is an identity matrix.
\end{theorem}
\begin{theorem}[Convergence to the NTK during training \cite{NTK}]
	Assume that \(\sigma\) is a Lipschitz nonlinearity function with bounded second derivative. Under mild assumptions, we have that, uniformly for any training time \(t \in [0, T]\), 
	$\Theta^{(L)}(t) \to \Theta_{\infty}^{(L)} \otimes {Id}_{n_{L}}$, where $\Theta^{(L)}(t)$ is the NTK at time $t$.
\end{theorem}
Since the NTK of an infinite-width neural network converges to a deterministic kernel for both initialization and during training, we can readily describe the convergence and generalization features of this neural network by depicting the corresponding kernel. For instance, Arora et al. \cite{Fine_grained_NTK} proposed the classical fine-grained analysis of convergence and generalization of overparameterized two-layer neural nets based on NTK. They considered a two-layer ReLU-activated neural network $f$ and the corresponding training objective $\Phi$ with input-label pairs $\{{\bf x}_i,y_i\}$:
\begin{align*}
	f_{{\bf W},{\bf a}}({\bf x}) = \frac{1}{\sqrt{m}}\sum_{r=1}^{m} a_r \sigma({{\bf w}_r^\top {\bf x}}),\;	\Phi({\bf W}) = 
	\frac{1}{2}\sum_{i=1}^{n}\left(y_i-f_{\bf W, \bf a}({\bf x}_i)\right)^2.
\end{align*}
Using the eigen-decomposition of the NTK matrix ${\bf H}^{\infty}= \sum_{i=1}^n \lambda_i {\bf v_i} {\bf v_i^\top}$ (${\bf H}^{\infty}\in{\mathbb R}^{n\times n}$ is the evaluation of the NTK function on input data points ${\bf x}_1,\cdots,{\bf x}_n$), the optimization dynamic and generalization error bound of this neural network can be analyzed. 
\begin{theorem}[Training convergence rate \cite{Fine_grained_NTK}]\label{thm:convergence_rate}
	Suppose ${\bf y}=[y_1,\cdots,y_n]$, $\lambda_0 = \lambda_{\min}({\bf H}^\infty) >0$,
	$\kappa = O\left( \frac{\epsilon \delta}{\sqrt n} \right)$,
	$m = \Omega\left( \frac{n^7}{\lambda_0^4 \kappa^2 \delta^4 \epsilon^2} \right)$ and $\eta = O\left( \frac{\lambda_0}{n^2} \right)$. 
	Then with probability at least $1-\delta$ over the random initialization, for all training steps $k=0, 1, 2, \ldots$ we have the training error follows
	\begin{equation} \label{eqn:u(k)-y_size}
		\|{\bf y}-[f_{{\bf W}(k), \bf a}({\bf x}_1),\cdots,f_{{\bf W}(k), \bf a}({\bf x}_n)]\|_2
		= \sqrt{\sum_{i=1}^{n}(1-\eta\lambda_i)^{2k} \left({\bf v}_i^\top {\bf y}\right)^2} \pm \epsilon.
	\end{equation}
\end{theorem}
\begin{figure}[t]
	\scriptsize
	\setlength{\tabcolsep}{0.9pt}
	\begin{center}
		\begin{tabular}{c}
			\centering
			\includegraphics[height=0.35\textwidth,width=0.7\textwidth]{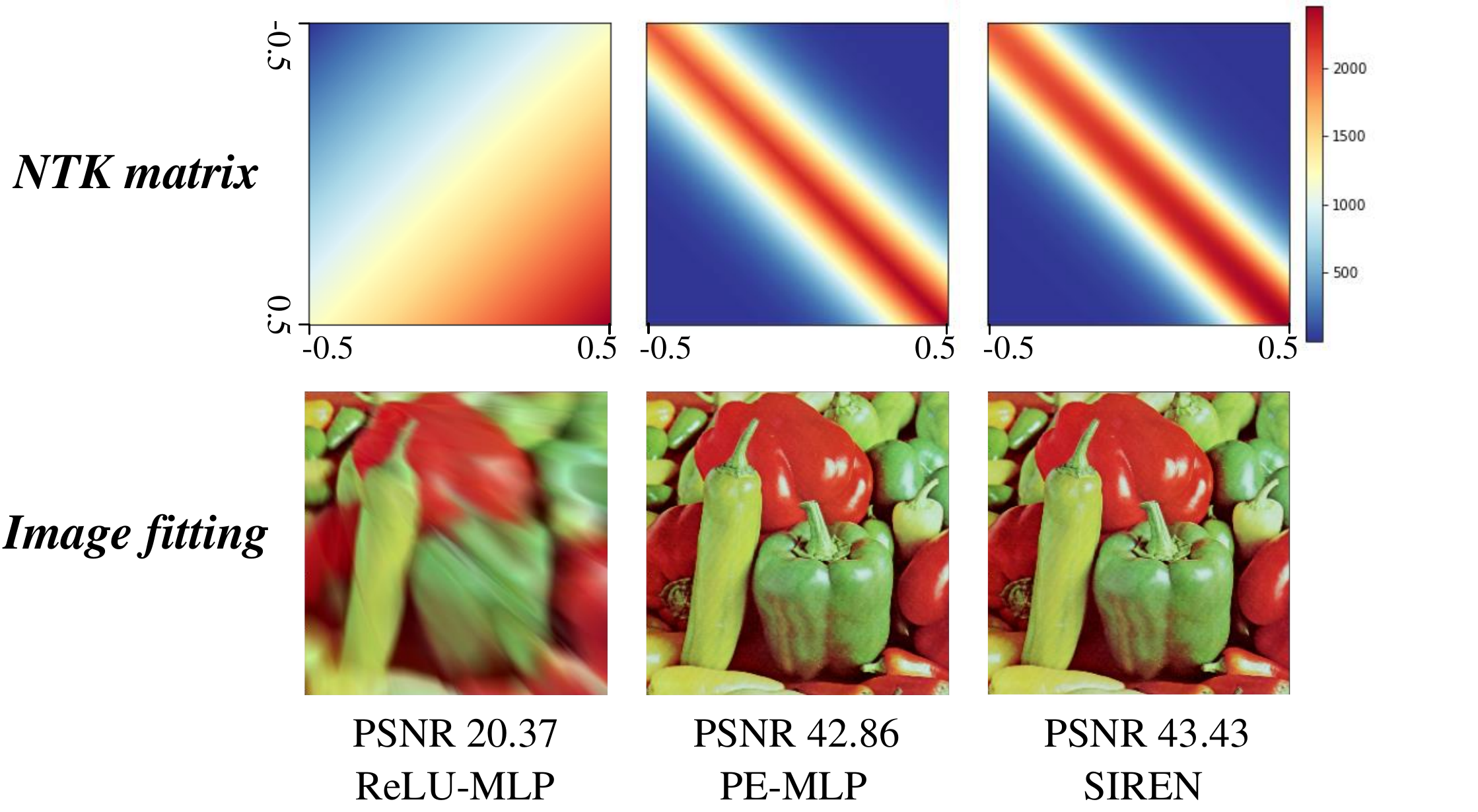}
		\end{tabular}
	\end{center}
	\vspace{-0.4cm}
	\caption{Neural tangent kernel diagrams of several neural networks: ReLU-activated MLP, positional encoding MLP (PE-MLP) \cite{PE}, and SIREN \cite{SIREN}. The INR methods PE-MLP and SIREN shift the NTK matrix from non-diagonal to diagonal, thus enabling more accurate continuous representations of an image.}
	\label{fig:NTK}
\end{figure}
Following the analysis in \cite{Fine_grained_NTK}, we define \(\xi_i(k) = (1 - \eta \lambda_i)^{2k} (\mathbf{v}_i^T \mathbf{y})^2\), and each sequence starts at \(\xi_i(0) = (\mathbf{v}_i^T \mathbf{y})^2\) and decreases at ratio \((1 - \eta \lambda_i)^2\). In other words, we can think of decomposing the label vector \(\mathbf{y}\) into its projections onto all eigenvectors \(\mathbf{v}_i\) of \(\mathbf{H}^\infty\): \(\|\mathbf{y}\|_2^2 = \sum_{i=1}^n (\mathbf{v}_i^T \mathbf{y})^2 = \sum_{i=1}^n \xi_i(0)\), and the \(i\)-th portion shrinks exponentially at ratio \((1 - \eta \lambda_i)^2\). The larger \(\lambda_i\) is, the faster \(\{\xi_i(k)\}_{k=0}^\infty\) decreases to 0. In order to have faster convergence, we would like the projections of \(\mathbf{y}\) onto top eigenvectors to be larger. For a set of labels \(\mathbf{y}\), if they align with the top eigenvectors, i.e., \((\mathbf{v}_i^T \mathbf{y})^2\) is large for large eigenvalues \(\lambda_i\), then gradient descent converges quickly. Otherwise, the optimization would converge slowly to components that correspond to smaller eigenvalues. This property induces the well-known spectral bias in neural networks. Addressing the spectral bias issue is the motivation of many INR methods. For instance, the Fourier reparameterized INR method \cite{INR_Reparameterize} reconfigures the eigenvalue distribution of the NTK through reparameterization of MLP layers, hence enabling more projections onto top eigenvectors, leading to faster convergence towards high-frequency components. The method in \cite{FINER} alleviates the spectral bias of INR by flexible bandwidth tuning of the NTK through a delicately designed activation function.\par
Since the training of a neural network using gradient descent mimics the kernel gradient descent w.r.t. the NTK, we can consider the scenario of using the kernel regression for continuous representation (i.e., mapping coordinates to values using kernel regression). If the kernel $k(x,x')$ is non-diagonal (i.e., input coordinates $x$ and $x'$ that are close to each other do not result in high responses $k(x,x')$), it is hard to use this kernel to fit a good regression model in the continuous space. Otherwise, if the kernel $k(x,x')$ is diagonal and shift-invariant, it would be more suitable. This is the underlying principle of many feature encoding-based INR methods, such as the Fourier feature PE \cite{PE} and SIREN \cite{SIREN}. The Fourier feature encoding shifts the NTK of a ReLU-activated neural network from non-diagonal to diagonal, hence enabling effective continuous representation (see {Fig. \ref{fig:NTK}}).\par 
In terms of generalization error bound, Arora et al. \cite{Fine_grained_NTK} developed an NTK-based bound that fully depends on data pairs $\{{\bf x}_i,y_i\}$ and the structure of the neural network through the NTK matrix ${\bf H}^{\infty}$. The generalization theory states that, suppose the data \(S = \{ ({\bf  x}_i, {y}_i) \}_{i=1}^n\) are i.i.d. samples from a distribution \(D\) and consider any loss function \(\ell: \mathbb{R} \times \mathbb{R} \to [0, 1]\) that is \(1\)-Lipschitz and sufficiently wide neural network, then with high probability $1-\delta$, the two-layer neural network \(f_{{\bf W}(k), \mathbf{a}}\) trained by gradient descent has generalization loss \(L_D(f_{{\bf W}(k), \mathbf{a}}) = \mathbb{E}_{({\bf  x}, y) \sim D} \left[ \ell(f_{{\bf W}(k), \mathbf{a}}({\bf  x}), y) \right]\) bounded as:
	\[
	L_D(f_{{\bf W}(k), \mathbf{a}}) \leq \sqrt{\frac{2{\bf y}^{\top} (\mathbf{H}^{\infty})^{-1}{\bf y}}{n}} + O \left( \sqrt{\frac{\log \frac{n}{\lambda_0 \delta}}{n}} \right),
	\]
where the first term $\sqrt{\frac{2{\bf y}^{\top} (\mathbf{H}^{\infty})^{-1}{\bf y}}{n}}$ can be viewed as a complexity measure of data that one can use to predict the test accuracy of the
learned neural network. This result gives an illustrative example on how to use the generalization theory of traditional kernel methods to analyze the generalization behavior of sufficiently wide neural networks, bringing new theoretical tools for neural nets. Indeed, enhancing the generalization capabilities of INRs for unseen coordinates remains a crucial research objective \cite{WIRE_ICLR_25}. Rigorous theoretical analysis of generalization error bounds for INR methods is warranted, as it would facilitate the development of principled approaches with provable generalization guarantees.\par
Many subsequent works have provided further analysis to NTK-based frameworks. Arora et al. \cite{wideNN_NTK} gave the first exact algorithm for computing the NTK of a convolutional neural network, and proposed a non-asymptotic proof that a wide neural network is equivalent to the NTK regression. They also proposed an efficient GPU implementation of this algorithm, and reported a state-of-the-art performance using a pure kernel-based method for image classification. While conventional NTK analysis does not work when there is a standard $\ell_2$-norm regularizer (known as weight decay), Wei et al. \cite{l2_reg_NTK} showed that regularized neural nets have better generalization abilities than the corresponding NTK without regularizer, and developed the convergence of $\ell_2$-norm regularized neural nets. Specifically, they proved that for infinite-width two-layer neural nets, gradient descent optimizes the regularized neural net loss to a global minimum. Chen et al. \cite{RKHS_norm_reg} proposed a function-space regularization for training neural nets instead of the conventional $\ell_2$-norm regularizer. This method approximates the norm of neural network functions by the RKHS norm under the NTK and uses it as a function-space regularizer. The authors proved that neural networks trained using this regularizer are arbitrarily close to kernel ridge regression solutions under the NTK. Furthermore, they provided a generalization error bound under the RKHS norm regularizer and empirically demonstrated improved generalization on downstream tasks. \par
While conventional NTK analysis deals with gradient descent with full-batch settings (noise-less setting), understanding the training dynamic of neural nets under stochastic optimization is of crucial importance. Chen et al. \cite{NTK_two_layer_noisyGD} proposed a more generalized NTK analysis of two-layer neural net, which considers the setting with noisy gradient descent (e.g., mini-batch) and weight decay, and such generalized setting also exhibit a ``kernel-like'' behavior, implying that the training loss converges linearly. This is a more practical optimization setting since full-batch gradient descent is usually computationally heavy in practice. Accelerating the convergence of deep neural nets via preconditioning is also a widely studied topic. For instance, Geifman et al. \cite{TMLR_NTK_control} proposed a family of modified spectrum kernels (MSK's) and introduced the MSK's-based preconditioned gradient descent by altering eigenvalues of NTK for accelerating the convergence of wide neural networks, which gets rid of the dependency of convergence speed on NTK. Recently, Chng et al. \cite{INR_precondition} proposed a stochastic training scheme for INRs by using the curvature-aware diagonal preconditioners for second-order optimization methods, which are especially useful for non-traditional activation-based INR training such as SIREN \cite{SIREN} and WIRE \cite{WIRE}.\par 
As relative theoretical frameworks, Alkhouri et al. \cite{untrained_survey_2025} summarized the theories and algorithms of using untrained DNNs (for instance, untrained neural networks including INRs and deep image priors \cite{DIP}) for data reconstruction, which establishes NTK dynamics analysis and reconstruction error estimations for such an untrained paradigm. This work connects the broad category of untrained neural net methods with theoretical frameworks of NTK to explain the underlying effectiveness of these unsupervised methods. Recently, Audia et al. \cite{InstantNGP_NTK} demonstrated that grid encoding continuous representations (such as InstantNGP \cite{InstantNGP}) recovers fine details better than conventional INRs from the NTK eigenvalue perspective. Specifically, they showed that the smallest eigenvalue of the NTK matrix induced by InstantNGP is greater than that of an MLP, and the expressiveness mainly comes from the additional learnable grid parameters, which uncovers the theoretical insights about the stronger representation abilities of grid encoding continuous representation methods.\par
The NTK provides a powerful framework for analyzing the properties and theoretical advantages of continuous representation models, such as INRs. By leveraging NTK theory, researchers can examine key characteristics of INRs, including analyzing the shift-invariant property of the NTK (a desired property for continuous representation) when passing the input coordinates through a Fourier mapping \cite{PE}, the bandwidth of NTK and its corresponding parameter configurations/initialization schemes of INR \cite{FINER}, and the desired eigenvalue distribution of NTK for faster convergence of INR \cite{INR_Reparameterize}. One can utilize the NTK framework for a wider range of analysis involving non-conventional neural structures, such as the theoretical insights of PDE-constrained neural solvers \cite{NeurIPS_nonlinear_PINN_NTK} and neural operator learning \cite{NO_NTK}. Utilizing the NTK analysis tool to advance fundamental theoretical understanding for various applications such as full waveform inversion \cite{IFWI_JGR} and computational biology \cite{STINR} are also promising interdisciplinary research directions in future work.
\subsection{Implicit Regularization}
Implicit regularization \cite{Implicit_regularization_NIPS_2019,Implicit_reg_CNN} refers to the theoretical analysis that characterizes the implicit bias or structure constraint of an optimization algorithm over a parameterization model (e.g., neural networks), even without explicit regularization or constrained conditions. These theoretical analyses, in part, explain the good generalization ability of deep neural networks even with over-parameterization. In particular, continuous representations (e.g., INRs \cite{SIREN,PE}) are representative instances that, even with over-parameterization, could still learn robust and generalizable features (e.g., to unseen coordinates) from limited training data, which can be understood from the implicit regularization perspective of training dynamics. Compared with NTK analysis, implicit regularization analysis that based on matrix/tensor factorization dose not depend on infinite-width assumption.\par
Gunasekar et al. \cite{IR_NIPS17} firstly developed the implicit regularization of gradient descent over the matrix factorization of the form $X=UU^T\in{\mathbb R}^{n\times n}$ where $U\in{\mathbb R}^{n\times d}$ is an over-parameterized matrix factor (i.e., $n=d$). They showed that with small enough step sizes and initialization close enough to the origin, gradient descent on such a full-dimensional matrix factorization converges to the minimum nuclear norm solution. Consequently, Arora et al. \cite{Implicit_regularization_NIPS_2019} studied the implicit regularization of gradient flow over a deep matrix factorization of the form $X=W_NW_{N-1}\cdots W_1$, where $W_n$ are over-parameterized matrix factors. The main result in \cite{Implicit_regularization_NIPS_2019} states that gradient descent over the following matrix sensing minimization problem
\begin{equation}\label{IR}
\min_{W_1,\dots, W_N} \phi(W_1(t), \ldots, W_N(t)):=\frac{1}{2}\sum_{i=1}^m (y_i - \langle A_i, W_NW_{N-1}\cdots W_1\rangle)^2\,
\end{equation}
converges to the minimal nuclear norm solution, a conjecture similar to that of \cite{IR_NIPS17}. 
\begin{theorem}[Implicit regularization of deep matrix factorization \cite{Implicit_regularization_NIPS_2019}]
Let $W_{\text{deep},\infty}(\alpha) := \lim_{t \to \infty} W_N(t)$ \\$W_{N-1}(t) \cdots W_1(t)$ where \(W_j(0) = \alpha I\) and \(W_j(t) = -\frac{\partial \phi}{\partial W_j}(W_1(t), \ldots, W_N(t))\) for \(t \in \mathbb{R}_{\geq 0}\). Suppose \(N \geq 3\), and that the matrices \(A_1, \ldots, A_m\) commute. Then, if \(\bar{W}_{\text{deep}} := \lim_{\alpha \to 0} W_{\text{deep},\infty}(\alpha)\) exists and is a global optimum for \eqref{IR} with zero loss, then \(\bar{W}_{\text{deep}}\) is a global optimum with minimal nuclear norm.
\end{theorem}
The tendency of gradient-based optimization over deep matrix factorization towards a low nuclear norm solution shed light on the generalization mystery of deep neural networks, i.e., a low ``complexity'' model would generalize better by such implicit regularization of gradient descent. To further distinguish the influence of depth $N$ on the implicit regularization, the authors \cite{Implicit_regularization_NIPS_2019} proposed dynamical analysis of gradient flow with infinitesimally small
learning rate on deep matrix factorizations. They showed that the evolution of singular values of the recovered matrix $X$ is essentially related to the depth $N$. Specifically, the derivative of the $r$-th singular value $\sigma_{r}$ of the recovered matrix w.r.t. training time $t$ is formulated by
\[
\dot{\sigma}_{r}(t) = - {N} \cdot \left( \sigma_{r}^{2}(t) \right)^{1-1/N} \cdot \left\langle \nabla \ell(W(t)), \mathbf{u}_{r}(t) \mathbf{v}_{r}^{\top}(t) \right\rangle,
\]
where $N$ is the depth, $\nabla \ell(W(t))$ is the gradient of the loss function, and $\mathbf{u}_{r}(t),\mathbf{v}_{r}(t)$ are singular vectors of the recovered matrix at time $t$. This gives the evolution dynamic of the singular value w.r.t. time $t$. We can see that the dynamic enhances the movement of large singular values, and on the other hand attenuates that of small ones, ultimately leading to a low-rank solution. Moreover, the evolution of singular values depends on the depth $N$, and the attenuation of singular values becomes more significant as $N$ grows. This theoretically reflects the influence of depth on the convergence rate towards the low nuclear norm solution. It could be noted that the INR \cite{PE,InstantNGP} can be seen as a special case of the deep matrix factorization with input layer $W_1$ being coordinate embedding and without nonlinear activation. Further studying the implicit regularization of gradient optimization over INRs is an important theoretical direction towards understanding its effectiveness. \par 
Subsequently, Razin et al. \cite{Implicit_reg_CNN} studied the implicit regularization in hierarchical tensor factorization, a model similar to a certain deep convolutional neural network. Through dynamical analysis, they established implicit regularization of the model towards low hierarchical tensor rank, hence translating to an implicit regularization towards locality (i.e., low separation rank) for the associated convolutional network. 
Here, the separation rank of a function $f$ w.r.t. an index set $I$ is defined as the smallest number $R$ such as the function can be separated into
\[
f\left(\mathbf{x}^{(1)}, \ldots, \mathbf{x}^{(N)}\right) = \sum_{r=1}^{R} g_r \left( \left( \mathbf{x}^{(i)} \right)_{i \in I} \right) \cdot \bar{g}_r \left( \left( \mathbf{x}^{(j)} \right)_{j \in [N] \setminus I} \right)
\]
with some separated functions $g_r,\bar{g}_r$. The lower separation rank, the stronger local (short-range) dependencies between different input regions $\mathbf{x}^{(i)}$ and $\mathbf{x}^{(j)}$. Hence, to counter the locality implicit regularization and enhance the long-range dependency evacuation of the deep network, the authors \cite{Implicit_reg_CNN} designed an explicit regularization that discourages locality (i.e., discourages low separation rank), and demonstrated its effectiveness to improve the long-range dependency characterization of convolutional network for image classification performance.\par
Similar implicit regularization analyses in deep tensor factorizations are conducted in \cite{Implicit_regularization_deep_CP,Implicit_regularization_deep_Tucker}. Hariz et al. \cite{Implicit_regularization_deep_CP} studied implicit regularization in deep tensor CP factorization in the form of 
\[
{\cal W} = \sum_{r=1}^{R} \bigotimes_{n=1}^{N} \prod_{i=1}^{k_n} {\bf A}_i^{n,r} {\bf w}_r^n,
\]
where $\bigotimes$ denotes the outer-product and ${\bf A}_i^{n,r},{\bf w}_r^n$ are factors of the deep tensor CP factorization, with $k_n$ indicating the depth. They established the training dynamic of the vector component $\bigotimes_{n=1}^N {\bf w}_r^n(t)$ w.r.t. the training time $t$ using infinitely small learning rate of gradient descent, which writes   

\[
\frac{d}{dt} \left\| \bigotimes_{n=1}^N {\bf w}_r^n(t) \right\| = N\delta_r(t) \left\| \bigotimes_{n=1}^N {\bf w}_r^n(t) \right\|^{2 - \frac{2}{N} + \frac{k_1 + \cdots + k_N}{N}},
\]
where $\delta_r(t)$ can be expressed independent of depth. This shows that the evolution rates of the CP component $\bigotimes_{n=1}^N {\bf w}_r^n(t)$ are proportional to $\left\| \bigotimes_{n=1}^N {\bf w}_r^n(t) \right\|^{2 - \frac{2}{N} + \frac{k_1 + \cdots + k_N}{N}}$. This evolution rate is polynomially related to the depth $k_n$, i.e., when the depth increases, the evolution becomes faster. Since the CP blocks with larger norms decay faster according to the evolution rate, these facts ultimately lead to low-CP-rank solution, even without explicit low-rank constraints. The implicit low-tensor-rank regularization could yield more accurate estimations and
better convergence properties for deep tensor factorizations. Similarly, their later work \cite{Implicit_regularization_deep_Tucker} studied the implicit regularization in deep tensor Tucker factorization, which shows that deep Tucker factorization trained by gradient descent induces a structured sparse regularization, leading to the solution with low-multilinear-rank. The deep Tucker factorization here is tightly connected to a certain deep neural network. These results could draw inspirations for depicting the implicit regularization of a class of functional tensor decomposition methods based on INR \cite{CoordX,LRTFR,Separable_PINN}, since these methods can be exactly formulated as deep tensor factorization models parameterized by deep neural networks, with the input layers being coordinate embedding. \par 
Recently, Bai et al. \cite{NeurIPS24_IR} studied the influence of the connectivity of the graph induced by the observed matrix on the implicit regularization. They showed that disconnected observations lead to low nuclear norm regularization while connected ones lead to low-rank solutions, supported by training dynamic analysis. Except for standard gradient descent, many studies have developed implicit bias analysis for other popular optimizers in deep learning. For instance, Xie and Li \cite{AdamW_implicit} studied the implicit bias of the widely utilized AdamW optimizer, and showed that, under full-batch training, it implicitly performs constrained optimization in terms of $\ell_\infty$-norm conditioned on the weight decay factor. Future researches can combine advanced optimizers with deep matrix/tensor factorizations, and consider more stochastic optimization (such as stochastic gradient descent) to deepen the insights into implicit regularization induced by different models and algorithms, which would be beneficial for explaining some specific deep models, such as continuous representation using INRs.\par
The implicit regularization encoded in deep models and optimization algorithms can explain the good generalization ability of INRs for unseen coordinates and across different images, such as the generalization of the arbitrary-scale image restoration model \cite{LIIF}. However, most current implicit regularization results rely on pure linear combination, and how to extend the results to scenarios with nonlinearity remains challenging. Moreover, deriving the theoretical implicit regularization of more complex models such as neural TV-induced optimization dynamics \cite{NeurTV}, differential equations represented by neural nets \cite{NeurIPS_nonlinear_PINN_NTK}, and neural operator learning \cite{DeepONet} are open areas.  
\section{Practical Data Recovery Applications of Continuous Regularizations}\label{sec:applications}
This section conducts a systematic review of representative applications of continuous representation methods, organized through four application paradigms: Low-level vision and graphics, scientific computing (particularly numerical PDE solutions), biomedical imaging and bioinformatics, and geosciences and remote sensing.  An overview of some representative application works using continuous representation methods is synthesized in Table \ref{tab:appl}.  
\begin{table}[t]
	\centering
	\scriptsize
	\belowrulesep=0pt
	\aboverulesep=0pt
	\setlength{\tabcolsep}{1.5pt}
	\renewcommand{\arraystretch}{1.1} 
	\caption{Review of some representative applications of continuous representation methods.\label{tab:appl}}
	\begin{tabular}{ll|l|c|c}
		\midrule
		\textbf{Category} & \textbf{Application} & \textbf{Description} & \textbf{Year} & \textbf{Ref.} \\
		\midrule
		\multirow{9}*{\tabincell{l}{Vision and \\ Graphics}}
		& \tabincell{l}{(DeepSDF) Shape \\representation}  & \tabincell{l}{Learn signed distance function (SDF) of shapes by continuous\\ neural representation that implicitly encodes the boundary.} & 2019 & {\cite{DeepSDF}}\\\cmidrule{2-5}
		 ~& \tabincell{l}{(NeRF) Neural\\ radiance fields}& \tabincell{l}{Represent scenes for view synthesis using continuous \\5D INR and volume rendering.} & 2020 & {\cite{NeRF}}\\
		\cmidrule{2-5}
		~& \tabincell{l}{(LIIF) Arbitrary-scale\\image super-resolution} & \tabincell{l}{Introduce the local implicit image function using INR for \\image super-resolution with arbitrary upsampling scales.} & 2021 & {\cite{LIIF}}\\
		\cmidrule{2-5}
		~& \tabincell{l}{(PCU) Point cloud \\upsampling} & \tabincell{l}{Self-supervised and arbitrary-scale PCU by seeking nearest \\projection points on the implicit surface of seed points.} & 2023 & {\cite{PAMI_PCU}}\\	\cmidrule{2-5}
		~& \tabincell{l}{Cross-scale \\image deblurring} & \tabincell{l}{Introduce self-supervised method for cross-scale blind image \\deblurring using INR to represent image and kernel.} & 2024 & {\cite{NeurIPS_Zhang}}\\
		\midrule
\multirow{11}*{\tabincell{l}{Scientific\\Computing}}
& \tabincell{l}{(DeepONet) Deep\\operator network} & \tabincell{l}{Learning nonlinear operators using two deep neural nets to\\ respectively take input functions and coordinates.}& 2021 & \cite{DeepONet} \\\cmidrule{2-5}
& \tabincell{l}{(FNO) Fourier \\neural operator} & \tabincell{l}{Parameterize the function integral kernel in the Fourier space \\for expressive operator learning.} & 2021 & \cite{FNO} \\\cmidrule{2-5}
& \tabincell{l}{Physical simulation \\using gird encoding} & \tabincell{l}{Utilize InstantNGP for neural physical simulation with \\a numerical gradient method and fast boundary sampling.} & 2024 & \cite{AAAI_PINN_InstantNGP} \\\cmidrule{2-5}
& \tabincell{l}{InstantNGP for\\ PINN} & \tabincell{l}{Use InstantNGP for physical-informed neural nets with finite\\-difference calculations of derivatives to address discontinuity.} & 2024 & \cite{InstantNGP_PINN_JCP} \\\cmidrule{2-5}
& \tabincell{l}{Tensor decomposed\\ PINN} & \tabincell{l}{Employ CP, Tensor-Train, and Tucker decompositions in \\PINN for efficient learning of multivariate functions.} & 2025 & \cite{FTD_PINN} \\\cmidrule{2-5}
& \tabincell{l}{(D-FNO) Decomposed\\ Fourier neural operator} & \tabincell{l}{Improve the efficiency of FNO ($O(N^3\log N)$ to $O(NP\log N)$)\\ by tensor decomposition and separability of Fourier transform.} & 2025 & \cite{D-FNO} \\
\midrule
		\multirow{13}*{\tabincell{l}{Medical Imaging\\and Bioinformatics}}
		& \tabincell{l}{(IREM) INR for MRI\\ Reconstruction} & \tabincell{l}{High-resolution 3D MRI reconstruction with arbitrary\\ up-sampling rate using INR.} & 2021 & \cite{MRI_INR_MICCAI} \\\cmidrule{2-5}
		& \tabincell{l}{(ArSSR) MRI arbitrary-scale\\super-resolution} & \tabincell{l}{Arbitrary-scale super-resolution for 3D MRI using 3D CNN\\ and local implicit image function.}& 2023 &\cite{MRI_LIIF}\\\cmidrule{2-5}
		& \tabincell{l}{Survey of INR for\\medical imaging} & \tabincell{l}{Deliver a comprehensive survey of INR for medical imaging\\ analysis and reconstruction.} & 2023 & \cite{Medical_imaging_survey} \\\cmidrule{2-5}
		& \tabincell{l}{INR for image\\registration} & \tabincell{l}{Medical image registration using generalized INR with latent\\ modulations obtained by a CNN encoder.} & 2024 & \cite{MICCAI_24_registration} \\\cmidrule{2-5}
		& \tabincell{l}{(Moner) Undersampled \\MRI reconstruction} & \tabincell{l}{Reconstruct MRI and motion parameters from undersampled \\measurements using INR and Fourier-slice theorem.} & 2025 & \cite{ICLR_MRI_25} \\\cmidrule{2-5}
& \tabincell{l}{(STINR) INR for spatial \\transcriptomics} & \tabincell{l}{Decipher multi-slice spatial transcriptomics data using INR\\ and single-cell references.} & 2025 & \cite{STINR} \\\cmidrule{2-5}
		& \tabincell{l}{(GASTON) Spatial gene\\expressions} & \tabincell{l}{Analyze spatial gene expressions through spatial gradients\\ and isodepth learned by INR.} & 2025 & \cite{NM_INR} \\
		\midrule
		\multirow{11}*{\tabincell{l}{Geosciences and \\Remote Sensing}} 
		&\tabincell{l}{HSI super-resolution\\using INR}&\tabincell{l}{Single HSI super-resolution using INR and a content-aware\\ hypernetwork that produces the weights of INR.} & 2023 & \cite{HSI_INR_23} \\\cmidrule{2-5}
				& \tabincell{l}{{(IFWI)} Full waveform \\inversion}&\tabincell{l}{Use INR to parameterize the velocity model for full waveform\\ inversion (FWI) with improved convergence.} & 2023 & \cite{IFWI_JGR} \\\cmidrule{2-5}
& \tabincell{l}{(SINR) Spectral \\reconstruction using INR}&\tabincell{l}{Continuous spectral amplification process using INR for snapshot \\spectral imaging with arbitrary recovered spectral bands.} & 2024 & \cite{HSI_INR_TCSVT} \\\cmidrule{2-5}
&\tabincell{l}{Arbitrary-scale HSI \\super-resolution}&\tabincell{l}{Utilize LIIF for HSI arbitrary-scale super-resolution by leveraging \\and fusing RGB spatial priors.} & 2024 & \cite{HSI_LIIF_24} \\\cmidrule{2-5}
		&\tabincell{l}{Hyperspectral \\unmixing}&\tabincell{l}{Nonnegative matrix functional factorization parameterized by\\ INRs for hyperspectral unmixing.} & 2024 & \cite{HSI_Unmixing_INR} \\\cmidrule{2-5}
		& \tabincell{l}{{(NeRSI)} Seismic data \\interpolation}&\tabincell{l}{Utilize INR to encode continuous seismic 5D wavefield to infer\\ missing trace amplitudes (interpolation).} & 2025 & \cite{5D_seismic_INR} \\
		\midrule
	\end{tabular}
\end{table}
\subsection{Low-Level Vision and Graphics}
Utilizing continuous representation models (particularly INRs) for low-level vision problems and applications in graphics has attracted extensive attention in recent years due to the continuous and differentiable nature of INRs, which make them especially suitable for representing continuous, resolution-independent, and irregular signals in vision and graphics. Here we introduce some representative applications of continuous representation in related fields.\par
\begin{figure}[t]
	\scriptsize
	\setlength{\tabcolsep}{0.9pt}
	\begin{center}
		\begin{tabular}{c}
			\centering
			\includegraphics[width=0.95\textwidth]{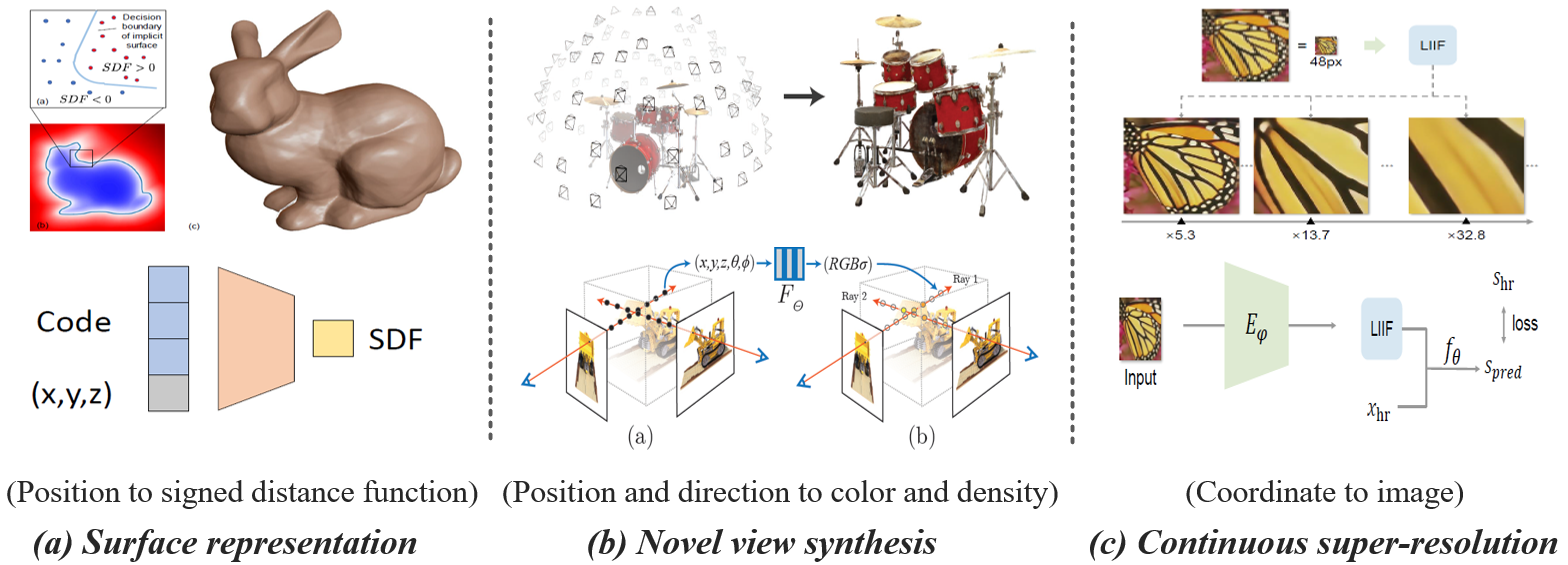}
		\end{tabular}
	\end{center}
	\vspace{-0.4cm}
	\caption{Representative applications of continuous representation models in vision and graphics. (a) Surface representation via signed distance function (SDF) \cite{DeepSDF}, which maps positions to the corresponding SDF to implicitly determine the surface boundary. (b) Novel view synthesis via neural radiance fields \cite{NeRF}, which maps positions and camera view directions to the corresponding colors and volume densities, followed by differentiable volume rendering for view synthesis. (c) Arbitrary-scale image super-resolution paradigm via local implicit image function \cite{LIIF}, which maps coordinates concatenated with latent features to the corresponding high-resolution image. Pictures were taken from \cite{DeepSDF,NeRF,LIIF}.}
	\label{fig:vision}
\end{figure}

 
The classical INR methods \cite{SIREN,PE} were applied to represent images, videos, and surfaces under a unified continuous representation framework, showing strong capability for capturing fine details of visual signals by compressing them into neural representations. Around 2020, there were many pioneer works that develop implicit and continuous methods for 3D scene representation. For instance, Park et al. \cite{DeepSDF} proposed the first signed distance function (SDF)-based implicit representation for shape representation (termed DeepSDF). It utilizes an INR to encode the SDF of a shape, where the SDF is a continuous function that, for a given spatial point in ${\mathbb R}^3$, outputs the point's distance to the closest surface, whose sign encodes whether the point is inside (negative) or outside (positive) of the surface. The underlying shape is implicitly represented by the surface of SDF$=0$ (see Fig. \ref{fig:vision}'s left subfigure). Sitzmann et al. \cite{scene_representation} proposed a continuous and 3D structure-aware scene representation that encodes both geometry and appearance by representing 3D scenes as continuous functions that map world coordinates to a feature representation of local scene properties. Jiang et al. \cite{CVPR_3D_scene} proposed local implicit grid representations for 3D scenes, which reconstructs 3D scenes from point clouds via optimization of the latent grid of an implicit network. Lars et al. \cite{CVPR19_occupancy} proposed the occupancy network, which implicitly represents the 3D surface as the continuous decision boundary of a deep neural network classifier, allowing to extract 3D meshes at any resolution. Yariv et al. \cite{NeurIPS_21_neural_implicit_surface} proposed to model the volume density of 3D scenes as a function of the geometry represented by SDF, which produces high-quality geometry reconstructions. Niemeyer et al. \cite{CVPR20_diff_volume} proposed the differentiable rendering formulation for implicit shape representations, which learns implicit shape and texture representations directly from RGB images by leveraging the analytic expression of the gradients of depth w.r.t. the implicit network parameters. \par 
In terms of shape representation, Genova et al. \cite{ICCV19_Shape} proposed to use structured implicit functions for learning template 3D shapes, and the learned shape templates support downstream applications such as shape exploration, correspondence, interpolation, etc. Chen and Zhang \cite{CVPR19_shape} proposed to use implicit fields for learning generative models of shapes, where an implicit field decoder was developed for improving the visual quality of the generated shapes. Chibane et al. \cite{CVPR20_Shape} proposed the implicit feature networks for 3D shape reconstruction and completion from 3D inputs (such as sparse voxel), which outputs continuous shapes from implicit functions. Recently, Schirmer et al. \cite{CG_SDF_survey} presented a comprehensive survey on geometric INR methods for signed distance functions in scene representation. \par
A famous scene representation model is the neural radiance field (NeRF) \cite{NeRF}. The NeRF represents a scene using an implicit network, which maps a single continuous 5D coordinate
(spatial location $(x,y,z)$ and viewing direction) to the volume density and view-dependent color at that spatial location. Then the classic volume rendering techniques are used to project
the output colors and densities into an image to synthesize novel views (see Fig. \ref{fig:vision}'s middle subfigure). The following works such as InstantNGP \cite{InstantNGP}, PlenOctrees \cite{PlenOctrees}, and Plenoxels \cite{Plenoxels} methods based on grid parametric encoding are more powerful continuous representation methods for scene representation of novel view synthesis based on NeRF. Chen et al. \cite{TensorNeRF} proposed tensorial radiance fields, which decomposes NeRF under the tensor block term decomposition to enable faster convergence without neural networks. Tang et al. \cite{TenosrNeRF_NIPS} proposed compressible and composable NeRF via rank residual decomposition. It learns a tensor decomposition model of NeRF, uses a rank-residual learning strategy to encourage the preservation of primary information in lower ranks, and gradually encodes fine details in higher ranks, allowing composition of scenes by concatenating along the rank dimension. Keil et al. \cite{Kplanes} proposed the $K$-planes representation, which represents a $d$-dimensional scene using $C_d^2$ 2D planes with multi-resolution grids and interpolation. The $K$-planes induce a natural decomposition of static and dynamic components, achieving state-of-the-art reconstruction fidelity with low memory usage for scene representation. For a detailed and comprehensive survey of NeRF-based methods, please refer to \cite{NeRF_review}.\par
Several works have explored the application of INR for image restoration tasks in low-level vision. Chen et al. \cite{LIIF} proposed the local implicit image function (LIIF) for continuous image representation, and applied this representation to arbitrary-scale image super-resolution (see Fig. \ref{fig:vision}'s right subfigure). The LIIF uses an encoder to obtain high-resolution features and feeds the features with image coordinates into an INR to generate the high-resolution image to enable resolution-independent representation of an image. Chen et al. \cite{INR_derain} proposed a multi-scale INR network for image deraining, which takes advantage of the continuous representation to construct multi-scale collaborative representations of an image. Nam et al. \cite{NIR_fusion} proposed neural image representations for multi-image fusion and layer separation, which estimates the homography, optical flow, and occlusion scene motions to effectively combine multiple input images into a single canonical view using coordinate-based neural representations. Cheng et al. \cite{INR_SAR_detection} proposed INRs with imaging geometry for synthetic aperture radar (SAR) target recognition. The target recognition problem can be seen as a view synthesis pipeline which predicts the density and intensity from the input views to render novel views. Zhang et al. \cite{NeurIPS_Zhang} applied the INR for cross-scale self-supervised image deblurring with unknown blur kernel, which represents the image and kernel with INRs to encode resolution-free property for image deblurring. Hu et al. \cite{Heatmap_INR_NIPS} proposed an INR-based continuous heatmap regression method for human pose estimation. It can output the predicted heatmaps at arbitrary resolution during inference, which easily achieves sub-pixel localization precision. Tang et al. \cite{INR_lowlight} proposed to use INR for cooperative low-light image enhancement, which unifies the diverse degradation factors of real-world low-light images using an INR normalization pipeline, thus enhancing robustness. Recently, Zhao et al. \cite{PAMI_PCU} proposed to use implicit surfaces for arbitrary-scale point cloud upsampling, which seeks the nearest projection points on the implicit surface of seed points to sample dense point clouds at arbitrary-scales.\par
There is a group of works that leverage implicit representations for videos in the continuous spatial-temporal domain. For instance, Chen et al. \cite{VideoINR} applied the INR for continuous space-time super-resolution of videos by estimating motion flow fields between frames. Chen et al. \cite{NeRV} proposed the neural representations for videos (NeRV), which represents videos as neural networks taking frame index as input and outputting the corresponding RGB images. The NeRV achieves efficient video compression by neural networks.
Li et al. \cite{E-NeRV} proposed the expedited NeRV by decomposing the image-wise INR into separate spatial and temporal contexts, which are fused by convolution stages. It achieves significantly faster speed on convergence as compared with vanilla NeRV for implicit video representations. Similarly, Yan et al. \cite{DS_NeRV} proposed the implicit NeRV with decomposed static and dynamic codes for video representation, which efficiently utilizes redundant static information while maintaining high-frequency details. \par
There are also several works that generalize INRs to graph-structured data on non-Euclidean spaces. For instance, Grattarola and Vandergheynst \cite{Spectral_embedding_graph} proposed the generalised INR for discrete graph representation on non-Euclidean domains. It utilizes the eigenvectors of the graph Laplacian matrix as the spectral embedding of INRs, which allows the training of INRs without knowing the underlying continuous domain, a scenario for most real-world graph signals. Xia et al. \cite{IGNR} proposed the implicit graphon neural representation, which parameterizes the adjacency matrix of graphs by INRs and enables efficient and flexible generation of arbitrary sized graphs.\par 
Representing and reconstructing vision and graphics signals using continuous representation (such as coordinate networks and spherical harmonics on voxel) has led to a group of increasingly popular research fields (NeRF \cite{NeRF}, LIIF \cite{LIIF}, SDF \cite{DeepSDF}, etc). It is foreseen that there will continue to be a large number of advancing methods developed for continuous signal representation in vision and graphics. For a comprehensive review of related fields, we refer readers to \cite{NeRF_review,CG_SDF_survey}. 
\subsection{Scientific Computing}
Resolving scientific computing problems in numerical mathematics using continuous representations, particularly numerical PDE solutions, has attracted significant interests in recent years with the emerging of many physics-informed continuous representation methods \cite{NRP_PINN}. Resolving a numerical PDE can be viewed as a data reconstruction paradigm that recovers continuous physical fields from interpretable physical rules or limited observations. We mainly focus on the physics-informed neural network and neural operator learning frameworks.\par 
The physics-informed neural network (PINN) is a class of continuous representation models designed to solve forward and inverse problems involving PDEs by integrating physical laws. The PINN was introduced by Raissi et al. \cite{PINN} in 2019, which leverages neural networks to approximate solutions of PDEs.  Let a PDE be defined as
\[
\mathcal{N}[u(\mathbf{x}, t)] = 0, \;\mathbf{x} \in \Omega, \; t \in [0, T],
\]  
with boundary conditions $\mathcal{B}[u(\mathbf{x}, t)] = 0, \; \mathbf{x} \in \partial\Omega$, where \( \mathcal{N} \) is a differential operator, \( u \) is the unknown solution, and \( \Omega \) is the spatial domain. A PINN approximates \( u(\mathbf{x}, t) \) using a neural network \( \hat{u}_{\theta}(\mathbf{x}, t) \), parameterized by learnable weights \( \theta \). The network is trained to minimize a composite differential loss function containing several terms
$\mathcal{L}(\theta) = \mathcal{L}_{\text{data}} + \lambda_{1} \mathcal{L}_{\text{PDE}} + \lambda_{2} \mathcal{L}_{\text{BC}}$,
where the data fidelity loss (if labeled data exists) encodes the knowledge of observed data $
\mathcal{L}_{\text{data}} = \sum_{i} \left| \hat{u}_{\theta}(\mathbf{x}_i, t_i) - u(\mathbf{x}_i, t_i) \right|^2$, the PDE loss encodes the physical law $
\mathcal{L}_{\text{PDE}} =  \sum_{j} \left| \mathcal{N}[\hat{u}_{\theta}(\mathbf{x}_j, t_j)] \right|^2$, followed by the boundary condition loss $
\mathcal{L}_{\text{BC}} = \sum_{k} \left| \mathcal{B}[\hat{u}_{\theta}(\mathbf{x}_k, t_k)] \right|^2$. The differential loss functions related to \( \mathcal{N}[\hat{u}_{\theta}] \) and \( \mathcal{B}[\hat{u}_{\theta}] \) are computed using automatic differentiation of neural networks, enabling exact gradient calculations without discretization errors. PINNs have been successfully applied to several scientific fields including fluid dynamics, material modeling, biomedical systems, and climate science. By unifying data and physical laws, PINNs hold the power of physics-guided machine learning in computational science. Notably, the PINN can be seen as an INR \cite{PE,SIREN} operating on a physically interpretable continuous fields, with tailored differential loss functions to model the these physical fields. The development of PINNs is a broader topic coupling machine learning and numerical mathematics. For comprehensive discussions and reviews of the PINN framework and related methods, we refer readers to \cite{NRP_PINN,PINN_review,PINN_review_Fluids}.\par 
A growing body of recent research has investigated the application of grid encoding parametric models \cite{InstantNGP} for PINNs by leveraging the multi-resolution grid structures to effectively represent fine details in physical fields. Kang et al. \cite{PIXEL} proposed the physics-informed cell representations, which utilizes multi-resolution grid encoding \cite{InstantNGP} with a cosine interpolation kernel for PINNs. Ge et al. \cite{CAI_PINN_InstantNGP} proposed the hash grid encoding methods (based on InstantNGP) for PINN using differentiable cubic interpolation, which are combined directly with auto-differentiation of neural nets. Wang et al. \cite{AAAI_PINN_InstantNGP} proposed the neural physical simulation with multi-resolution hash grid encoding, and used a numerical gradient method for computing high-order derivatives with boundary conditions and a range
analysis sample method for fast neural geometry boundary. Huang and Alkhalifah \cite{InstantNGP_PINN_JCP} proposed the efficient PINN using hash encoding, which replace the automatic differentiation with finite-difference calculations of the derivatives to address the discontinuous. They also shared the appropriate ranges of hash encoding hyperparameters to
obtain robust derivatives. Recently, Kang et al. \cite{PIG} proposed the physics-informed Gaussians for PDE solutions, which utilize Gaussian mixture models as parametric mesh representations followed by a lightweight neural network, and the derivatives of such a structure can be computed analytically.\par
Neural operator (NO) learning \cite{DeepONet} serves as a novel architecture designed to learn mappings between infinite-dimensional function spaces, making it particularly suited for solving parametric PDEs. Unlike PINNs, NOs directly operate on functions and thus enjoy generalization abilities by evaluating PDE solutions instantly for new parameter functions. Let \( \mathcal{A} \) and \( \mathcal{U} \) be function spaces (e.g., Sobolev spaces). Given observations of input-output function pairs $\{a_j(x), u_j(y)\}_{j=1}^N, \;a_j \in \mathcal{A}, \, u_j \in \mathcal{U}$, the goal of NO learning is to learn an operator \( \mathcal{G}^\dagger: \mathcal{A} \to \mathcal{U} \) such that $\mathcal{G}^\dagger(a)(y) = u(y)$. For parametric PDEs, \( a(x) \) might represent an initial condition or coefficient field, and \( u(y) \) the PDE solution. The deep operator network (DeepONet) \cite{DeepONet} forms an operator by using a branch net to encode function values and a trunk net to decode locations, which enjoys universal approximation theorem in function spaces. The DeepONet is formulated as
\[
\mathcal{G}(a)(y) = \sum_{k=1}^p \underbrace{b_k(a)}_{\text{branch net}} \cdot \underbrace{t_k(y)}_{\text{trunk net}},
\]
where \( b_k \) encodes the input function, and \( t_k \) decodes the output location. Both \( b_k \) and \( t_k \) could be parameterized by neural nets, and learned through function pairs. The Fourier neural operator (FNO) \cite{FNO,NO} maps a function $v$ through Fourier transform operator and frequency-domain filter. The learnable Fourier operator layer is formulated as
\[
\mathcal{K}(a)(y) = \mathcal{F}^{-1}\left(R \cdot \mathcal{F}(a)\right)(y),
\]
where \( \mathcal{F} \) is the Fourier transform operator and \( R \) is a learnable frequency-domain filter. Stacking multiple Fourier layers leads to the FNO. As compared with DeepONet, FNO can query both input and output at any positions, and is discretization invariant. The FNO is also grounded with universal approximation theorem for operators \cite{NO}. Schiaffini et al. \cite{ICML_NO} proposed the localized integral and differential kernels in FNO to replace the global convolution kernel $R$, which better capture local features and preserve the resolution-independence. NOs represent a paradigm shift in scientific machine learning, combining numerical analysis with generalizable deep learning to resolve scientific computing problems (especially numerical PDEs) efficiently. For comprehensive discussion and review of NOs, please refer to \cite{NRP_NO}.\par 
Recent studies have explored accelerating techniques and addressing curse of dimensionality of PINNs and operator learning through tensor decomposition paradigms. Li and Ye \cite{D-FNO} proposed the decomposed FNO, which decomposes the high-dimensional latent representation into a series of rank-1 tensor products, and the three-dimensional fast Fourier transform (FFT) is replaced by a series of one-dimensional FFT. This decomposed structure enhances the efficiency for large-scale PDE modeling from $O(N^2\log N)$ to $O(NP\log N)$, where the separation rank $P$ is much smaller than the grid size $N$. Cho et al. \cite{Separable_PINN} proposed the separable network architecture for PINNs, which operates on a per-axis basis to significantly reduce the number of network propagations in multi-dimensional PDEs based on tensor decomposition, which shares similarity with tensor functional decomposition methods \cite{LRTFR,ECCV_Jianli}. A similar idea is used in \cite{FTD_PINN}, which uses a tensor decomposed structure (e.g., Tucker, CP, and tensor-train) to accelerate numerical PDE based on PINN for internal learning of a multivariate function. \par
Several studies have investigated operator learning frameworks for generating INR functions. Xu et al. \cite{SP_for_INR} proposed the signal processing paradigm for INR, which maps an INR to another INR by a learned operator parameterized by convolutional neural network. This method inputs the derivatives of an INR into the operator network and generates another INR. It is capable of dealing with several visual problems such as image denoising and classification. Pal et al. \cite{O-INR} proposed the INR via operator learning, which maps input functions (often in the form of Fourier positional encoding functions \cite{PE}) to the corresponding signals functions (characterized by conventional INRs that map coordinates to pixels). It achieves theoretical consistency with operator learning and obtains better performances for image regression and robustness in neural weight interpolation.\par																
Efforts to understand the convergence and generalization behaviors of PINN or NO have also been widely explored. For instance, Wang et al. \cite{JCP_PINN_NTK} analyzed the training dynamics of PINNs using NTK and proposed a novel NTK-guided gradient descent algorithm for PINN. Bonfanti et al. \cite{NeurIPS_nonlinear_PINN_NTK} analyzed the training dynamics of PINN for nonlinear PDEs under the NTK framework. Nguyen and Mucke \cite{NO_NTK} introduced the NTK regime for two-layer NOs and analyzed their generalization properties. Following this direction, further research into the theoretical properties of PINNs and NOs is expected to advance their theoretical foundations and improved algorithmic design. Recently, Dummer et al. \cite{SIAM_INR} proposed to use resolution-independent INR for large deformation diffeomorphic metric mapping (LDDMM) coupled with LDDMM-based statistical latent modeling with applications in computer graphics and the medical domain. This work paves the way for future research into how Riemannian geometry, shape analysis, and continuous learning representations can be combined.
\subsection{Medical Imaging and Bioinformatics}
In medical imaging and bioinformatics fields, continuous representation methods (particularly INRs) enable precise and resolution-independent representations of biological data that benefit many downstream tasks \cite{Medical_imaging_survey,NM_INR}. The inherent smoothness of continuous methods preserves interpretable structures in medical scans \cite{CT_CVPR}, cellular relationships \cite{STINR}, and tissue organizations \cite{GNTD} in biological specimens. The continuous modeling framework enables natural integration with physical rules, medical imaging processes, and biomedical engineering frameworks for enhanced reconstruction fidelity of diverse types of medical images and biological data.\par
Image registration is an important computational problem for medical imaging such as MRI and computed tomography (CT) imaging. The image registration aims to establish spatial correspondence between two or more images by applying nonlinear and elastic transformations. There is a group of works that utilize INRs to address the image registration. Wolterink et al. \cite{INR_registration_Jacobian_reg} proposed the first INR model for deformable image registration by predicting the transformation between images using an INR. Later, Byra et al. \cite{MICCAI_23_registration} proposed INRs for joint decomposition and registration of gene expression images, which use several implicit networks combined with an image exclusion loss to jointly perform the registration and decompose the image into a support and residual image, where the image decomposition guides the registration to be more accurate. Lampretsa et al. \cite{spline_INR_registration} proposed a spline-enhanced INR for multi-modal image registration, which parameterizes the continuous deformable transformation represented by an INR using free form deformations, allowing for multi-modal registration while mitigating folding issues. Zimmer et al. \cite{MICCAI_24_registration} proposed the generalized INR for image registration, which encodes the fixed and moving image volumes to latent representations, and uses the latent representations to modulate the INRs to enable generalization across multiple instances.\par
The medical imaging reconstruction using INRs is another widely studied research direction. For instance, Sun et al. \cite{CoIL} proposed a coordinate-based internal learning paradigm for continuous representation of measurements, with applications in inverse problems of medical imaging. Reed et al. \cite{CT_CVPR} proposed a dynamic CT reconstruction method from limited views with INRs coupled with parametric motion fields. Wu et al. \cite{MRI_INR_MICCAI} proposed to use INR for high-resolution MRI reconstruction, and further proposed an arbitrary scale super-resolution approach for 3D MRI \cite{MRI_LIIF} via 3D convolutional neural networks and INRs. Their recent work \cite{ICLR_MRI_25} proposed an InstantNGP-based model for reconstructing MRI and motion parameters from undersampled measurements. Chu et al. \cite{MIA_diffusion} proposed to integrate the prior sampling of a pre-trained diffusion model with INRs for accelerated MRI reconstruction, which incorporates both the diffusion prior and the MRI physical model to ensure high data fidelity.\par
Advanced model architecture designs were also incorporated into medical imaging processing using INRs. For instance, Vo et al. \cite{3DV_RED_INR} proposed a sparse-view X-ray CT method by using the regularization-by-denoising (RED) \cite{RED_SIIMS,RED_TCI} framework to regularize the INR optimization. It enhances the efficiency of INR training for this task by decoupling the post-processing network (RED network) and INR optimization. Similarly, Iskender et al. \cite{RED_NeRF} proposed a continuous neural field to represent the dynamic object with RED, via a learned restoration operator using static supervised training. Yu et al. \cite{Bilevel_INR_MRI} proposed a bilevel optimization framework for INR, with applications for accelerated MRI reconstruction. This method automatically optimizes the hyperparameters of the INR for a given protocol, enabling a tailored reconstruction without training data. For a comprehensive review of INR methods for medical imaging applications, we refer readers to \cite{Medical_imaging_survey}.\par 
\begin{figure*}[t]
	\scriptsize
	\setlength{\tabcolsep}{0.9pt}
	\begin{center}\vspace{0.3cm}
		\begin{tabular}{ccccc}\vspace{-0.1cm}
			\includegraphics[height=0.172\textwidth,width=0.15\textwidth]{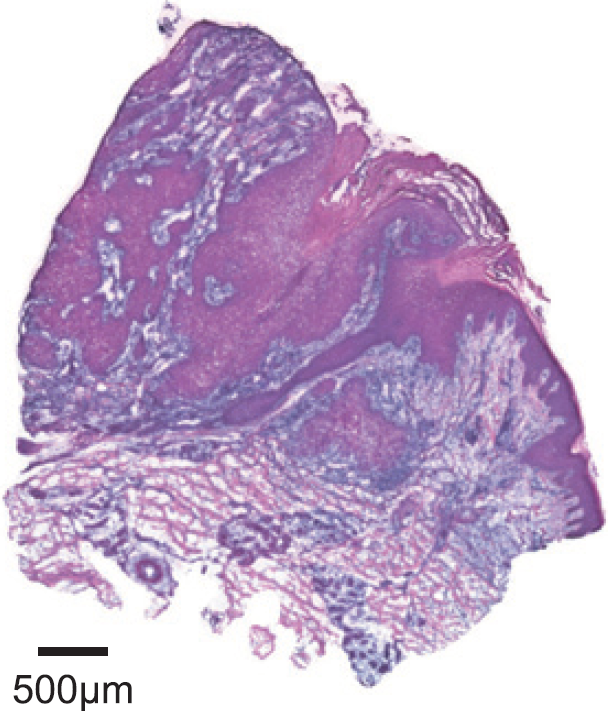}&
			\includegraphics[height=0.174\textwidth,width=0.15\textwidth]{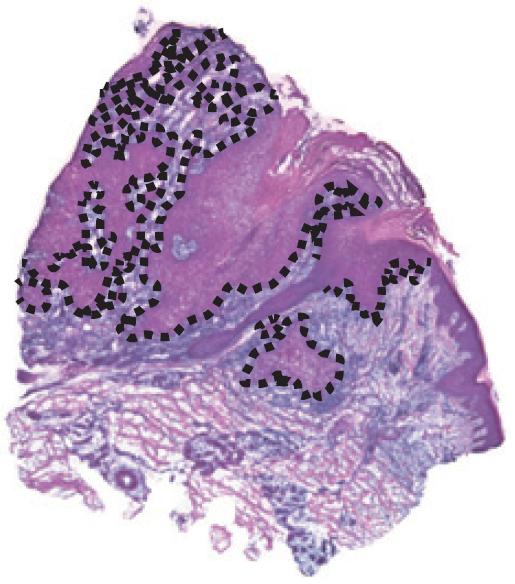}&
			\includegraphics[height=0.182\textwidth,width=0.16\textwidth]{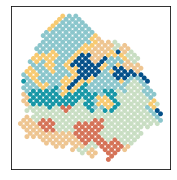}&
			\includegraphics[height=0.182\textwidth,width=0.16\textwidth]{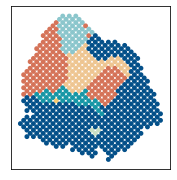}&
			\includegraphics[height=0.182\textwidth,width=0.205\textwidth]{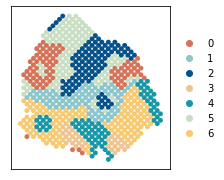}
			\\\vspace{-0.2cm}
			SCC tissue&Tumor leading edges&STAGATE&GraphST&\hspace{-0.6cm}{\bf STINR}\\
		\end{tabular}
		\vspace{-0.2cm}
	\end{center}
	\caption{An example of continuous representation methods for biological analysis of spatial transcriptomics data. The figures, taken from \cite{STINR}, show the spatial tissue segmentation results on the human squamous cell carcinoma (SCC) spatial transcriptomics. The continuous INR-based method STINR \cite{STINR} more accurately detects profiles and leading edges of tumor regions as compared with baselines STAGATE and GraphST.\label{fig:ST}}
\end{figure*}
Beyond medical imaging, continuous representation methods have also gained attraction in the field of bioinformatics, demonstrating their applicability across diverse computational biology applications. Zhong et al. \cite{cryo-EM} proposed the first neural network-based approach for cryo-electron microscopy to determine the structure of proteins under an inverse problem setting. The method leverages coordinate-based deep neural networks with positional encoding to model continuous generative factors of structural heterogeneity within proteins. Another application of continuous models in bioinformatics is the representation of spatial transcriptomic (ST) data \cite{NM_2021}. The ST is a novel technology that enables gene expression profiles within spatial positional context. This approach provides insights into tissue organization and cell-cell interactions in the space, advancing research developments in biology. However, the irregular spatial profile and variability of genes make it challenging to model ST data in a computational framework. There are several recent studies that utilize continuous representation to model the irregular and non-grid structure of ST data. For instance, Song et al. \cite{GNTD} proposed the graph-guided neural tensor decomposition model for reconstructing ST from incomplete measurements, where the model takes the coordinate and gene index as inputs and outputs the corresponding gene expression. The model is regularized by spatial and functional relations-informed graph regularizers. Chitra et al. \cite{NM_INR} proposed a coordinate-based neural network to learn a continuous and differentiable function of ST (termed GASTON), which enables gradient and isodepth analysis of ST to accurately identify spatial tissue domains. Li et al. \cite{Shihua_ST_INR_NAR} proposed STAGE, a computational framework that generates high-density ST from low-quality measurements by using a spatial coordinate-based generator. This spatial coordinate is generated from the observed ST in an encoder-decoder architecture. Recently, Luo et al. \cite{STINR} proposed an INR-based ST representation framework (termed STINR), which encodes ST along with single-cell references using an INR. Benefit from the implicit local correlation of INRs, STINR achieves higher accuracy in detecting spatial tissue domains from raw ST data (see {Fig. \ref{fig:ST}} for examples). Zhu et al. \cite{SUICA} introduced SUICA, which models ST in a continuous manner by INRs to improve both the spatial resolution and the gene expression. Overall, continuous representation frameworks, particularly INRs, have emerged as increasingly prominent computational approaches for ST analysis in bioinformatics. It is foreseeable that the continuous representation framework for ST will gain growing attention in future research.
\subsection{Geosciences and Remote Sensing}
Continuous function representation methods are increasingly emerging paradigms for geophysical and remote sensing data processing by addressing critical challenges in scenarios including hyperspectral imaging, seismic data analysis, and waveform inversion, by leveraging the inherent advantages of INRs such as parameter efficiency, resolution independence, and seamless integration with physical constraints.\par
The hyperspectral imaging (HSI) community has seen extensive research efforts focusing on leveraging continuous representation and INRs to efficiently encode high-dimensional HSI with hundreds of spectral bands. Zhang et al. \cite{HSI_INR_23} proposed to use INR for HSI super-resolution, where a content-aware hypernetwork is employed to produce INR weights for each HSI in a meta-learning manner. Chen et al. \cite{HSI_INR_TCSVT} proposed a spectral-wise INR method for coded aperture snapshot spectral imaging reconstruction, which can reconstruct an unlimited number of spectral bands. Ge et al. \cite{AAAI24_pan} proposed an INR method with progressive high-frequency reconstruction for pan-sharpening by effectively representing and fusing spatial and spectral features in the continuous domain. Chen et al. \cite{HSI_LIIF_24} proposed an arbitrary-scale HSI super-resolution from a fusion perspective with spatial priors. This method utilizes an RGB encoder trained with high-quality RGB datasets to extract spatial features, and utilizes a spectral transformer for feature fusion, enabling more accurate reconstruction. Liang et al. \cite{HSI_fusion_24} proposed a Fourier-enhanced INR fusion network for multispectral and
hyperspectral image fusion, which utilizes a spatial and frequency implicit fusion function in the Fourier domain to capture high-frequency information and expand the receptive field. Recently, Wang et al. \cite{HSI_Unmixing_INR} proposed a non-negative matrix function factorization framework parameterized by INRs for HSI unmixing, which could handle nonuniform spectral sampling due to the continuous representation.\par 
\begin{figure*}[t]
	\scriptsize
	\setlength{\tabcolsep}{0.9pt}
	\begin{center}\vspace{0.3cm}
		\begin{tabular}{c}\vspace{-0.1cm}
			\includegraphics[height=0.35\textwidth,width=0.82\textwidth]{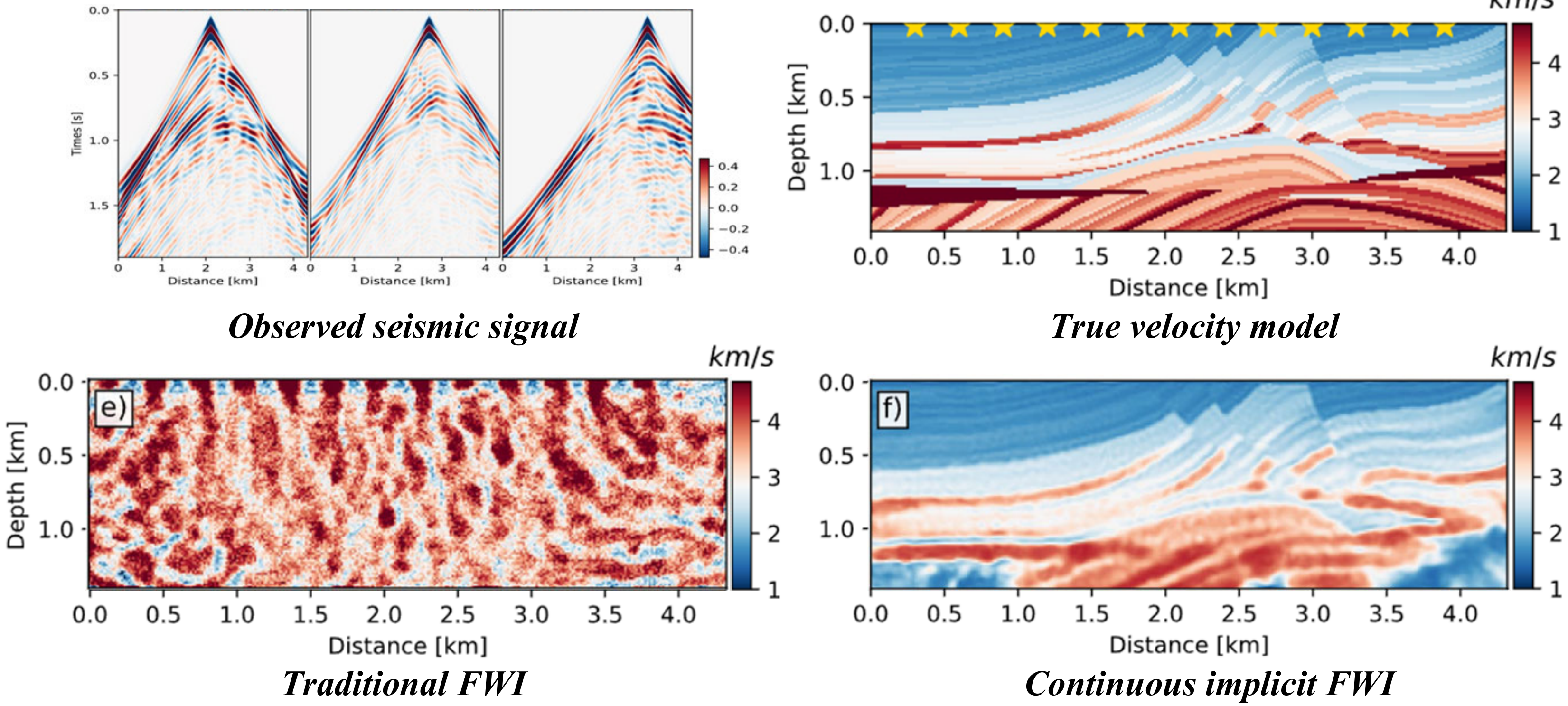}\\
		\end{tabular}
		\vspace{-0.4cm}
	\end{center}
	\caption{An example of continuous representation methods for full waveform inversion (FWI) of P-wave velocity model from observed geophysical seismic signals. The figures, taken from \cite{IFWI_JGR}, show the inversion results on the Marmousi model. As compared with traditional FWI, the implicit FWI \cite{IFWI_JGR} using INR shows robustness capability even with a random initial model.\label{fig:FWI}}
\end{figure*}
INRs have also been considered in geophysical data processing, such as seismic data. Gao et al. \cite{5D_seismic_INR} proposed to use INR for 5D seismic data interpolation, which encodes continuous seismic 5D wavefields and infers missing trace amplitudes from their coordinates. Recently, Wang et al. \cite{KAN_NLRR} proposed an efficient Kolmogorov-Arnold network-empowered neural low-rank representation for seismic denoising, where the model is parameterized by INRs to encode smoothness. The full waveform inversion (FWI) is another important technique in geophysics, which aims to reconstruct the underlying subsurface velocity model from observed seismic data, utilizing various waveforms. Sun et al. \cite{IFWI_JGR} proposed the implicit seismic FWI framework by using INR to parameterize the velocity model, and optimize the object  
\[ \min_\Theta\| \mathbf{d}_{\text{obs}} - \mathbf{F}(\mathbf{N}_{\Theta}(\mathbf{x})) \|^2, \]
where $\mathbf{N}_{\Theta}(\mathbf{x})$ is the predicted velocity model parameterized by an INR $\mathbf{N}_{\Theta}$ with input coordinates $\bf x$, ${\bf d}_{\text{obs}}$ is the observed seismic signal, and $\bf F$ denotes the forward modeling operator for wave propagation. The forward operator $\bf F$ can be expressed by an acoustic second-order finite difference operator to solve a numerical wave propagation PDE
\[
\frac{\partial^2 \mu(x, t)}{\partial t^2} = {\bf m}(x)^2 \nabla^2 \mu + f(x, t),
\]
where \( \mu(x, t) \) is the reconstructed waveform, \( {\bf m}(x) \) is the velocity model obtained by the INR $\mathbf{N}_{\Theta}(\mathbf{x})$, and \( f(x, t) \) is an initial condition. It is shown that the low-frequency preference of INR benefits a more robust and accurate inversion of velocity model under the FWI framework (see {Fig. \ref{fig:FWI} for examples}). Their later work \cite{IFWI_TGRS} applies an INR with GELU activation function to enhance the robustness of implicit FWI with the absence of prior and low-frequency information (such as noisy cases), further enhancing the practical value and robustness of of implicit FWI. \par
The inherent advantages of continuous representations (e.g., parameter efficiency and resolution independence) make them particularly suited for geoscience applications where data acquisition is costly, sparse, or contain noise. These advantages would position continuous methods as transformative tools for important geophysical applications such as multi-parameter inversion, 3D inversion, and uncertainty quantification in Earth system modeling \cite{earth} in future developments. 
\section{Future Directions and Perspectives}\label{sec:future}
We have delivered a comprehensive overview of continuous representation methods, theories, and applications. Notably, our analysis reveals substantial cross-pollination interplay between different subareas within the introduced works. For instance, the interplay between neural representation learning and computational mathematics reveals interdisciplinary parallels, e.g., the arbitrary-scale imaging paradigm \cite{LIIF} shares structural parallels with operator learning frameworks \cite{NO,FNO} from numerical PDE analysis, as both architectures map low-dimensional positional embeddings (augmented with latent codes/input function responses) to continuous fields through learned operators. Similarly, the tensor factorization paradigms underlying radiance field reconstruction (e.g., TensorRF \cite{TensorNeRF}) demonstrate mathematical similarities with tensor decomposed functional representations \cite{LRTFR,FunTT,Functional_CP_Longitudinal} developed for high-dimensional data approximation. Classical Fourier basis function approximations \cite{TSP_CP,ICML_STD} are similar to more recent neural representations employing frequency domain feature embeddings \cite{PE,NeRF}. These facts lead to many opportunities to incorporate interdisciplinary research directions, highlighting synergies between machine learning and computational mathematics primitives. Inspired by these intrinsic connections, we propose potential research directions, perspectives, and identify key challenges in developing advancing continuous representation methods and theories for data reconstruction.
\subsection{Future Directions on Continuous Representation Parametric Models}
Future research may explore the interdisciplinary integration of diverse parametric models to enhance performance and flexibility of continuous representations. For instance, we can combine statistical modeling techniques with neural representations to better capture periodic and regular patterns in temporal and time-series data. An example was already given in a recent work\cite{Shikai_ICLR_24}, where the neural representation coupled with statistical modeling shows better performances for tensor recovery. It is promising to incorporate basis function representations into neural frameworks to improve interpretability (e.g., model identifiability), or leverage randomized neural networks \cite{Wang_JCP} under least-squares frameworks to enhance INR efficiency. It would be beneficial to integrate advanced architectures—such as convolutional layers, attention mechanisms, transformers, mamba networks, and mixture-of-experts (MoE) models \cite{MoE_INR}—into neural continuous representations for complex pattern modeling. Integrating continuous representations with generative diffusion priors \cite{InverseBench,MIA_diffusion} for more expressive modeling and generation of continuous signals is also an interesting direction.\par 
Though some works have combined INR with tensor decompositions \cite{LRTFR,DRO-TFF,Separable_PINN}, it would be valuable to further combine INR with more stochastic tensor network decompositions (e.g., tensor-train or tensor-ring), and further learning adaptive tensor network decompositions that automatically determine factor connections through INR's intrinsic knowledge-sharing properties. It is also interesting to explore the integration of tensor decomposition with grid-based parametric models (e.g., InstantNGP \cite{InstantNGP}) to reduce grid scales and improve computational efficiency. In terms of frequency characterization, we can investigate frequency-aware or frequency-decoupled continuous representation models to enhance multi-frequency modeling abilities for real-world signals.\par 
For image super-resolution, we can develop tensor decomposition-based methodologies to lower the computational complexity of arbitrary-scale super-resolution \cite{LIIF} for large-scale images. It is interesting to further embed physics-informed constraints into INRs and tensor functional networks to learn physically consistent continuous fields while enhancing interpretability. Exploring more in-depth meta-learning strategies to enable the generalization of INRs across datasets, or using continuous representations as meta-learners to enable model generation are also promising future directions. Furthermore, it would be interesting to investigate model averaging or fusion strategies (e.g., hybridizing Tucker \cite{LRTFR} and CP \cite{TSP_CP} functional models) to adaptively select optimal parametric representations for different samples.
\subsection{Future Directions on Continuous Structural Modeling Methods}
Imposing explicit or implicit structural constraints and regularizations for continuous representations are effective for enhancing robustness and interpretability. In future research, we can develop online optimization techniques for continuous models for streaming data analysis \cite{streaming_survey}, with the focus on regularization strategies that can handle irregular data streams, maintain physical consistency of dynamically updated continuous models, and avoid catastrophic forgetting during online optimization.\par 
It is required to generalize existing continuous regularizations to more flexible formulations, such as weighted low-rank factorization models or group sparsity-based continuous regularizations, to boost performances. It is also crucial to extend current continuous regularizations to high-dimensional cases to enhance scalability, such as extending the neural TV \cite{NeurTV} to characterize multi-directional local correlations. We can also attempt implementing these continuous regularization methods for various physical models, such as neural radiance fields, signed distance function fields, and Bayesian statistical frameworks, to enhance structural consistence in real-world applications.\par
Another potential is to explore continuous neural regularizations for implicit low-rank adaptation (LoRA) \cite{LoRA} in large model fine-tuning, where the low-rank weights can be parameterized via continuous representations to enhance the efficiency of fine-tuning large language models. We can further extend the fine-tuning approach to class-incremental continual learning paradigms \cite{SD-LoRA}, by employing implicit low-rank function representations to dynamically optimize INR weights for task-adaptive low-rank adapters.\par 
Current tensor function decomposition methods for INRs \cite{CoordX,Separable_PINN,LRTFR} are inefficient when handling irregular non-grid data. Hence, addressing the efficiency challenges of continuous tensor decomposition methods \cite{CoordX,LRTFR} when processing irregularly sampled data (non-tensor formats) is also important and challenging. This challenge can be resolved through potential integration with grid encoding-based approaches \cite{InstantNGP} or voxel projection methods, which transform non-grid structured data to girds or voxels to enable efficient processing.
\subsection{Future Directions on Theoretical Analysis for Continuous Methods}
Future research on continuous representation methods offers several promising theoretical directions worthy of exploration. We outline some directions to inspire future research on understanding the theoretical effectiveness of continuous methods. For instance, we can investigate the theoretical properties of neural network-based scientific computing frameworks (e.g., PINNs and neural operators) based on well-established mathematical tools from machine learning, such as implicit regularization and NTK theory. We can study the approximation and representation theories of INR-based matrix/tensor function decomposition models \cite{LRTFR,Separable_PINN,CoordX}, such as low-rank approximation error in the function space and the decay rate of kernel singular values. Furthermore, we can analyze the implicit bias or optimization dynamics induced by explicit continuous regularizations (e.g., neural TV \cite{NeurTV}, Jacobian regularization \cite{INR_registration_Jacobian_reg}, or neural nonlocal regularization \cite{CRNL}) beyond their explicit formulations. To exploit this, mathematical tools from NTK-based analyses of differential loss functions in PINNs \cite{NeurIPS_nonlinear_PINN_NTK} could provide inspirations.\par 
It is also important to characterize the implicit bias of optimization methods for continuous models under more practical settings, such as mini-batch algorithms (critical for computationally expensive applications \cite{MRI_INR_MICCAI}) instead of full-batch optimization (suitable for efficient representations like tensor decompositions \cite{LRTFR,CoordX}). Understanding the joint influence of both model architectures and optimization strategies is also important to uncover the scalability of continuous representation methods. The joint influence might be tractable by coupling convergence analysis \cite{NTK} with implicit bias characterization \cite{Implicit_regularization_NIPS_2019}.\par
With insightful theoretical developments, it is important to develop theoretical-algorithmic co-design, such as using explicit regularizations to counter the implicit and spectral bias \cite{Implicit_reg_CNN}, developing spectral bias-aware activation functions and rank-adaptive tensor function formats, to establish continuous methods that are both theoretically sound and computationally efficient.\par
Although implicit regularizations for linear matrix/tensor factorizations have been well-studied \cite{Implicit_regularization_NIPS_2019,Implicit_regularization_deep_CP}, establishing distinct implicit regularization theories induced by continuous representation models with smooth inputs (e.g., coordinate inputs in INR frameworks) remains unexplored and challenging. It is important to distinguish continuous representations from conventional deep matrix factorization models \cite{Implicit_regularization_NIPS_2019} to explain their practical performance gap and reveal unique properties of INR-based approaches brought by the continuous paradigm. Also, to interpret the generalization capabilities of continuous representation methods, it is required to incorporate data-dependent and problem-dependent generalization error bounds \cite{Li_bound} to theoretically assess the cross-domain and multi-modal modeling capabilities of advanced INR methods, such as the generalization error analysis of INR for full waveform inversion \cite{IFWI_JGR}, arbitrary-scale imaging \cite{LIIF}, and view synthesis \cite{NeRF}.\par 
Finally, studying the implicit bias from the untrained neural network perspective warrants discussions. For instance, we can consider the double over-parameterization paradigm \cite{YiMa_DIP} (over-parameterize both the low-rank matrix and sparse corruption) using implicit low-rank neural representations, and potentially enhance current results using discrete CNNs \cite{YiMa_DIP}. We can also investigate the implicit smoothness bias \cite{ICML_untrained} of untrained continuous representation neural networks, and explore the discrepancy principle for early stopping of untrained neural networks \cite{SIAM_untrained} to enable stable convergence of continuous representations from randomly initialized neural networks, particularly addressing the initialization sensitivity challenge.
\subsection{Future Directions on Applications of Continuous Methods}
Extending existing continuous representation methods and theoretical frameworks to diverse data representation and reconstruction applications opens up numerous cross-disciplinary research opportunities, warranting further investigation. For example, we can utilize advanced operator learning methods for diverse low-level vision tasks, such as the continuous image super-resolution, or conversely, employ grid-encoding-based INRs \cite{InstantNGP} to solve numerical PDEs and related scientific computing problems. We can also apply the continuous regularization to address the image restoration challenge, such as deraining \cite{SCIS_rain,INR_derain}, where real-world rain streaks often display explicit or implicit continuous patterns.\par
In AI for science, INRs \cite{SIREN,PE} and low-rank function representations \cite{CoordX,LRTFR} offer strong potential for processing high-velocity, large-scale real-world data, such as black hole imaging \cite{InverseBench}, infrared imaging \cite{NeurSTT}, and light-field tensor analysis \cite{ICCV_LF}. These real-world tensor datasets inherently possess local and global structural correlations, which can be efficiently encoded using continuous representation methods.\par 
Future researches may integrate computational and interpretability analysis tools—such as gradient-based saliency maps and dimension reduction methods—to balance interpretability with model complexity and identify key features of INRs, thus avoiding redundancy in continuous representation learning. Furthermore, we can leverage pruning, fine-tuning, or adaptive neural architecture search to reduce the model size of continuous representations, thereby improving efficiency.\par 
There are still key challenges in continuous methods for real-world applications, including scalability to high-dimensional data, computational burden, spectral bias toward low-frequency components, and out-of-distribution generalization. Further research on scalability and efficiency is crucial to advancing real-world applications. It is also important to further investigate cross-modal generalization, such as integrating cross-modal priors (e.g., fusing hematoxylin and eosin (H\&E) stain imaging \cite{HE_NC} with gene expression data \cite{STINR,NM_INR} under a continuous representation framework to link morphological features with transcriptomics). Exploring unified continuous representation frameworks for heterogeneous data fusion (e.g., hyperspectral and multispectral images, MRI and genomics, infrared and RGB images) are also promising research directions.\par
In summary, future research could consider both theory-guided and application-oriented developments, and couple mathematical analysis with algorithm improvements to enable continuous representation methods for interdisciplinary scientific discovery and applications. It is required to develop deeper collaborations between machine learning, applied mathematical, and domain scientists, with innovative modeling techniques to promote advanced data reconstruction paradigms, therefore addressing multi-modal, multi-scale, and physics-induced data science problems in the era of big data.
\section{Conclusions}
Continuous representation methods have achieved significant progress in data representation and reconstruction. This review synthesizes recent research advancements across three dimensions: methodological designs, theoretical foundations, and practical applications. Future studies could prioritize interdisciplinary collaboration, combining advances in neural architectures with domain knowledge (e.g., physical principles) to enhance continuous representation efficiency and real-world applicability. Additionally, integrating these methods with emerging technologies—such as foundation models, multi-modal models, and large language models—could lead to new breakthroughs in continuous data reconstruction paradigms. It is essential to establish closer collaboration among machine learning researchers, applied mathematicians, and domain experts. Such efforts would improve existing techniques and provide new transformative computational tools based on current continuous family methods, benefiting scientific discovery for broader research fields.
\bibliographystyle{plain}
\bibliography{ref}

\begin{thebibliography}{100}

\bibitem{streaming_survey}
K.~Abed-Meraim, N.~L. Trung, Adel Hafiane, et~al.
\newblock A contemporary and comprehensive survey on streaming tensor
  decomposition.
\newblock {\em IEEE Transactions on Knowledge and Data Engineering},
  35(11):10897--10921, 2022.

\bibitem{untrained_survey_2025}
Ismail Alkhouri, Evan Bell, Avrajit Ghosh, Shijun Liang, Rongrong Wang, and
  Saiprasad Ravishankar.
\newblock Understanding untrained deep models for inverse problems: Algorithms
  and theory.
\newblock {\em Arxiv: 2502.18612}, 2025.

\bibitem{Implicit_regularization_NIPS_2019}
Sanjeev Arora, Nadav Cohen, Wei Hu, and Yuping Luo.
\newblock Implicit regularization in deep matrix factorization.
\newblock In {\em Proceedings of the 33rd International Conference on Neural
  Information Processing Systems (NeurIPS)}, pages 7413--7424, 2019.

\bibitem{Fine_grained_NTK}
Sanjeev Arora, Simon Du, Wei Hu, Zhiyuan Li, and Ruosong Wang.
\newblock Fine-grained analysis of optimization and generalization for
  overparameterized two-layer neural networks.
\newblock In {\em Proceedings of the 36th International Conference on Machine
  Learning (ICML)}, volume~97, pages 322--332, 2019.

\bibitem{wideNN_NTK}
Sanjeev Arora, Simon~S. Du, Wei Hu, Zhiyuan Li, Ruslan Salakhutdinov, and
  Ruosong Wang.
\newblock On exact computation with an infinitely wide neural net.
\newblock In {\em Proceedings of the 33rd International Conference on Neural
  Information Processing Systems (NeurIPS)}, pages 8141--8150, 2019.

\bibitem{InstantNGP_NTK}
Samuel Audia, Soheil Feizi, Matthias Zwicker, and Dinesh Manocha.
\newblock How learnable grids recover fine detail in low dimensions: A neural
  tangent kernel analysis of multigrid parametric encodings.
\newblock In {\em The Thirteenth International Conference on Learning
  Representations (ICLR)}, 2025.

\bibitem{NRP_NO}
Kamyar Azizzadenesheli, Nikola Kovachki, Zongyi Li, Miguel Liu-Schiaffini, Jean
  Kossaifi, and Anima Anandkumar.
\newblock Neural operators for accelerating scientific simulations and design.
\newblock {\em Nature Reviews Physics}, 6:320--328, 2024.

\bibitem{NeurIPS24_IR}
Zhiwei Bai, Jiajie Zhao, and Yaoyu Zhang.
\newblock Connectivity shapes implicit regularization in matrix factorization
  models for matrix completion.
\newblock In {\em The Thirty-eighth Annual Conference on Neural Information
  Processing Systems}, volume~37, pages 45914--45955, 2024.

\bibitem{NeuralExperts}
Yizhak Ben-Shabat, Chamin Hewa~Koneputugodage, Sameera Ramasinghe, and Stephen
  Gould.
\newblock Neural experts: Mixture of experts for implicit neural
  representations.
\newblock In {\em Advances in Neural Information Processing Systems},
  volume~37, pages 101641--101670, 2024.

\bibitem{NeurIPS_nonlinear_PINN_NTK}
Andrea Bonfanti, Giuseppe Bruno, and Cristina Cipriani.
\newblock The challenges of the nonlinear regime for physics-informed neural
  networks.
\newblock In {\em Advances in Neural Information Processing Systems},
  volume~37, pages 41852--41881, 2024.

\bibitem{MICCAI_23_registration}
Michal Byra, Charissa Poon, Tomomi Shimogori, and Henrik Skibbe.
\newblock Implicit neural representations for joint decomposition
  and registration of gene expression images in the marmoset brain.
\newblock In {\em Medical Image Computing and Computer Assisted Intervention
  (MICCAI)}, pages 645--654, 2023.

\bibitem{Batch_normalization_INR}
Zhicheng Cai, Hao Zhu, Qiu Shen, Xinran Wang, and Xun Cao.
\newblock Batch normalization alleviates the spectral bias in coordinate
  networks.
\newblock In {\em 2024 IEEE/CVF Conference on Computer Vision and Pattern
  Recognition (CVPR)}, pages 25160--25171, 2024.

\bibitem{TensorNeRF}
Anpei Chen, Zexiang Xu, Andreas Geiger, Jingyi Yu, and Hao Su.
\newblock Tensorf: Tensorial radiance fields.
\newblock In {\em 17th European Conference on Computer Vision (ECCV)}, page
  333–350, 2022.

\bibitem{HSI_LIIF_24}
Guochao Chen, Jiangtao Nie, Wei Wei, Lei Zhang, and Yanning Zhang.
\newblock Arbitrary-scale hyperspectral image super-resolution from a fusion
  perspective with spatial priors.
\newblock {\em IEEE Transactions on Geoscience and Remote Sensing}, 62:1--11,
  2024.

\bibitem{NeRV}
Hao Chen, Bo~He, Hanyu Wang, Yixuan Ren, Ser~Nam Lim, and Abhinav Shrivastava.
\newblock {NeRV}: Neural representations for videos.
\newblock In {\em Advances in Neural Information Processing Systems},
  volume~34, pages 21557--21568, 2021.

\bibitem{HSI_INR_TCSVT}
Huan Chen, Wangcai Zhao, Tingfa Xu, Guokai Shi, Shiyun Zhou, Peifu Liu, and
  Jianan Li.
\newblock Spectral-wise implicit neural representation for hyperspectral image
  reconstruction.
\newblock {\em IEEE Transactions on Circuits and Systems for Video Technology},
  34(5):3714--3727, 2024.

\bibitem{Shikai_25_LRTFR}
Panqi Chen, Lei Cheng, Jianlong Li, Weichang Li, Weiqing Liu, Jiang Bian, and
  Shikai Fang.
\newblock Generalized temporal tensor decomposition with rank-revealing
  latent-ode.
\newblock {\em Arxiv: 2502.06164}, 2025.

\bibitem{ICML_feature_function_matrix}
Tianqi Chen, Hang Li, Qiang Yang, and Yong Yu.
\newblock General functional matrix factorization using gradient boosting.
\newblock In {\em Proceedings of the 30th International Conference on
  International Conference on Machine Learning (ICML)}, page 436–444, 2013.

\bibitem{INR_derain}
Xiang Chen, Jinshan Pan, and Jiangxin Dong.
\newblock Bidirectional multi-scale implicit neural representations for image
  deraining.
\newblock In {\em 2024 IEEE/CVF Conference on Computer Vision and Pattern
  Recognition (CVPR)}, pages 25627--25636, 2024.

\bibitem{LIIF}
Yinbo Chen, Sifei Liu, and Xiaolong Wang.
\newblock Learning continuous image representation with local implicit image
  function.
\newblock In {\em 2021 IEEE/CVF Conference on Computer Vision and Pattern
  Recognition}, pages 8624--8634, 2021.

\bibitem{meta_transformer_INR}
Yinbo Chen and Xiaolong Wang.
\newblock Transformers as meta-learners for implicit neural representations.
\newblock In {\em European Conference on Computer Vision (ECCV)}, pages
  170--187, 2022.

\bibitem{VideoINR}
Zeyuan Chen, Yinbo Chen, Jingwen Liu, Xingqian Xu, Vidit Goel, Zhangyang Wang,
  Humphrey Shi, and Xiaolong Wang.
\newblock Videoinr: Learning video implicit neural representation for
  continuous space-time super-resolution.
\newblock In {\em 2022 IEEE/CVF Conference on Computer Vision and Pattern
  Recognition (CVPR)}, pages 2037--2047, 2022.

\bibitem{CVPR19_shape}
Zhiqin Chen and Hao Zhang.
\newblock Learning implicit fields for generative shape modeling.
\newblock In {\em 2019 IEEE/CVF Conference on Computer Vision and Pattern
  Recognition (CVPR)}, pages 5932--5941, 2019.

\bibitem{NTK_two_layer_noisyGD}
Zixiang Chen, Yuan Cao, Quanquan Gu, and Tong Zhang.
\newblock A generalized neural tangent kernel analysis for two-layer neural
  networks.
\newblock In {\em Proceedings of the 34th International Conference on Neural
  Information Processing Systems (NeurIPS)}, pages 13363--13373, 2020.

\bibitem{RKHS_norm_reg}
Zonghao Chen, Xupeng Shi, Tim G.~J. Rudner, Qixuan Feng, Weizhong Zhang, and
  Tong Zhang.
\newblock A neural tangent kernel perspective on function-space regularization
  in neural networks.
\newblock In {\em International Conference on Neural Information Processing
  Systems (NeurIPS) Workshop on Optimization for Machine Learning}, 2022.

\bibitem{INR_SAR_detection}
Ziheng Cheng, Yucheng Ding, Chunhui Qu, and Bo~Chen.
\newblock Implicit neural representation with imaging geometry for sar target
  recognition.
\newblock {\em IEEE Transactions on Aerospace and Electronic Systems}, 2025.
\newblock doi={10.1109/TAES.2025.3538571}.

\bibitem{CVPR20_Shape}
Julian Chibane, Thiemo Alldieck, and Gerard Pons-Moll.
\newblock Implicit functions in feature space for 3d shape reconstruction and
  completion.
\newblock In {\em 2020 IEEE/CVF Conference on Computer Vision and Pattern
  Recognition (CVPR)}, pages 6968--6979, 2020.

\bibitem{NM_INR}
Uthsav Chitra, Brian~J. Arnold, Hirak Sarkar, Kohei Sanno, Cong Ma, Sereno
  Lopez-Darwin, and Benjamin~J. Raphael.
\newblock Mapping the topography of spatial gene expression with interpretable
  deep learning.
\newblock {\em Nature Methods}, 22:298--309, 2025.

\bibitem{INR_precondition}
Shin-Fang Chng, Hemanth Saratchandran, and Simon Lucey.
\newblock Preconditioners for the stochastic training of neural fields.
\newblock In {\em Proceedings of the IEEE/CVF Conference on Computer Vision and
  Pattern Recognition (CVPR)}, 2025.

\bibitem{Separable_PINN}
Junwoo Cho, Seungtae Nam, Hyunmo Yang, Seok-Bae Yun, Youngjoon Hong, and
  Eunbyung Park.
\newblock Separable physics-informed neural networks.
\newblock In {\em Thirty-seventh Conference on Neural Information Processing
  Systems}, pages 23761--23788, 2023.

\bibitem{MIA_diffusion}
Jiayue Chu, Chenhe Du, Xiyue Lin, Xiaoqun Zhang, Lihui Wang, Yuyao Zhang, and
  Hongjiang Wei.
\newblock Highly accelerated mri via implicit neural representation guided
  posterior sampling of diffusion models.
\newblock {\em Medical Image Analysis}, 100:103398, 2025.

\bibitem{RED_SIIMS}
Regev Cohen, Michael Elad, and Peyman Milanfar.
\newblock Regularization by denoising via fixed-point projection ({RED-PRO}).
\newblock {\em SIAM Journal on Imaging Sciences}, 14(3):1374--1406, 2021.

\bibitem{MRA}
Ingrid Daubechies.
\newblock Orthonormal bases of compactly supported wavelets.
\newblock {\em Communications on Pure and Applied Mathematics}, 41(7):909--996,
  1988.

\bibitem{spl_smooth}
Otto Debals, Marc Van~Barel, and Lieven De~Lathauwer.
\newblock Nonnegative matrix factorization using nonnegative polynomial
  approximations.
\newblock {\em IEEE Signal Processing Letters}, 24(7):948--952, 2017.

\bibitem{IFWI_TGRS}
Bo~Du, Jian Sun, Anqi Jia, Ning Wang, and Huaishan Liu.
\newblock Physics-informed robust and implicit full waveform inversion without
  prior and low-frequency information.
\newblock {\em IEEE Transactions on Geoscience and Remote Sensing}, 62:1--12,
  2024.

\bibitem{SIAM_INR}
Sven Dummer, Nicola Strisciuglio, and Christoph Brune.
\newblock Rda-inr: Riemannian diffeomorphic autoencoding via implicit neural
  representations.
\newblock {\em SIAM Journal on Imaging Sciences}, 17(4):2302--2330, 2024.

\bibitem{SIIMS_denoising_survey}
Michael Elad, Bahjat Kawar, and Gregory Vaksman.
\newblock Image denoising: The deep learning revolution and beyond—a survey
  paper.
\newblock {\em SIAM Journal on Imaging Sciences}, 16(3):1594--1654, 2023.

\bibitem{Universal_Ellacott_1994}
S.W. Ellacott.
\newblock Aspects of the numerical analysis of neural networks.
\newblock {\em Acta Numerica}, 3:145–202, 1994.

\bibitem{Season_trend}
Julien Fageot.
\newblock Variational seasonal-trend decomposition with sparse
  continuous-domain regularization.
\newblock {\em arXiv}, 2505.10486, 2025.

\bibitem{Shikai_ICML_24}
Shikai Fang, Qingsong Wen, Yingtao Luo, Shandian Zhe, and Liang Sun.
\newblock {B}ay{OTIDE}: {B}ayesian online multivariate time series imputation
  with functional decomposition.
\newblock In {\em Proceedings of the 41st International Conference on Machine
  Learning (ICML)}, volume 235, pages 12993--13009, 2024.

\bibitem{NeurIPS_Fang}
Shikai Fang, Xin Yu, Shibo Li, Zheng Wang, Robert~M. Kirby, and Shandian Zhe.
\newblock Streaming factor trajectory learning for temporal tensor
  decomposition.
\newblock In {\em Proceedings of the 37th International Conference on Neural
  Information Processing Systems}, pages 56849--56870, 2023.

\bibitem{Shikai_ICLR_24}
Shikai Fang, Xin Yu, Zheng Wang, Shibo Li, Mike Kirby, and Shandian Zhe.
\newblock Functional bayesian tucker decomposition for continuous-indexed
  tensor data.
\newblock In {\em The Twelfth International Conference on Learning
  Representations (ICLR)}, 2024.

\bibitem{CycleINR}
Wei Fang, Yuxing Tang, Heng Guo, Mingze Yuan, Tony~C.W. Mok, Ke~Yan, Jiawen
  Yao, Xin Chen, Zaiyi Liu, Le~Lu, Ling Zhang, and Minfeng Xu.
\newblock Cycleinr: Cycle implicit neural representation for arbitrary-scale
  volumetric super-resolution of medical data.
\newblock In {\em 2024 IEEE/CVF Conference on Computer Vision and Pattern
  Recognition (CVPR)}, pages 11631--11641, 2024.

\bibitem{MFN}
Rizal Fathony, Anit~Kumar Sahu, Devin Willmott, and J~Zico Kolter.
\newblock Multiplicative filter networks.
\newblock In {\em International Conference on Learning Representations}, 2021.

\bibitem{ICCV_LF}
Brandon~Yushan Feng and Amitabh Varshney.
\newblock Signet: Efficient neural representation for light fields.
\newblock In {\em 2021 IEEE/CVF International Conference on Computer Vision
  (ICCV)}, pages 14204--14213, 2021.

\bibitem{Kplanes}
Sara Fridovich-Keil, Giacomo Meanti, Frederik~Rahbæk Warburg, Benjamin Recht,
  and Angjoo Kanazawa.
\newblock K-planes: Explicit radiance fields in space, time, and appearance.
\newblock In {\em 2023 IEEE/CVF Conference on Computer Vision and Pattern
  Recognition (CVPR)}, pages 12479--12488, 2023.

\bibitem{Plenoxels}
Sara Fridovich-Keil, Alex Yu, Matthew Tancik, Qinhong Chen, Benjamin Recht, and
  Angjoo Kanazawa.
\newblock Plenoxels: Radiance fields without neural networks.
\newblock In {\em 2022 IEEE/CVF Conference on Computer Vision and Pattern
  Recognition (CVPR)}, pages 5491--5500, 2022.

\bibitem{5D_seismic_INR}
Wenbin Gao, Dawei Liu, Wenchao Chen, Mauricio~D. Sacchi, and Xiaokai Wang.
\newblock Ne{RSI}: Neural implicit representations for 5d seismic data
  interpolation.
\newblock {\em Geophysics}, 90(1):V29--V42, 2025.

\bibitem{TMLR_NTK_control}
Amnon Geifman, Daniel Barzilai, Ronen Basri, and Meirav Galun.
\newblock Controlling the inductive bias of wide neural networks by modifying
  the kernel{\textquoteright}s spectrum.
\newblock {\em Transactions on Machine Learning Research}, 2024.

\bibitem{ICCV19_Shape}
Kyle Genova, Forrester Cole, Daniel Vlasic, Aaron Sarna, William Freeman, and
  Thomas Funkhouser.
\newblock Learning shape templates with structured implicit functions.
\newblock In {\em 2019 IEEE/CVF International Conference on Computer Vision
  (ICCV)}, pages 7153--7163, 2019.

\bibitem{FunTT}
Alex Gorodetsky, Sertac Karaman, and Youssef Marzouk.
\newblock A continuous analogue of the tensor-train decomposition.
\newblock {\em Computer Methods in Applied Mechanics and Engineering},
  347:59--84, 2019.

\bibitem{TT_regression}
Alex~A. Gorodetsky and John~D. Jakeman.
\newblock Gradient-based optimization for regression in the functional
  tensor-train format.
\newblock {\em Journal of Computational Physics}, 374:1219--1238, 2018.

\bibitem{Spectral_embedding_graph}
Daniele Grattarola and Pierre Vandergheynst.
\newblock Generalised implicit neural representations.
\newblock In {\em Proceedings of the 36th International Conference on Neural
  Information Processing Systems}, pages 30446--30458, 2022.

\bibitem{SIAM_Bivariate_function_SVD}
Michael Griebel and Guanglian Li.
\newblock On the decay rate of the singular values of bivariate functions.
\newblock {\em SIAM Journal on Numerical Analysis}, 56(2):974--993, 2018.

\bibitem{IR_NIPS17}
Suriya Gunasekar, Blake~E Woodworth, Srinadh Bhojanapalli, Behnam Neyshabur,
  and Nati Srebro.
\newblock Implicit regularization in matrix factorization.
\newblock In {\em Advances in Neural Information Processing Systems},
  volume~30, pages 6152--6160, 2017.

\bibitem{JASA_FunTSVD}
Rungang Han, Pixu Shi, and Anru R.~Zhang and.
\newblock Guaranteed functional tensor singular value decomposition.
\newblock {\em Journal of the American Statistical Association},
  119(546):995--1007, 2024.

\bibitem{MoE_INR}
Zekun Hao, Arun Mallya, Serge Belongie, and Ming-Yu Liu.
\newblock Implicit neural representations with levels-of-experts.
\newblock In {\em Advances in Neural Information Processing Systems}, pages
  2564--2576, 2022.

\bibitem{Implicit_regularization_deep_CP}
Kais Hariz, Hachem Kadri, Stephane Ayache, Maher Moakher, and Thierry Artieres.
\newblock Implicit regularization with polynomial growth in deep tensor
  factorization.
\newblock In {\em Proceedings of the 39th International Conference on Machine
  Learning (ICML)}, volume 162, pages 8484--8501, 2022.

\bibitem{Implicit_regularization_deep_Tucker}
Kais Hariz, Hachem Kadri, Stéphane Ayache, Maher Moakher, and Thierry
  Artières.
\newblock Implicit regularization in deep tucker factorization: Low-rankness
  via structured sparsity.
\newblock In {\em International Conference on Artificial Intelligence and
  Statistics (AISTATS)}, pages 2359--2367, 2024.

\bibitem{SIAM_cheb}
Behnam Hashemi and Lloyd~N. Trefethen.
\newblock Chebfun in three dimensions.
\newblock {\em SIAM Journal on Scientific Computing}, 39(5):C341--C363, 2017.

\bibitem{ICML_untrained}
Reinhard Heckel and Mahdi Soltanolkotabi.
\newblock Compressive sensing with un-trained neural networks: Gradient descent
  finds a smooth approximation.
\newblock In {\em Proceedings of the 37th International Conference on Machine
  Learning}, volume 119, pages 4149--4158, 2020.

\bibitem{ICLR_geometric}
Hyeongjun Heo, Seonghun Oh, Jae~Yong Lee, Young~Min Kim, and Yonghyeon Lee.
\newblock Isometric regularization for manifolds of functional data.
\newblock In {\em The Thirteenth International Conference on Learning
  Representations}, 2025.

\bibitem{GRSM_HSI}
Danfeng Hong, Wei He, Naoto Yokoya, Jing Yao, Lianru Gao, Liangpei Zhang,
  Jocelyn Chanussot, and Xiaoxiang Zhu.
\newblock Interpretable hyperspectral artificial intelligence: When nonconvex
  modeling meets hyperspectral remote sensing.
\newblock {\em IEEE Geoscience and Remote Sensing Magazine}, 9(2):52--87, 2021.

\bibitem{SCIS_rain}
Minghan Li Qian Zhao Deyu~Meng Hong~Wang, Yichen~Wu.
\newblock Survey on rain removal from videos or a single image.
\newblock {\em Science China Information Sciences}, 65(1):111101, 2022.

\bibitem{Universal_NN}
Kurt Hornik, Maxwell Stinchcombe, and Halbert White.
\newblock Multilayer feedforward networks are universal approximators.
\newblock {\em Neural Networks}, 2(5):359--366, 1989.

\bibitem{LoRA}
Edward~J Hu, yelong shen, Phillip Wallis, Zeyuan Allen-Zhu, Yuanzhi Li, Shean
  Wang, Lu~Wang, and Weizhu Chen.
\newblock Lo{RA}: Low-rank adaptation of large language models.
\newblock In {\em International Conference on Learning Representations (ICLR)},
  2022.

\bibitem{PINN_review}
Haoteng Hu, Lehua Qi, and Xujiang Chao.
\newblock Physics-informed neural networks ({PINN}) for computational solid
  mechanics: Numerical frameworks and applications.
\newblock {\em Thin-Walled Structures}, 205:112495, 2024.

\bibitem{Heatmap_INR_NIPS}
Shengxiang Hu, Huaijiang Sun, Dong Wei, Xiaoning Sun, and Jin Wang.
\newblock Continuous heatmap regression for pose estimation via implicit neural
  representation.
\newblock In {\em The Thirty-eighth Annual Conference on Neural Information
  Processing Systems (NeurIPS)}, volume~37, pages 102036--102055, 2024.

\bibitem{InstantNGP_PINN_JCP}
Xinquan Huang and Tariq Alkhalifah.
\newblock Efficient physics-informed neural networks using hash encoding.
\newblock {\em Journal of Computational Physics}, 501:112760, 2024.

\bibitem{ICML_STD}
Masaaki Imaizumi and Kohei Hayashi.
\newblock Tensor decomposition with smoothness.
\newblock In {\em International Conference on Machine Learning}, volume~70,
  pages 1597--1606, 2017.

\bibitem{RED_NeRF}
Berk Iskender, Sushan Nakarmi, Nitin Daphalapurkar, Marc~L. Klasky, and Yoram
  Bresler.
\newblock Rsr-nf: Neural field regularization by static restoration priors for
  dynamic imaging.
\newblock {\em Arxiv: 2503.10015}, 2025.

\bibitem{NTK}
Arthur Jacot, Franck Gabriel, and Cl\'{e}ment Hongler.
\newblock Neural tangent kernel: convergence and generalization in neural
  networks.
\newblock In {\em Proceedings of the 32nd International Conference on Neural
  Information Processing Systems (NeurIPS)}, page 8580–8589, 2018.

\bibitem{SIAM_untrained}
Tim Jahn and Bangti Jin.
\newblock Early stopping of untrained convolutional neural networks.
\newblock {\em SIAM Journal on Imaging Sciences}, 17(4):2331--2361, 2024.

\bibitem{WIRE_ICLR_25}
Dhananjaya Jayasundara, Heng Zhao, Demetrio Labate, and Vishal~M. Patel.
\newblock {PIN}: Prolate spheroidal wave function-based implicit neural
  representations.
\newblock In {\em The Thirteenth International Conference on Learning
  Representations (ICLR)}, 2025.

\bibitem{CVPR_3D_scene}
Chiyu Jiang, Avneesh Sud, Ameesh Makadia, Jingwei Huang, Matthias Nießner, and
  Thomas Funkhouser.
\newblock Local implicit grid representations for 3d scenes.
\newblock In {\em 2020 IEEE/CVF Conference on Computer Vision and Pattern
  Recognition (CVPR)}, pages 6000--6009, 2020.

\bibitem{CAI_PINN_InstantNGP}
Ge~Jin, Deyou Wang, Jian~Cheng Wong, and Shipeng Li.
\newblock Differentiable hash encoding for physics-informed neural networks.
\newblock In {\em 2024 IEEE Conference on Artificial Intelligence (CAI)}, pages
  444--447, 2024.

\bibitem{PIXEL}
Namgyu Kang, Byeonghyeon Lee, Youngjoon Hong, Seok-Bae Yun, and Eunbyung Park.
\newblock Pixel: physics-informed cell representations for fast and accurate
  pde solvers.
\newblock In {\em Proceedings of the Thirty-Seventh AAAI Conference on
  Artificial Intelligence}, pages 8186--8194, 2023.

\bibitem{PIG}
Namgyu Kang, Jaemin Oh, Youngjoon Hong, and Eunbyung Park.
\newblock {PIG}: Physics-informed gaussians as adaptive parametric mesh
  representations.
\newblock In {\em The Thirteenth International Conference on Learning
  Representations}, 2025.

\bibitem{Nonlinear_AAAI}
Nikos Kargas and Nicholas~D. Sidiropoulos.
\newblock Nonlinear system identification via tensor completion.
\newblock In {\em Proceedings of the AAAI Conference on Artificial
  Intelligence}, volume~34, pages 4420--4427, 2020.

\bibitem{TSP_CP}
Nikos Kargas and Nicholas~D. Sidiropoulos.
\newblock Supervised learning and canonical decomposition of multivariate
  functions.
\newblock {\em IEEE Transactions on Signal Processing}, 69:1097--1107, 2021.

\bibitem{NRP_PINN}
George~Em Karniadakis, Ioannis~G. Kevrekidis, Lu~Lu, Paris Perdikaris, Sifan
  Wang, and Liu Yang.
\newblock Physics-informed machine learning.
\newblock {\em Nature Reviews Physics}, 3:422--440, 2021.

\bibitem{INCODE}
Amirhossein Kazerouni, Reza Azad, Alireza Hosseini, Dorit Merhof, and Ulas
  Bagci.
\newblock Incode: Implicit neural conditioning with prior knowledge embeddings.
\newblock In {\em 2024 IEEE/CVF Winter Conference on Applications of Computer
  Vision (WACV)}, pages 1287--1296, 2024.

\bibitem{SIAM_review}
Tamara~G. Kolda and Brett~W. Bader.
\newblock Tensor decompositions and applications.
\newblock {\em SIAM Review}, 51(3):455--500, 2009.

\bibitem{NO}
Nikola Kovachki, Zongyi Li, Burigede Liu, Kamyar Azizzadenesheli, Kaushik
  Bhattacharya, Andrew Stuart, and Anima Anandkumar.
\newblock Neural operator: learning maps between function spaces with
  applications to pdes.
\newblock {\em Journal of Machine Learning Research}, 24(1), 2023.

\bibitem{Universal_Approximation}
Anastasis Kratsios.
\newblock The universal approximation property: Characterization, construction,
  representation, and existence.
\newblock {\em Annals of Mathematics and Artificial Intelligence},
  89(5-6):435--469, 2021.

\bibitem{smooth_matrix_factorization_1991}
Peter Kunkel and Volker Mehrmann.
\newblock Smooth factorizations of matrix valued functions and their
  derivatives.
\newblock {\em Numerische Mathematik}, 60:115--131, 1991.

\bibitem{AAAI24_Fourierbases}
Jason Chun~Lok Li, Chang Liu, Binxiao Huang, and Ngai Wong.
\newblock Learning spatially collaged fourier bases for implicit neural
  representation.
\newblock In {\em Proceedings of the AAAI Conference on Artificial
  Intelligence}, volume~38, pages 13492--13499, 2024.

\bibitem{Li_bound}
Jian Li, Xuanyuan Luo, and Mingda Qiao.
\newblock On generalization error bounds of noisy gradient methods for
  non-convex learning.
\newblock In {\em International Conference on Learning Representations}, 2020.

\bibitem{Superpixel_INR_ECCV}
Jiayi Li, Xile Zhao, Jianli Wang, Chao Wang, and Min Wang.
\newblock Superpixel-informed implicit neural representation for
  multi-dimensional data.
\newblock In {\em 18th European Conference on Computer Vision (ECCV)}, page
  258–276, 2024.

\bibitem{D-FNO}
Kangjie Li and Wenjing Ye.
\newblock {D-FNO}: A decomposed {F}ourier neural operator for large-scale
  parametric partial differential equations.
\newblock {\em Computer Methods in Applied Mechanics and Engineering},
  436:117732, 2025.

\bibitem{Shihua_ST_INR_NAR}
Shang Li, Kuo Gai, Kangning Dong, Yiyang Zhang, and Shihua Zhang.
\newblock High-density generation of spatial transcriptomics with stage.
\newblock {\em Nucleic Acids Research}, 52(9):4843--4856, 2024.

\bibitem{DRO-TFF}
Yanyi Li, Xi~Zhang, Yisi Luo, and Deyu Meng.
\newblock Deep rank-one tensor functional factorization for multi-dimensional
  data recovery.
\newblock In {\em Proceedings of the AAAI Conference on Artificial
  Intelligence}, volume~39, pages 18539--18547, 2025.

\bibitem{Wang_JCP}
Yunlong Li and Fei Wang.
\newblock Local randomized neural networks with finite difference methods for
  interface problems.
\newblock {\em Journal of Computational Physics}, 529:113847, 2025.

\bibitem{INRR}
Zhemin Li, Hongxia Wang, and Deyu Meng.
\newblock Regularize implicit neural representation by itself.
\newblock In {\em 2023 IEEE/CVF Conference on Computer Vision and Pattern
  Recognition (CVPR)}, pages 10280--10288, 2023.

\bibitem{E-NeRV}
Zizhang Li, Mengmeng Wang, Huaijin Pi, Kechun Xu, Jianbiao Mei, and Yong Liu.
\newblock {E-NeRV}: Expedite neural video representation with disentangled
  spatial-temporal context.
\newblock In {\em 17th European Conference on Computer Vision (ECCV)}, page
  267–284, 2022.

\bibitem{FNO}
Zongyi Li, Nikola~Borislavov Kovachki, Kamyar Azizzadenesheli, Burigede liu,
  Kaushik Bhattacharya, Andrew Stuart, and Anima Anandkumar.
\newblock Fourier neural operator for parametric partial differential
  equations.
\newblock In {\em International Conference on Learning Representations}, 2021.

\bibitem{CoordX}
Ruofan Liang, Hongyi Sun, and Nandita Vijaykumar.
\newblock Coord{X}: Accelerating implicit neural representation with a split
  {MLP} architecture.
\newblock In {\em International Conference on Learning Representations (ICLR)},
  2022.

\bibitem{HSI_fusion_24}
Yu-Jie Liang, Zihan Cao, Shangqi Deng, Hong-Xia Dou, and Liang-Jian Deng.
\newblock Fourier-enhanced implicit neural fusion network for multispectral and
  hyperspectral image fusion.
\newblock In {\em Advances in Neural Information Processing Systems (NeurIPS)},
  volume~37, pages 63441--63465, 2024.

\bibitem{FINER}
Zhen Liu, Hao Zhu, Qi~Zhang, Jingde Fu, Weibing Deng, Zhan Ma, Yanwen Guo, and
  Xun Cao.
\newblock Finer: Flexible spectral-bias tuning in implicit neural
  representation by variableperiodic activation functions.
\newblock In {\em 2024 IEEE/CVF Conference on Computer Vision and Pattern
  Recognition (CVPR)}, pages 2713--2722, 2024.

\bibitem{KAN}
Ziming Liu, Yixuan Wang, Sachin Vaidya, Fabian Ruehle, James Halverson, Marin
  Soljacic, Thomas~Y. Hou, and Max Tegmark.
\newblock {KAN}: {K}olmogorov-{A}rnold networks.
\newblock In {\em The Thirteenth International Conference on Learning
  Representations}, 2025.

\bibitem{ICML_NO}
Miguel Liu-Schiaffini, Julius Berner, Boris Bonev, Thorsten Kurth, Kamyar
  Azizzadenesheli, and Anima Anandkumar.
\newblock Neural operators with localized integral and differential kernels.
\newblock In {\em Proceedings of the 41st International Conference on Machine
  Learning}, pages 32576--32594, 2024.

\bibitem{DeepONet}
Lu~Lu, Pengzhan Jin, Guofei Pang, Zhongqiang Zhang, and George~Em Karniadakis.
\newblock Learning nonlinear operators via {DeepONet} based on the universal
  approximation theorem of operators.
\newblock {\em Nature Machine Intelligence}, 3:218--229, 2021.

\bibitem{LRTFR}
Yisi Luo, Xile Zhao, Zhemin Li, Michael~K. Ng, and Deyu Meng.
\newblock Low-rank tensor function representation for multi-dimensional data
  recovery.
\newblock {\em IEEE Transactions on Pattern Analysis and Machine Intelligence},
  46(5):3351--3369, 2024.

\bibitem{CRNL}
Yisi Luo, Xile Zhao, and Deyu Meng.
\newblock Revisiting nonlocal self-similarity from continuous representation.
\newblock {\em IEEE Transactions on Pattern Analysis and Machine Intelligence},
  47(1):450--468, 2025.

\bibitem{NeurTV}
Yisi Luo, Xile Zhao, Kai Ye, and Deyu Meng.
\newblock Neurtv: Total variation on the neural domain.
\newblock {\em SIAM Journal on Imaging Sciences}, 18(2):1101--1140, 2025.

\bibitem{STINR}
Yisi Luo, Xile Zhao, Kai Ye, and Deyu Meng.
\newblock Stinr: Deciphering spatial transcriptomics via implicit neural
  representation.
\newblock In {\em Proceedings of the IEEE/CVF Conference on Computer Vision and
  Pattern Recognition (CVPR)}, 2025.

\bibitem{NM_2021}
Vivien Marx.
\newblock Method of the year: spatially resolved transcriptomics.
\newblock {\em Nature Methods}, 18:9--14, 2021.

\bibitem{AAAI24_pan}
Ge~Meng, Jingjia Huang, Yingying Wang, Zhenqi Fu, Xinghao Ding, and Yue Huang.
\newblock Progressive high-frequency reconstruction for pan-sharpening with
  implicit neural representation.
\newblock In {\em Proceedings of the AAAI Conference on Artificial
  Intelligence}, volume~38, pages 4189--4197, 2024.

\bibitem{CVPR19_occupancy}
Lars Mescheder, Michael Oechsle, Michael Niemeyer, Sebastian Nowozin, and
  Andreas Geiger.
\newblock Occupancy networks: Learning 3d reconstruction in function space.
\newblock In {\em 2019 IEEE/CVF Conference on Computer Vision and Pattern
  Recognition (CVPR)}, pages 4455--4465, 2019.

\bibitem{NeRF}
Ben Mildenhall, Pratul~P. Srinivasan, Matthew Tancik, Jonathan~T. Barron, Ravi
  Ramamoorthi, and Ren Ng.
\newblock {NeRF}: Representing scenes as neural radiance fields for view
  synthesis.
\newblock In {\em European Conference on Computer Vision}, pages 405--421,
  2020.

\bibitem{Medical_imaging_survey}
Amirali Molaei, Amirhossein Aminimehr, Armin Tavakoli, Amirhossein Kazerouni,
  Bobby Azad, Reza Azad, and Dorit Merhof.
\newblock Implicit neural representation in medical imaging: A comparative
  survey.
\newblock In {\em Proceedings of the IEEE/CVF International Conference on
  Computer Vision (ICCV) Workshops}, pages 2381--2391, 2023.

\bibitem{InstantNGP}
Thomas M\"{u}ller, Alex Evans, Christoph Schied, and Alexander Keller.
\newblock Instant neural graphics primitives with a multiresolution hash
  encoding.
\newblock {\em ACM Transactions on Graphics}, 41(4), 2022.

\bibitem{NIR_fusion}
Seonghyeon Nam, Marcus~A. Brubaker, and Michael~S. Brown.
\newblock Neural image representations for multi-image fusion and layer
  separation.
\newblock In {\em 17th European Conference on Computer Vision (ECCV)}, pages
  216--232, 2022.

\bibitem{NO_NTK}
Mike Nguyen and Nicole Mücke.
\newblock Optimal convergence rates for neural operators.
\newblock {\em ArXiv:2412.17518}, 2025.

\bibitem{Traffic_LRTFR}
Tong Nie, Guoyang Qin, Wei Ma, and Jian Sun.
\newblock Spatiotemporal implicit neural representation as a generalized
  traffic data learner.
\newblock {\em Transportation Research Part C: Emerging Technologies},
  169:104890, 2024.

\bibitem{CVPR20_diff_volume}
Michael Niemeyer, Lars Mescheder, Michael Oechsle, and Andreas Geiger.
\newblock Differentiable volumetric rendering: Learning implicit 3d
  representations without 3d supervision.
\newblock In {\em 2020 IEEE/CVF Conference on Computer Vision and Pattern
  Recognition (CVPR)}, pages 3501--3512, 2020.

\bibitem{TTF}
I.~V. Oseledets.
\newblock Constructive representation of functions in low-rank tensor formats.
\newblock {\em Constructive Approximation}, 37:1--18, 2013.

\bibitem{O-INR}
Sourav Pal, Harshavardhan Adepu, Clinton Wang, Polina Golland, and Vikas Singh.
\newblock Implicit representations via operator learning.
\newblock In {\em Forty-first International Conference on Machine Learning},
  pages 39022--39041, 2024.

\bibitem{earth}
Xiaoduo Pan, Deliang Chen, Baoxiang Pan, Xiaozhong Huang, Kun Yang, Shilong
  Piao, Tianjun Zhou, Yongjiu Dai, Fahu Chen, and Xin Li.
\newblock Evolution and prospects of earth system models: Challenges and
  opportunities.
\newblock {\em Earth-Science Reviews}, 260:104986, 2025.

\bibitem{DeepSDF}
Jeong~Joon Park, Peter Florence, Julian Straub, Richard Newcombe, and Steven
  Lovegrove.
\newblock Deepsdf: Learning continuous signed distance functions for shape
  representation.
\newblock In {\em 2019 IEEE/CVF Conference on Computer Vision and Pattern
  Recognition (CVPR)}, pages 165--174, 2019.

\bibitem{Numerical_Fourier_Analysis}
Gerlind Plonka, Daniel Potts, Gabriele Steidl, and Manfred Tasche.
\newblock {\em Numerical Fourier Analysis}, chapter~7, pages 235--259.
\newblock John Wiley \& Sons, Ltd, 2020.

\bibitem{SIAM_review_TV}
Monica Pragliola, Luca Calatroni, Alessandro Lanza, and Fiorella Sgallari.
\newblock On and beyond total variation regularization in imaging: The role of
  space variance.
\newblock {\em SIAM Review}, 65(3):601--685, 2023.

\bibitem{PINN}
M.~Raissi, P.~Perdikaris, and G.E. Karniadakis.
\newblock Physics-informed neural networks: A deep learning framework for
  solving forward and inverse problems involving nonlinear partial differential
  equations.
\newblock {\em Journal of Computational Physics}, 378:686--707, 2019.

\bibitem{ECCV_22_unified_activation_INR}
Sameera Ramasinghe and Simon Lucey.
\newblock Beyond periodicity: Towards a unifying framework for activations in
  coordinate-mlps.
\newblock In {\em 17th European Conference on Computer Vision (ECCV)}, page
  142–158, 2022.

\bibitem{Implicit_reg_CNN}
Noam Razin, Asaf Maman, and Nadav Cohen.
\newblock Implicit regularization in hierarchical tensor factorization and deep
  convolutional neural networks.
\newblock In {\em Proceedings of the 39th International Conference on Machine
  Learning (ICML)}, volume 162, pages 18422--18462, 2022.

\bibitem{CT_CVPR}
Albert~W. Reed, Hyojin Kim, Rushil Anirudh, K.~Aditya Mohan, Kyle Champley,
  Jingu Kang, and Suren Jayasuriya.
\newblock Dynamic ct reconstruction from limited views with implicit neural
  representations and parametric motion fields.
\newblock In {\em 2021 IEEE/CVF International Conference on Computer Vision
  (ICCV)}, pages 2238--2248, 2021.

\bibitem{RED_TCI}
Edward~T. Reehorst and Philip Schniter.
\newblock Regularization by denoising: Clarifications and new interpretations.
\newblock {\em IEEE Transactions on Computational Imaging}, 5(1):52--67, 2019.

\bibitem{WIRE_theory}
T~Mitchell Roddenberry, Vishwanath Saragadam, Maarten~V. de~Hoop, and Richard
  Baraniuk.
\newblock Implicit neural representations and the algebra of complex wavelets.
\newblock In {\em The Twelfth International Conference on Learning
  Representations (ICLR)}, 2024.

\bibitem{WIRE}
Vishwanath Saragadam, Daniel LeJeune, Jasper Tan, Guha Balakrishnan, Ashok
  Veeraraghavan, and Richard~G. Baraniuk.
\newblock Wire: Wavelet implicit neural representations.
\newblock In {\em 2023 IEEE/CVF Conference on Computer Vision and Pattern
  Recognition (CVPR)}, pages 18507--18516, 2023.

\bibitem{MultiScale_INR}
Vishwanath Saragadam, Jasper Tan, Guha Balakrishnan, Richard~G. Baraniuk, and
  Ashok Veeraraghavan.
\newblock Miner: Multiscale implicit neural representation.
\newblock In {\em 17th European Conference on Computer Vision (ECCV)}, pages
  318--333, 2022.

\bibitem{CG_SDF_survey}
Luiz Schirmer, Tiago Novello, Vinícius {da Silva}, Guilherme Schardong, Daniel
  Perazzo, Hélio Lopes, Nuno Gonçalves, and Luiz Velho.
\newblock Geometric implicit neural representations for signed distance
  functions.
\newblock {\em Computers \& Graphics}, 125:104085, 2024.

\bibitem{INR_Reparameterize}
Kexuan Shi, Xingyu Zhou, and Shuhang Gu.
\newblock Improved implicit neural representation with fourier reparameterized
  training.
\newblock In {\em 2024 IEEE/CVF Conference on Computer Vision and Pattern
  Recognition (CVPR)}, pages 25985--25994, 2024.

\bibitem{spline_INR_registration}
Vasiliki Sideri-Lampretsa, Julian McGinnis, Huaqi Qiu, Magdalini Paschali,
  Walter Simson, and Daniel Rueckert.
\newblock {SINR}: Spline-enhanced implicit neural representation for
  multi-modal registration.
\newblock In {\em International Conference on Medical Imaging with Deep
  Learning}, 2024.

\bibitem{SIREN}
Vincent Sitzmann, Julien Martel, Alexander Bergman, David Lindell, and Gordon
  Wetzstein.
\newblock Implicit neural representations with periodic activation functions.
\newblock In {\em International Conference on Neural Information Processing
  Systems}, volume~33, pages 7462--7473, 2020.

\bibitem{scene_representation}
Vincent Sitzmann, Michael Zollh\"{o}fer, and Gordon Wetzstein.
\newblock Scene representation networks: continuous 3d-structure-aware neural
  scene representations.
\newblock In {\em Proceedings of the 33rd International Conference on Neural
  Information Processing Systems}, pages 1121--1132, 2019.

\bibitem{GNTD}
Tianci Song, Charles Broadbent, and Rui Kuang.
\newblock {GNTD}: reconstructing spatial transcriptomes with graph-guided
  neural tensor decomposition informed by spatial and functional relations.
\newblock {\em Nature Communications}, 14:8276, 2023.

\bibitem{Functional_CP_Longitudinal}
Lucas Sort, Laurent~Le Brusquet, and Arthur Tenenhaus.
\newblock Latent functional parafac for modeling multidimensional longitudinal
  data.
\newblock {\em arXiv}, 2410.18696, 2024.

\bibitem{DVGO}
Cheng Sun, Min Sun, and Hwann-Tzong Chen.
\newblock Direct voxel grid optimization: Super-fast convergence for radiance
  fields reconstruction.
\newblock In {\em 2022 IEEE/CVF Conference on Computer Vision and Pattern
  Recognition (CVPR)}, pages 5449--5459, 2022.

\bibitem{IFWI_JGR}
Jian Sun, Kristopher Innanen, Tianze Zhang, and Daniel Trad.
\newblock Implicit seismic full waveform inversion with deep neural
  representation.
\newblock {\em Journal of Geophysical Research: Solid Earth},
  128(3):e2022JB025964, 2023.

\bibitem{CoIL}
Yu~Sun, Jiaming Liu, Mingyang Xie, Brendt Wohlberg, and Ulugbek~S. Kamilov.
\newblock Coil: Coordinate-based internal learning for tomographic imaging.
\newblock {\em IEEE Transactions on Computational Imaging}, 7:1400--1412, 2021.

\bibitem{Meta_Init_INR}
Matthew Tancik, Ben Mildenhall, Terrance Wang, Divi Schmidt, Pratul~P.
  Srinivasan, Jonathan~T. Barron, and Ren Ng.
\newblock Learned initializations for optimizing coordinate-based neural
  representations.
\newblock In {\em Proceedings of the IEEE/CVF Conference on Computer Vision and
  Pattern Recognition (CVPR)}, pages 2846--2855, 2021.

\bibitem{PE}
Matthew Tancik, Pratul Srinivasan, Ben Mildenhall, Sara Fridovich-Keil, Nithin
  Raghavan, Utkarsh Singhal, Ravi Ramamoorthi, Jonathan Barron, and Ren Ng.
\newblock Fourier features let networks learn high frequency functions in low
  dimensional domains.
\newblock In {\em International Conference on Neural Information Processing
  Systems}, volume~33, pages 7537--7547, 2020.

\bibitem{TenosrNeRF_NIPS}
Jiaxiang Tang, Xiaokang Chen, Jingbo Wang, and Gang Zeng.
\newblock Compressible-composable {N}e{RF} via rank-residual decomposition.
\newblock In {\em Advances in Neural Information Processing Systems (NeurIPS)},
  volume~35, pages 14798--14809, 2022.

\bibitem{TT_function_EUSIPCO}
Petr Tichavský and Ondřej Straka.
\newblock Tensor train approximation of multivariate functions.
\newblock In {\em 2024 32nd European Signal Processing Conference (EUSIPCO)},
  pages 2262--2266, 2024.

\bibitem{DIP}
D.~Ulyanov, A.~Vedaldi, and V.~Lempitsky.
\newblock Deep image prior.
\newblock {\em International Journal of Computer Vision}, 128:1867–1888,
  2020.

\bibitem{FTD_PINN}
Sai~Karthikeya Vemuri, Tim B{\"u}chner, Julia Niebling, and Joachim Denzler.
\newblock Functional tensor decompositions for physics-informed neural
  networks.
\newblock In {\em International Conference on Pattern Recognition}, pages
  32--46, 2025.

\bibitem{FTD-INR}
Sai~Karthikeya Vemuri, Tim Büchner, and Joachim Denzler.
\newblock {F-INR}: Functional tensor decomposition for implicit neural
  representations.
\newblock {\em arXiv:2503.21507}, 2025.

\bibitem{3DV_RED_INR}
Romain Vo, Julie Escoda, Caroline Vienne, and Étienne Decencière.
\newblock Neural field regularization by denoising for 3d sparse-view x-ray
  computed tomography.
\newblock In {\em 2024 International Conference on 3D Vision (3DV)}, pages
  1166--1176, 2024.

\bibitem{transfer_learning_INR_NIPS}
Kushal Vyas, Ahmed~Imtiaz Humayun, Aniket Dashpute, Richard Baraniuk, Ashok
  Veeraraghavan, and Guha Balakrishnan.
\newblock Learning transferable features for implicit neural representations.
\newblock In {\em The Thirty-eighth Annual Conference on Neural Information
  Processing Systems (NeurIPS)}, volume~37, pages 42268--42291, 2024.

\bibitem{NeurIPS_FtSVD}
Andong Wang, Yuning Qiu, Mingyuan Bai, Zhong Jin, Guoxu Zhou, and Qibin Zhao.
\newblock Generalized tensor decomposition for understanding multi-output
  regression under combinatorial shifts.
\newblock In {\em Advances in Neural Information Processing Systems},
  volume~37, pages 47559--47635, 2024.

\bibitem{AAAI_PINN_InstantNGP}
Haoxiang Wang, Tao Yu, Tianwei Yang, Hui Qiao, and Qionghai Dai.
\newblock Neural physical simulation with multi-resolution hash grid encoding.
\newblock {\em Proceedings of the AAAI Conference on Artificial Intelligence},
  38(6):5410--5418, 2024.

\bibitem{ECCV_Jianli}
Jianli Wang and Xile Zhao.
\newblock Functional transform-based low-rank tensor factorization for
  multi-dimensional data recovery.
\newblock In {\em 18th European Conference on Computer Vision (ECCV)}, page
  39–56, 2024.

\bibitem{KAN_NLRR}
Shengrui Wang, Yisi Luo, Sanfu Li, Jiangjun Peng, and Bangyu Wu.
\newblock Efficient seismic random noise attenuation via kan-empowered neural
  low-rank representation.
\newblock {\em IEEE Transactions on Geoscience and Remote Sensing}, 63:1--15,
  2025.

\bibitem{JCP_PINN_NTK}
Sifan Wang, Xinling Yu, and Paris Perdikaris.
\newblock When and why {PINNs} fail to train: A neural tangent kernel
  perspective.
\newblock {\em Journal of Computational Physics}, 449:110768, 2022.

\bibitem{HSI_Unmixing_INR}
Ting Wang, Jizhou Li, Michael~K. Ng, and Chao Wang.
\newblock Nonnegative matrix functional factorization for hyperspectral
  unmixing with nonuniform spectral sampling.
\newblock {\em IEEE Transactions on Geoscience and Remote Sensing}, 62:1--13,
  2024.

\bibitem{l2_reg_NTK}
Colin Wei, Jason~D. Lee, Qiang Liu, and Tengyu Ma.
\newblock {\em Regularization matters: generalization and optimization of
  neural nets v.s. their induced kernel}, pages 9712--9724.
\newblock 2019.

\bibitem{INR_registration_Jacobian_reg}
Jelmer~M Wolterink, Jesse~C Zwienenberg, and Christoph Brune.
\newblock Implicit neural representations for deformable image registration.
\newblock In {\em Proceedings of The 5th International Conference on Medical
  Imaging with Deep Learning}, volume 172, pages 1349--1359, 2022.

\bibitem{NeurSTT}
Fengyi Wu, Simin Liu, Haoan Wang, Bingjie Tao, Junhai Luo, and Zhenming Peng.
\newblock Neural spatial-temporal tensor representation for infrared small
  target detection.
\newblock {\em ArXiv:2412.17302}, 2024.

\bibitem{ICLR_MRI_25}
Qing Wu, Chenhe Du, Xuanyu Tian, Jingyi Yu, Yuyao Zhang, and Hongjiang Wei.
\newblock Moner: Motion correction in undersampled radial {MRI} with
  unsupervised neural representation.
\newblock In {\em The Thirteenth International Conference on Learning
  Representations (ICLR)}, 2025.

\bibitem{MRI_LIIF}
Qing Wu, Yuwei Li, Yawen Sun, Yan Zhou, Hongjiang Wei, Jingyi Yu, and Yuyao
  Zhang.
\newblock An arbitrary scale super-resolution approach for 3d mr images via
  implicit neural representation.
\newblock {\em IEEE Journal of Biomedical and Health Informatics},
  27(2):1004--1015, 2023.

\bibitem{MRI_INR_MICCAI}
Qing Wu, Yuwei Li, Lan Xu, Ruiming Feng, Hongjiang Wei, Qing Yang, Boliang Yu,
  Xiaozhao Liu, Jingyi Yu, and Yuyao Zhang.
\newblock Irem: High-resolution magnetic resonance image reconstruction via
  implicit neural representation.
\newblock In {\em Medical Image Computing and Computer Assisted Intervention
  (MICCAI)}, pages 65--74, 2021.

\bibitem{STP_TIP}
Tongle Wu and Jicong Fan.
\newblock Smooth tensor product for tensor completion.
\newblock {\em IEEE Transactions on Image Processing}, 33:6483--6496, 2024.

\bibitem{SD-LoRA}
Yichen Wu, Hongming Piao, Long-Kai Huang, Renzhen Wang, Wanhua Li, Hanspeter
  Pfister, Deyu Meng, Kede Ma, and Ying Wei.
\newblock {SD}-lo{RA}: Scalable decoupled low-rank adaptation for class
  incremental learning.
\newblock In {\em The Thirteenth International Conference on Learning
  Representations (ICLR)}, 2025.

\bibitem{IGNR}
Xinyue Xia, Gal Mishne, and Yusu Wang.
\newblock Implicit graphon neural representation.
\newblock In {\em Proceedings of The 26th International Conference on
  Artificial Intelligence and Statistics}, volume 206, pages 10619--10634,
  2023.

\bibitem{AdamW_implicit}
Shuo Xie and Zhiyuan Li.
\newblock Implicit bias of adamw: $\ell_\infty$-norm constrained optimization.
\newblock In {\em Proceedings of the 41st International Conference on Machine
  Learning}, pages 54488--54510, 2024.

\bibitem{SP_for_INR}
Dejia Xu, Peihao Wang, Yifan Jiang, Zhiwen Fan, and Zhangyang Wang.
\newblock Signal processing for implicit neural representations.
\newblock In {\em Advances in Neural Information Processing Systems},
  volume~35, pages 13404--13418, 2022.

\bibitem{DS_NeRV}
Hao Yan, Zhihui Ke, Xiaobo Zhou, Tie Qiu, Xidong Shi, and Dadong Jiang.
\newblock {DS-NeRV}: Implicit neural video representation with decomposed
  static and dynamic codes.
\newblock In {\em 2024 IEEE/CVF Conference on Computer Vision and Pattern
  Recognition (CVPR)}, pages 23019--23029, 2024.

\bibitem{INR_lowlight}
Shuzhou Yang, Moxuan Ding, Yanmin Wu, Zihan Li, and Jian Zhang.
\newblock Implicit neural representation for cooperative low-light image
  enhancement.
\newblock In {\em 2023 IEEE/CVF International Conference on Computer Vision
  (ICCV)}, pages 12872--12881, 2023.

\bibitem{HE_NC}
Yitao Yang, Yang Cui, Xin Zeng, Yubo Zhang, Martin Loza, Sung-Joon Park, and
  Kenta Nakai.
\newblock {STAIG}: Spatial transcriptomics analysis via image-aided graph
  contrastive learning for domain exploration and alignment-free integration.
\newblock {\em Nature Communications}, 16:1067, 2025.

\bibitem{NeRF_review}
Mingyuan Yao, Yukang Huo, Yang Ran, Qingbin Tian, Ruifeng Wang, and Haihua
  Wang.
\newblock Neural radiance field-based visual rendering: A comprehensive review.
\newblock {\em arXiv:2404.00714}, 2024.

\bibitem{NeurIPS_21_neural_implicit_surface}
Lior Yariv, Jiatao Gu, Yoni Kasten, and Yaron Lipman.
\newblock Volume rendering of neural implicit surfaces.
\newblock In {\em Advances in Neural Information Processing Systems},
  volume~34, pages 4805--4815, 2021.

\bibitem{SP_smooth}
Tatsuya Yokota, Rafal Zdunek, Andrzej Cichocki, and Yukihiko Yamashita.
\newblock Smooth nonnegative matrix and tensor factorizations for robust
  multi-way data analysis.
\newblock {\em Signal Processing}, 113:234--249, 2015.

\bibitem{YiMa_DIP}
Chong You, Zhihui Zhu, Qing Qu, and Yi~Ma.
\newblock Robust recovery via implicit bias of discrepant learning rates for
  double over-parameterization.
\newblock In {\em Advances in Neural Information Processing Systems},
  volume~33, pages 17733--17744, 2020.

\bibitem{PlenOctrees}
Alex Yu, Ruilong Li, Matthew Tancik, Hao Li, Ren Ng, and Angjoo Kanazawa.
\newblock Plenoctrees for real-time rendering of neural radiance fields.
\newblock In {\em 2021 IEEE/CVF International Conference on Computer Vision
  (ICCV)}, pages 5732--5741, 2021.

\bibitem{CF-INR}
Chang Yu, Yisi Luo, Kai Ye, Xile Zhao, and Deyu Meng.
\newblock Cross-frequency implicit neural representation with self-evolving
  parameters.
\newblock {\em arXiv:2504.10929}, 2025.

\bibitem{Bilevel_INR_MRI}
Hongze Yu, Jeffrey~A. Fessler, and Yun Jiang.
\newblock Bilevel optimized implicit neural representation for scan-specific
  accelerated mri reconstruction.
\newblock {\em Arxiv: 2502.21292}, 2025.

\bibitem{Dictionary_INR}
Gizem Y\"uce, Guillermo Ortiz-Jim\'enez, Beril Besbinar, and Pascal Frossard.
\newblock A structured dictionary perspective on implicit neural
  representations.
\newblock In {\em Proceedings of the IEEE/CVF Conference on Computer Vision and
  Pattern Recognition (CVPR)}, pages 19228--19238, 2022.

\bibitem{ICML_teacher}
Chen Zhang, Steven Tin~Sui Luo, Jason Chun~Lok Li, Yik~Chung Wu, and Ngai Wong.
\newblock Nonparametric teaching of implicit neural representations.
\newblock In {\em Proceedings of the 41st International Conference on Machine
  Learning (ICML)}, volume 235, pages 59435--59458, 2024.

\bibitem{HSI_INR_23}
Kaiwei Zhang, Dandan Zhu, Xiongkuo Min, and Guangtao Zhai.
\newblock Implicit neural representation learning for hyperspectral image
  super-resolution.
\newblock {\em IEEE Transactions on Geoscience and Remote Sensing}, 61:1--12,
  2023.

\bibitem{NeurIPS_Zhang}
Tianjing Zhang, Yuhui Quan, and Hui Ji.
\newblock Cross-scale self-supervised blind image deblurring via implicit
  neural representation.
\newblock In {\em Advances in Neural Information Processing Systems},
  volume~37, pages 7060--7094, 2024.

\bibitem{PINN_review_Fluids}
Chi Zhao, Feifei Zhang, Wenqiang Lou, Xi~Wang, and Jianyong Yang.
\newblock A comprehensive review of advances in physics-informed neural
  networks and their applications in complex fluid dynamics.
\newblock {\em Physics of Fluids}, 36(10):101301, 2024.

\bibitem{PAMI_PCU}
Wenbo Zhao, Xianming Liu, Deming Zhai, Junjun Jiang, and Xiangyang Ji.
\newblock Self-supervised arbitrary-scale implicit point clouds upsampling.
\newblock {\em IEEE Transactions on Pattern Analysis and Machine Intelligence},
  45(10):12394--12407, 2023.

\bibitem{InverseBench}
Hongkai Zheng, Wenda Chu, Bingliang Zhang, Zihui Wu, Austin Wang, Berthy Feng,
  Caifeng Zou, Yu~Sun, Nikola~Borislavov Kovachki, Zachary~E Ross, Katherine
  Bouman, and Yisong Yue.
\newblock Inversebench: Benchmarking plug-and-play diffusion priors for inverse
  problems in physical sciences.
\newblock In {\em The Thirteenth International Conference on Learning
  Representations}, 2025.

\bibitem{cryo-EM}
Ellen~D. Zhong, Tristan Bepler, Joseph~H. Davis, and Bonnie Berger.
\newblock Reconstructing continuous distributions of 3d protein structure from
  cryo-em images.
\newblock In {\em International Conference on Learning Representations}, 2020.

\bibitem{LR_survey}
Xiaowei Zhou, Can Yang, Hongyu Zhao, and Weichuan Yu.
\newblock Low-rank modeling and its applications in image analysis.
\newblock {\em ACM Computing Surveys}, 47(2), 2014.

\bibitem{MLP+InstantNGP}
Hao Zhu, Fengyi Liu, Qi~Zhang, Zhan Ma, and Xun Cao.
\newblock {RHINO}: Regularizing the hash-based implicit neural representation.
\newblock {\em Science China Information Sciences}, 2025.

\bibitem{DINER}
Hao Zhu, Shaowen Xie, Zhen Liu, Fengyi Liu, Qi~Zhang, You Zhou, Yi~Lin, Zhan
  Ma, and Xun Cao.
\newblock Disorder-invariant implicit neural representation.
\newblock {\em IEEE Transactions on Pattern Analysis and Machine Intelligence},
  46(8):5463--5478, 2024.

\bibitem{SUICA}
Qingtian Zhu, Yumin Zheng, Yuling Sang, Yifan Zhan, Ziyan Zhu, Jun Ding, and
  Yinqiang Zheng.
\newblock Suica: Learning super-high dimensional sparse implicit neural
  representations for spatial transcriptomics.
\newblock {\em Arxiv: 2412.01124}, 2024.

\bibitem{MICCAI_24_registration}
Veronika~A. Zimmer, Kerstin Hammernik, Vasiliki Sideri-Lampretsa, Wenqi Huang,
  Anna Reithmeir, Daniel Rueckert, and Julia~A. Schnabel.
\newblock Towards generalised neural implicit representations for image
  registration.
\newblock In {\em Medical Image Computing and Computer Assisted Intervention
  (MICCAI)}, pages 45--55, 2024.

\end{thebibliography}
	
\end{document}